%% file: thesis.tex
\documentclass[12pt,leqno]{report}

\input{header}

\definecolor{rblue}{RGB}{0,70,127}
\definecolor{dyellow}{RGB}{255,221,0}
\definecolor{uorange}{RGB}{248,151,26}
\definecolor{coolgray}{RGB}{215,217,218}
\definecolor{darkgreen}{RGB}{88,123,124}
\definecolor{ured}{RGB}{239,51,34}

\psset{levelsep=2.5em,treesep=0pt,treefit=tight}
\psset{arrowscale=2}

\begin{document}

\title{Tackling Graphical NLP problems with Graph Recurrent Networks}
\author{Linfeng Song}
\thesissupervisor{Professor Daniel Gildea}

\maketitle

\thispagestyle{plain}
\newenvironment{dedication}
{\cleardoublepage \thispagestyle{plain} \vspace*{\stretch{1}}
  \begin{center} \em}
  {\end{center} \vspace*{\stretch{3}} }

\begin{dedication}
To my family
\end{dedication}
\tableofcontents

\begin{curriculumvitae}

Linfeng Song was born and grew up in Yingkou, Liaoning in China.
He graduated with a Bachelor of Software Engineering degree in June 2010 from Northeastern University (China).
He then earned his Master of Computer Science degree in June 2014 from the Chinese Academy of Science in Beijing.
From November 2013 to February 2014, he was a visiting scholar at Singapore University of Technology and Design, working with Professor Yue Zhang.
In September 2014, he started his PhD program in the Department of Computer Science at the University of Rochester, supervised by Professor Daniel Gildea. 
He interned three times at IBM Thomas J. Watson Research Center in 2015, 2017 and 2018, and interned at Bosch Research and Technology Center in 2016.
He is honored to have been working with Dr. Zhiguo Wang, Dr. Haitao Mi, Dr. Lin Zhao, Dr. Abe Ittycheriah, Dr. Wael Hamza, Dr. Radu Florian and Dr. Salim Roukos during the internships. 

The following publications were the result of work conducted during his doctoral study:
\nobibliography*
\begin{itemize}
    \item \bibentry{song2019sembleu}
    \item \bibentry{song-tacl19}
    \item \bibentry{yin2019neural}
    \item \bibentry{song2018nary}
    \item \bibentry{P18-1150}
    \item \bibentry{P18-1030}
    \item \bibentry{peng-acl18}
    \item \bibentry{song-naacl18}
    \item \bibentry{song2018exploring}
    \item \bibentry{song-inlg18}
    \item \bibentry{song-acl17}
    \item \bibentry{song-emnlp16}
    \item \bibentry{song-starsem16}
    \item \bibentry{peng2015conll}
\end{itemize}
\end{curriculumvitae}
\begin{acknowledgments}
First and foremost, I would like to thank my advisor Daniel Gildea.
He has always been very farsighted and knowledgeable. 
During the five years working with him, he was always supportive and patient, especially during my most difficult days in my second year.
Despite having a very busy schedule, he always welcomed my occasional drop in for idea discussions, in addition to maintaining regular weekly meetings.
Even though he had his own research interest and ideas, he never imposed those ideas on me, but encouraged me to pursue research interests of my own. 
I hope to follow his example and become a knowledgeable and sincere researcher in my later academic career.

I would like to thank the people I have worked with over the years. Xiaochang and I have cooperated in several projects and we have written a few papers together. 
I would like to thank Professor Yue Zhang, a significant collaborator since 2014, and Dr. Zhiguo Wang, the mentor of my IBM internships, for discussing and brainstorming on my graph recurrent network development.
I am grateful to Dr. Mo Yu and Dr. Lin Zhao for idea discussion and other accommodation during my internships.
I would also like to thank my other committee members, Lenhart Schubert and Jiebo Luo, for accommodating their time, and providing their valuable feedback.

I am grateful to the wonderful people I have met in the department: Michelle, Xiaochang, Dong, Jianbo, Chencheng, Quanzeng, Taylan, Yapeng, Jing, Iftekhar, and a lot of other graduate students. 
I started working on NLP research as a graduate student advised by Professor Qun Liu from Chinese Academy of Science, and I really appreciate all the help and suggestions I got in the past few years.

Lastly, and more importantly, I want to thank my wife and my parents for their love and support during all these years.
\end{acknowledgments}

\begin{abstract}
How to properly model graphs is a long-existing and important problem in natural language processing, where several popular types of graphs are knowledge graphs, semantic graphs and dependency graphs.
Comparing with other data structures, such as sequences and trees, graphs are generally more powerful in representing complex correlations among entities.
For example, a knowledge graph stores real-word entities (such as ``Barack\_Obama'' and ``U.S.'') and their relations (such as ``live\_in'' and ``lead\_by'').
Properly encoding a knowledge graph is beneficial to user applications, such as question answering and knowledge discovery.
Modeling graphs is also very challenging, probably because graphs usually contain massive and cyclic relations.
For instance, a tree with $n$ nodes has $n-1$ edges (relations), while a complete graph with $n$ nodes can have $O(n^2)$ edges (relations).

Recent years have witnessed the success of deep learning, especially RNN-based models, on many NLP problems, including machine translation \cite{cho-EtAl:2014:EMNLP2014} and question answering \cite{shen2017reasonet}.
Besides, RNNs and their variations have been extensively studied on several graph problems and showed preliminary successes.
Despite the successes that have been achieved, RNN-based models suffer from several major drawbacks.
First, they can only consume sequential data, thus linearization is required to serialize input graphs, resulting in the loss of important structural information.
In particular, originally closely located graph nodes can be very far away after linearization, and this introduces great challenge for RNNs to model their relation.
Second, the serialization results are usually very long, so it takes a long time for RNNs to encode them.

In this thesis, we propose a novel graph neural network, named graph recurrent network (GRN).
GRN takes a hidden state for each graph node, and it relies on an iterative message passing framework to update these hidden states in parallel. 
Within each iteration, neighboring nodes exchange information between each other, so that they absorb more global knowledge.
Different from RNNs, which require absolute orders (such as left-to-right orders) for execution, our GRN only require relative neighboring information, making it very general and flexible on a variety of data structures.

We study our GRN model on 4 very different tasks, such as machine reading comprehension, relation extraction and machine translation.
Some tasks (such as machine translation) require generating sequences, while others only require one decision (classification).
Some take undirected graphs without edge labels, while the others have directed ones with edge labels.
To consider these important differences, we gradually enhance our GRN model, such as further considering edge labels and adding an RNN decoder.
Carefully designed experiments show the effectiveness of GRN on all these tasks.

\end{abstract}

\begin{contributors}
This work was supervised by a dissertation committee consisting of Professors Daniel Gildea (advisor), Lenhart Schubert, and Jiebo Luo from the Department of Computer Science and Professor Yue Zhang from the Westlake University.
Chapter~\ref{chap:mhqa} is based on~\citet{song2018exploring}, which is supervised by my internship co-mentors Zhiguo Wang and Mo Yu.
Chapter~\ref{chap:nary} is based on~\citet{song2018nary}, and Chapter~\ref{chap:amrgen} is based on~\citet{P18-1150}. 
Yue Zhang and Zhiguo Wang are highly involved in proposing the graph recurrent network. 
Chapter~\ref{chap:semnmt} is based on~\citet{song-tacl19}.
This material is based upon work supported by National Science Foundation awards \#IIS-1813823, IIS-1446996, NYS Center of Excellence award and a gift from Google Inc. Any opinions, findings, and conclusions or recommendations expressed in this material are those of the author(s) and do not necessarily reflect the views of above-named organizations.
\end{contributors}
\listoftables
\listoffigures
\chapter{Introduction}
\input{1-chap-intro}
\break

\chapter{Background: Encoding graphs with Neural Networks}
\label{chap:background}
\input{2-chap-background}

\break

\chapter{Graph Recurrent Network for Multi-hop Reading Comprehension}
\label{chap:mhqa}
\input{3-chap-MHQA}

\break

\chapter{Graph Recurrent Network for $n$-ary Relation Extraction}
\label{chap:nary}
\input{4-chap-nary}
\break

\chapter{Graph Recurrent Network for AMR-to-text Generation}
\label{chap:amrgen}
\input{5-chap-amrgen}
\break

\chapter{Graph Recurrent Network for Semantic NMT using AMR}
\label{chap:semnmt}
\input{6-chap-semnmt}
\break

\chapter{Conclusion}
\label{chap:conclusion}
In this dissertation, we have introduced graph recurrent networks (GRN), which is general enough to encode graphs of arbitrary types without destroying the original graph structure.
We investigated our GRN on 3 types of graphs across 4 different tasks, including multi-hop reading comprehension (Chapter \ref{chap:mhqa}), $n$-ary relation extraction (Chapter \ref{chap:nary}), AMR-to-text generation (Chapter \ref{chap:amrgen}) and semantic neural machine translation (Chapter \ref{chap:semnmt}).
On each task, our carefully designed experiments show that GRN is significantly better than baselines based on sequence-to-sequence models and DAG networks.
Besides, our GRN allows better parallelism than sequence-to-sequence and DAG models, and as a result it is much faster than these baselines.
We also theoretically compare GRN with other graph neural networks, such as graph convolutional network (GCN) \citep{kipf2017semi} and gated graph neural network (GGNN) \citep{li2015gated}, with a message passing framework (Chapter \ref{chap:background}).

Despite being successful on several tasks, there is still much room for improving GRN, thus I consider refining GRN as part of my future work.
One direction of refinement is to further differentiate the node states after each GRN step.
Previously, we rely on development experiments to carefully choose a proper number of GRN steps, and different tasks require very different number.
To alleviate this problem, we can use the node states of all steps instead of the last step. 
By using the GRN outputs of all steps, we ask the model to pick node states of proper GRN steps based on the task performance.
Another direction for improving GRN is to alleviate the massive memory usage (This problem exists for the other graph neural networks too) by sampling.
\citet{chen2018fastgcn} have investigated importance sampling according to the significance of the nodes in a citation graph.
However, relations are also very important for many types of graphs in the NLP area, such as the semantic graphs and dependency graphs.
We will study a proper relation-oriented sampling approach to reduce the memory usage of GRN.

\break

\bibliographystyle{urcsbiblio}
\bibliography{all}
\appendix

\end{document}

%% file: header.tex
\usepackage{urcsbiblio,urcsthesis}
\usepackage{url}
\usepackage{latexsym}
\usepackage{amsmath}
\usepackage{bibentry}
\usepackage{natbib}
\usepackage{graphicx}
\usepackage{multirow}
\usepackage[utf8]{inputenc}
\usepackage{pst-all}
\usepackage{color}
\usepackage{psfrag}
\usepackage{pstricks}
\usepackage{pstricks-add}
\usepackage{amsmath,bm}
\usepackage{array}
\usepackage{palatino}
\usepackage{mathpazo}
\usepackage{subcaption}
\usepackage{tikz-qtree}
\usepackage{tikz}

\usepackage{avm}
\avmfont{\sc}
\avmvalfont{\sf}
\font\seveni=cmti7
\avmsortfont{\seveni} 

\tikzset{
    events/.style={ellipse, draw, align=center},
}
\usepackage{caption}
\usepackage{subcaption}
\usepackage{multirow}
\usepackage{rotating} 

\usepackage{enumerate}
\psset{levelsep=2.5em,treesep=0pt,treefit=tight}
\usepackage{amsmath,amssymb}
\newcommand{\boxnum}[1]{{\setlength{\fboxsep}{1pt}\raisebox{1pt}
{\hspace{1pt}\fbox{\tiny{\ensuremath{#1}}}\hspace{1pt}}}}
\newcommand{\ind}[1]{\ensuremath{^{\kern-0.5pt\boxnum{#1}}}}

\newcommand{\eat}[1]{}

\usepackage{rotating} 

\newcommand{\qed}{\nobreak \ifvmode \relax \else
      \ifdim\lastskip<1.5em \hskip-\lastskip
            \hskip1.5em plus0em minus0.5em \fi \nobreak
                  \vrule height0.75em width0.5em depth0.25em\fi}

\usepackage{xspace}

\usepackage{booktabs}

\usepackage{latexsym}
\usepackage{natbib}
\usepackage{tabularx}
\usepackage{graphicx}
\usepackage{amsmath}
\usepackage{amssymb}
\usepackage{dsfont}
\usepackage{color}
\usepackage{pifont}
\usepackage{bm}
\usepackage{tablefootnote}
\usepackage{threeparttable}

\usepackage{url}
\usepackage{color}
\usepackage{graphicx}
\usepackage{amsmath}
\usepackage{amssymb}
\usepackage{tabularx}
\usepackage{multirow}
\usepackage{mathtools}

\usepackage{graphicx}
\usepackage{url}
\usepackage{amsmath}
\usepackage{amssymb}
\usepackage[linesnumbered]{algorithm2e}
\usepackage{color}
\usepackage{tabularx}
\usepackage{tablefootnote}



%% file: 1-chap-intro.tex
\section{Graph problems in NLP}

\begin{figure}
    \centering
    \begin{subfigure}[b]{0.38\textwidth}
        \includegraphics[width=\textwidth]{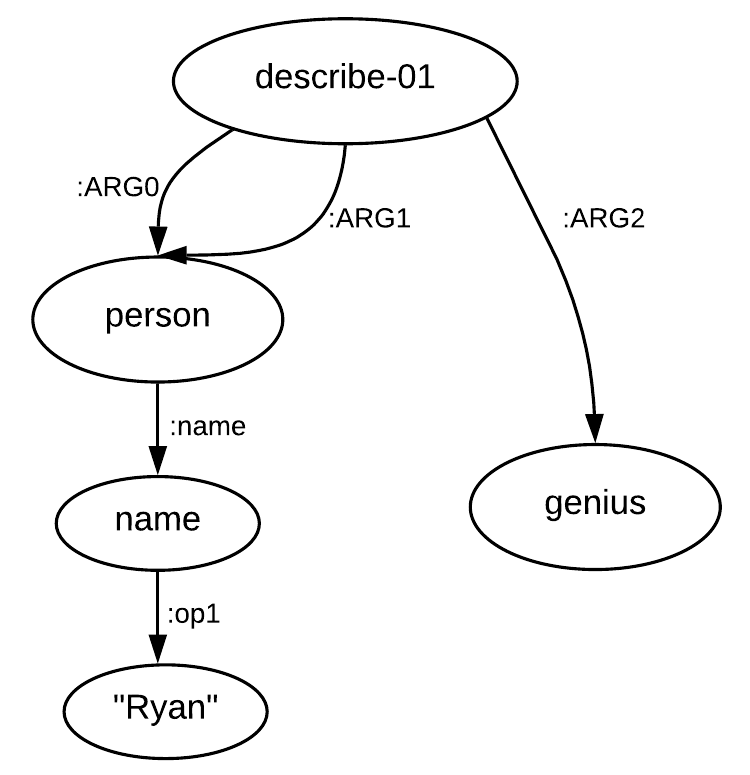}
        \caption{A semantic graph meaning ``Ryan's description of himself: a genius.''}
        \label{fig:1_amr}
    \end{subfigure}
    ~ 
    \begin{subfigure}[b]{0.45\textwidth}
        \includegraphics[width=\textwidth]{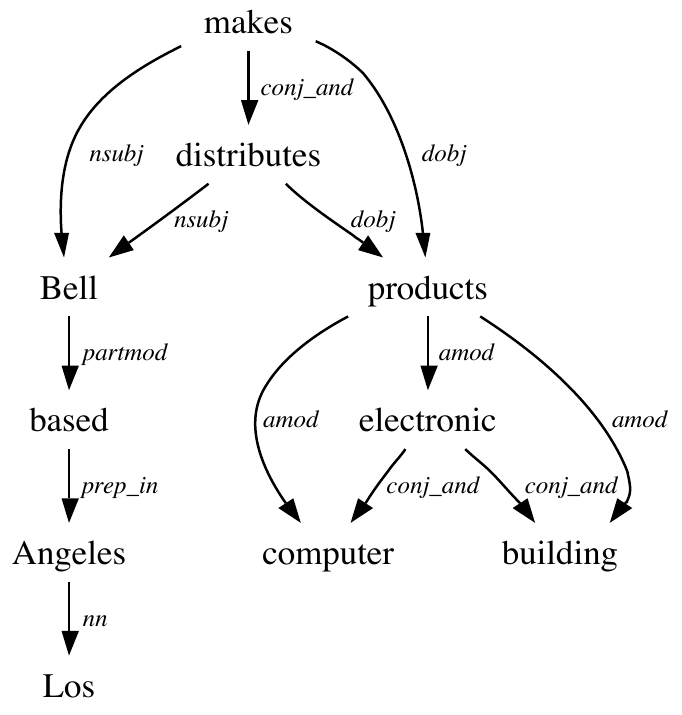}
        \caption{A dependency graph.}
        \label{fig:1_dep}
    \end{subfigure}
    ~ 
    \begin{subfigure}[b]{0.8\textwidth}
        \includegraphics[width=\textwidth]{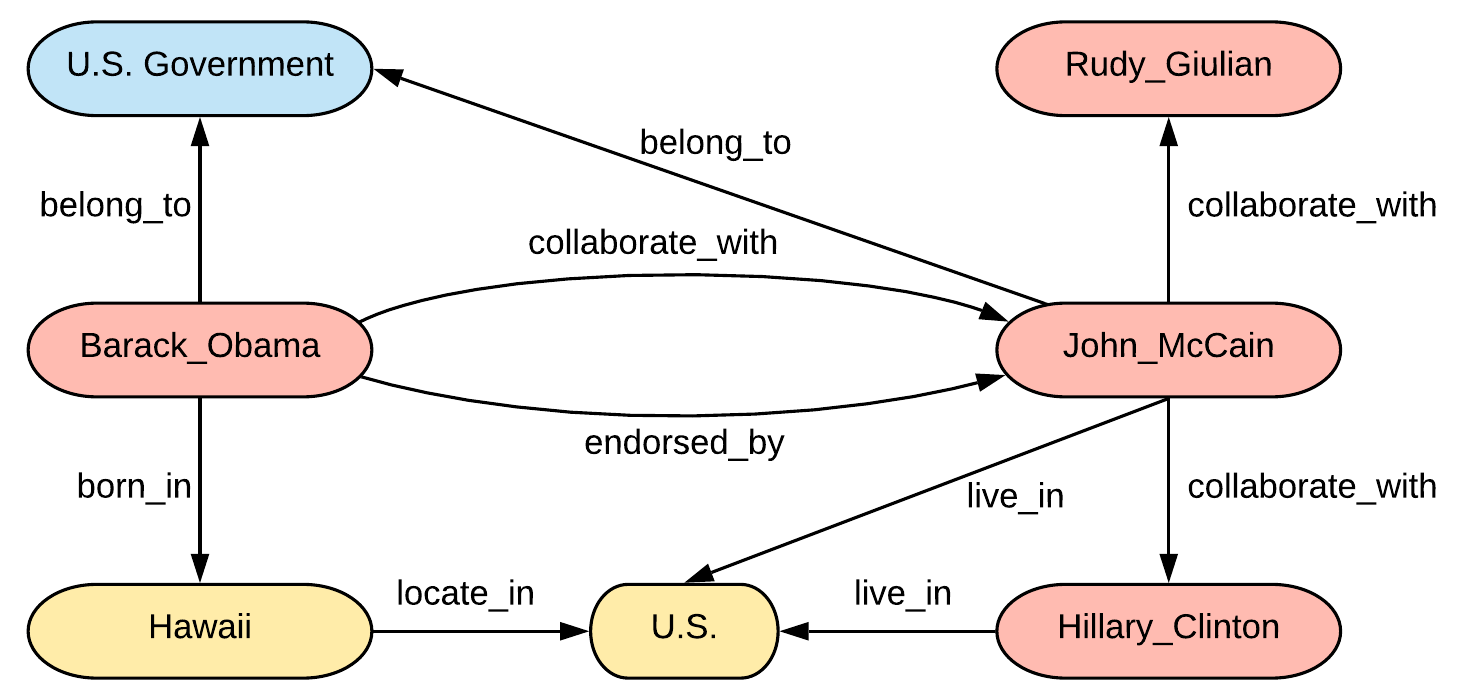}
        \caption{A small fraction of a large knowledge graph.}
        \label{fig:1_kb}
    \end{subfigure}
    \caption{Several types of graphs in NLP area.}\label{fig:1_graph_demo}
\end{figure}

There are plenty of graph problems in the natural language processing area.
These graphs include semantic graphs, dependency graphs, knowledge graphs and so on.
Figure \ref{fig:1_graph_demo} shows an example for several types of graphs, where a semantic graph (Figure \ref{fig:1_amr}) visualizes the underlining meaning (such as who does what to whom) of a given sentence by abstracting the sentence into several concepts (such as ``describe-01'' and ``person'') and their relations (such as ``:ARG0'' and ``:ARG1'').
On the other hand, a dependency graph (Figure \ref{fig:1_dep}) simply captures word-to-word dependencies, such as ``Bell'' being the subject (\emph{subj}) of ``makes''.
Finally, a knowledge graph (Figure \ref{fig:1_kb}) represents real-world knowledge by entities (such as ``U.S. Government'' and ``Barack\_Obama'') and their relations (such as ``belong\_to'' and ``born\_in'').
Since there is massive information in the world level, a knowledge graph (such as Freebase\footnote{https://developers.google.com/freebase/} and DBPedia\footnote{https://wiki.dbpedia.org/}) are very large.

\section{Previous approaches for modeling graphs}

How to properly model these graphs has been a long-standing and important topic, as this directly contributes to natural language understanding, one of the most important key problems in NLP.
Previously, statistical or rule-based approaches have been introduced to model graphs.
For instance, synchronous grammar-based methods have been proposed to model semantic graphs \citep{jones2012semantics,jeff2016amrgen,song-acl17} and dependency graphs \citep{xie2011novel,meng2013translation} for machine translation and text generation.
For modeling knowledge graphs, very different approaches have been adopted, probably due to the fact that their scale is too large.
One popular method is random walk, which has been investigated for knowledge base completion \citep{lao2011random} and entity linking \citep{han2011collective,xue2019neural}.

Recently, research on analyzing graphs with deep learning models has been receiving more and more attention.
This is because they have demonstrated strong learning power and other superior properties, i.e. not needing feature engineering and benefiting from large-scale data.
To date, people have studied several types of neural networks.

\begin{figure}
    \centering
    \includegraphics[width=0.8\textwidth]{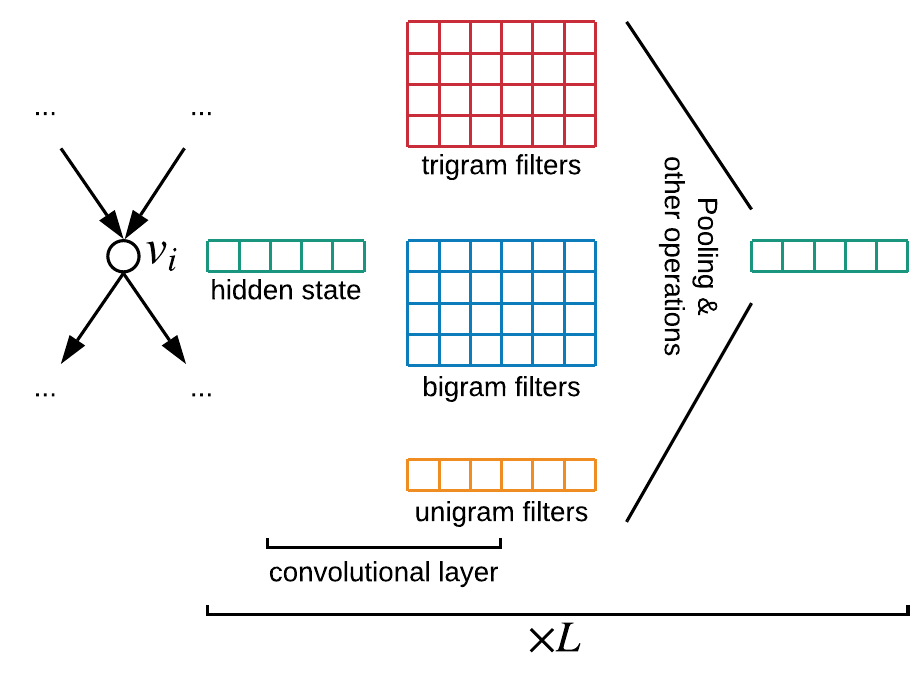}
    \caption{Graph encoding with $L$ CNN layers.}
    \label{fig:1_cnn_example}
\end{figure}

\subparagraph{CNN}
One group of models \citep{defferrard2016convolutional,niepert2016learning,duvenaud2015convolutional,henaff2015deep} adopt convolutional neural networks (CNN) \citep{lecun1995convolutional} for encoding graphs. 
As shown in Figure \ref{fig:1_cnn_example}, these models adopts multiple convolution layers, each capturing the local correspondences within $n$-gram windows.
By stacking the layers, more global correspondences can be captured.
One drawback is the large amount of computation, because CNNs calculate features by enumerating all $n$-gram windows, and there can be a large number of $n$-gram ($n>$1) windows for a very dense graph.
In particular, each node in a complete graph of $N$ nodes has $N$ left and $N$ right neighbors, so there are $O(N^2)$ bigram, $O(N^3)$ trigram and $O(N^4)$ 4-gram windows, respectively.
On the other hand, previous attempts at modeling text with CNN do not suffer from this problem, as a sentence with $N$ words only has $O(N)$ windows for each $n$-gram.
This is because a sentence can be viewed as a chain graph, where each node has only one left neighbor and one right neighbor.

\begin{figure}
    \centering
    \includegraphics[width=\textwidth]{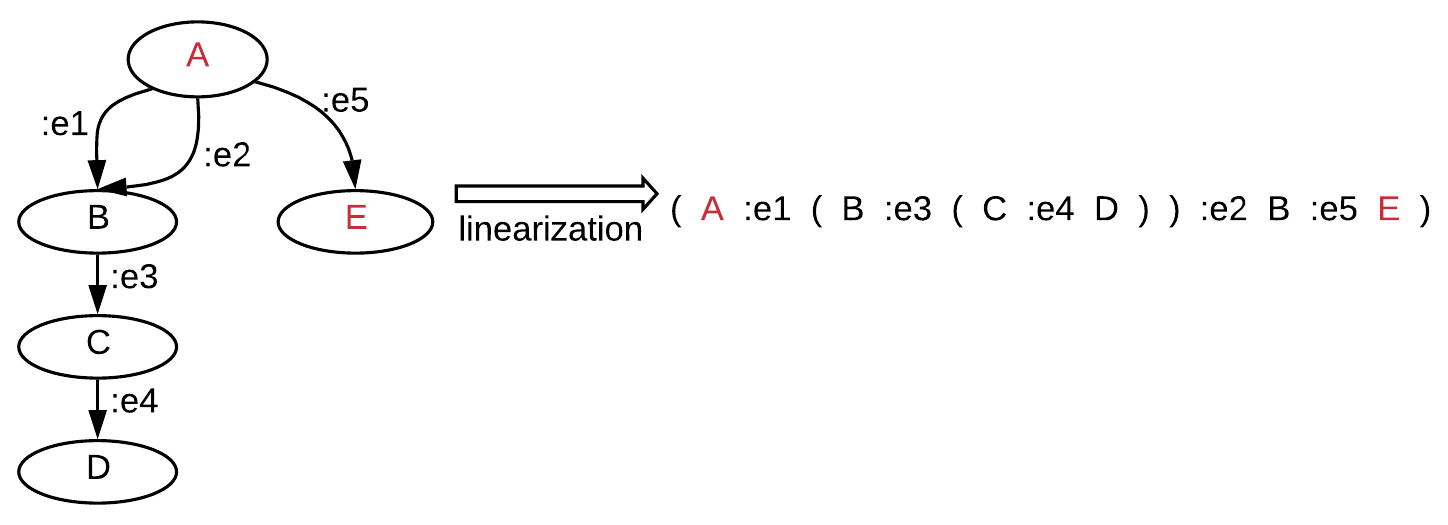}
    \caption{Graph linearization by depth-first traversal. ``A'' and ``E'' are directly connected in the original graph, but fell apart after linearization.}
    \label{fig:1_graph_serielize}
\end{figure}

\subparagraph{RNN}
Another direction is applying RNNs on linearized graphs \citep{konstas-EtAl:2017:Long,li-EtAl:2017:Long} based on depth-first traversal algorithms.
Usually a bidirectional RNN is adopted to capture global dependencies within the whole graph.
Comparing with CNNs, the computations of a RNN is only linear in terms of graph scale.
However, one drawback of this direction is that some structural information is lost after linerization.
Figure \ref{fig:1_graph_serielize} shows a linearization result, where nodes ``A'' and ``E'' are far apart, while they are directly connected in the original graph.
To alleviate this problem, previous methods insert brackets into their linearization results to indicate the original structure, and they hope RNNs can figure that out with the aid of brackets.
But it is still uncertain this method can recover all the information loss.
Later work \citep{P18-1150,song-tacl19} show large improvements by directly modeling original graphs without linearization.
This indicates that simply inserting brackets does not handle this problem very well.

\section{Motivation and overview of our model}
In this dissertation, we want to explore better alternatives for encoding graphs, which are general enough to be applied on arbitrary graphs without destroying the original structures.
We introduce graph recurrent networks (GRNs) and show that they are successful in handling a variety of graph problems in the NLP area. 
Note that there are other types of graph neural networks, such as graph convolutional network (GCN) and gated graph neural network (GGNN).
I will give a comprehensive comparison in Chapter \ref{sec:2_gnn_comp}.

Given an input graph, GRN adopts a hidden state for each graph node.
In order to capture non-local interaction between nodes, it allows information exchange between neighboring nodes through time.
At each time step, each node propagates its information to each other node that has a direct connection so that every node absorbs more and more global information through time.
Each resulting node representation contains information from a large context surrounding it, so the final representations can be very expressive for solving other graph problems, such as graph-to-sequence learning or graph classification.
We can see that our GRN model only requires local and \emph{relative} neighboring information, rather than an \emph{absolute} topological order of all graph nodes required by RNNs.
As a result, GRN can work on arbitrary graph structures with or without cycles.
From the global view, GRN takes the collection of all node states as the state for the entire graph, and the neighboring information exchange through time can be considered as a graph state transition process.
The graph state transition is a recurrent process, where the state is recurrently updated through time (This is reason we name it ``graph recurrent network'').
Comparatively, a regular RNN absorbs a new token to update its state at each time.
On the other hand, our GRN lets node states exchange information for updating the graph state through time.

In this thesis, I will introduce the application of GRN on 3 types of popular graphs in the NLP area, which are dependency graphs, semantic graphs and another type of graphs (evidence graphs) that are constructed from textual input for modeling entities and their relations.
The evidence graphs are created from documents to represent entities (such as ``Time Square'', ``New York City'' and ``United States'') and their relations for QA-oriented reasoning.

\section{Thesis Outline}
The remainder of the thesis is organized as follows.
\begin{itemize}
\item \textbf{Chapter~\ref{chap:background}: Background}
In this chapter, we briefly introduce previous deep learning models for encoding graphs. 
We first discuss applying RNNs and DAG networks on graphs, including their shortcomings.
As a next step, we describe several types of graph neural networks (GNNs) in more detail, then systematically compare GNNs with RNNs.
Finally, I point out one drawback of GNNs when encoding large-scale graphs, before showing some existing solutions.
\item \textbf{Chapter~\ref{chap:mhqa}: Graph Recurrent Network for Multi-hop Reading Comprehension}
In this chapter, I will first describe the multi-hop reading comprehension task, then propose a graph representation for each input document.
To encode the graphs for global reasoning, we introduce 3 models for this task, including one RNN baseline, one baseline with DAG network and our GRN-based model.
For fair comparison, all three models are in the same framework, with the only difference being how to encode the graph representations.
Finally, our comprehensive experiments show the superiority of our GRN.
\item \textbf{Chapter~\ref{chap:nary}: Graph Recurrent Network for $n$-ary Relation Extraction} 
In this chapter, we extend our GRN from undirected and edge-unlabeled graphs (as in Chapter \ref{chap:mhqa}) to dependency graphs for solving a medical relation extraction (classification) problem.
The goal is to determine whether a given medicine is effective on cancers caused by a type of mutation on a certain gene.
Previous work has shown the effectiveness of incorporating rich syntactic and discourse information.
The previous state of the art propose DAN networks by splitting the dependency graphs into two DAGs.
Arguing that important information is lost by splitting the original graphs, we adapt GRN on the dependency graphs without destroying any graph structure.
\item \textbf{Chapter~\ref{chap:amrgen}: Graph Recurrent Network for AMR-to-text Generation}
In this chapter, we propose a graph-to-sequence model by extending GRN with an attention-based LSTM, and evaluate our model on AMR-to-text generation.
AMR is a semantic formalism based on directed and edge-labelled graphs, and the task of AMR-to-text generation aims at recovering the original sentence of a given AMR graph.
In our extensive experiments, our model show consistently better performance than a sequence-to-sequence baseline with a Bi-LSTM encoder under the same configuration, demonstrating the superiority of our GRN over other sequential encoders.
\item \textbf{Chapter~\ref{chap:semnmt}: Graph Recurrent Network for Semantic NMT using AMR} 
In this chapter, we further adapt our GRN-based graph-to-sequence model on AMR-based semantic neural machine translation.
In particular, our model is extended by another Bi-LSTM encoder for modeling source sentences, as they are crucial for translation.
The AMRs are automatically obtained by parsing the source sentences.
Experiments show that using AMR outperforms other common syntactic and semantic representations, such as dependency and semantic role.
We also show that GRN-based encoder is better than a Bi-LSTM encoder using linearized AMRs.
This is consistent with the results of Chapter \ref{chap:amrgen}.
\item \textbf{Chapter~\ref{chap:conclusion}: Conclusion} Finally, I summarize the main contributions of this thesis, and propose some future research directions for further improving our graph recurrent network.
\end{itemize}

%% file: 2-chap-background.tex
Graphs are a kind of structure for representing a set of concepts (graph nodes) and their relations (graph edges).
Studying graphs is a very important topic in research, with one reason being that there have been numerous types of graphs in our daily lives, such as social networks, traffic network and molecules.
Having a deeper understanding of these graphs can lead to better friend recommendation via social media, less traffic jam in busy cities and more effective medicine.
However, existing probabilistic approaches and statistical models are not very successful on modeling graphs.
It is possibly because graphs are usually very complex and large, making these approaches inefficient or not distinguishing enough.
Recent years have witnessed the success of deep learning approaches, which have been shown to be more expressive and can benefit more from large-scale training. 
Also, recent advances on GPUs, especially the massive parallelism for tensor operations, make the deep neural-network models very efficient.
As a result, there have been many attempts on investigating neural networks on dealing with graphs.

In this chapter, we will focus on the graph problems in the natural language processing (NLP) area. 
In particular, I will first introduce several conventional neural approaches (such as recurrent neural networks) for dealing with these graph problems, before giving a deeper investigation on the more recent graph neural networks (GNN).

\section{Encoding graphs with RNN or DAG network}

\begin{figure}
    \centering
    \includegraphics{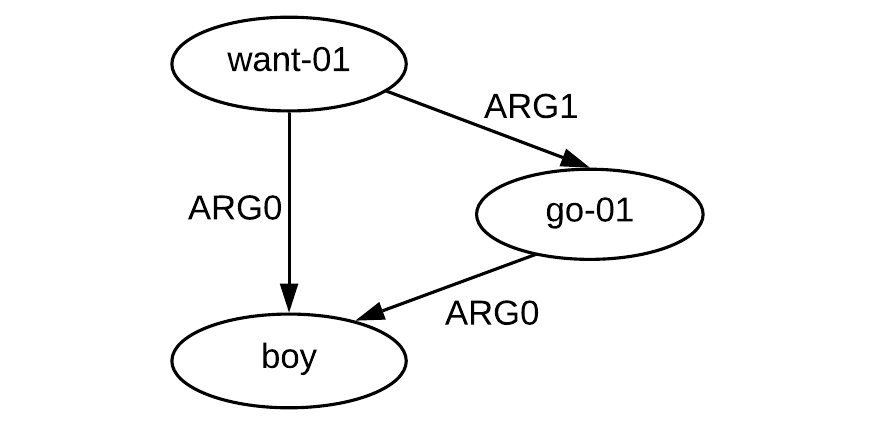}
    \caption{An AMR graph representing ``The boy wants to go.''}
    \label{fig:2_amr_example}
\end{figure}

Since the first great breakthrough of recurrent neural networks (RNN) on machine translation \citep{cho-EtAl:2014:EMNLP2014,bahdanau2015neural}, people have investigated the usefulness of RNN and several of its extensions for solving graph problems in the NLP area.
Below we take abstract meaning representation (AMR) \citep{banarescu-EtAl:2013:LAW7-ID} as an example to demonstrate several existing ways for encoding graphs with RNNs. 
As shown in Figure \ref{fig:2_amr_example}, AMRs are rooted and directed graphs, where the graph nodes (such as ``want-01'' and ``boy'') represent the concepts and edges (such as ``ARG0'' and ``ARG1'') represent the relations between nodes.

\subparagraph{Encoding with RNNs}
To encode AMRs, one kind of approaches \citep{konstas-EtAl:2017:Long} first linearize their inputs with depth-first traversal, before feeding the linearization results into a multi-layer LSTM encoder.
We can see that the linearization causes loss of the structural information.
For instance, originally closely-located graph nodes (such as parents and children) can be very far away, especially when the graph is very large.
In addition, there is not a specific order among the children for a graph node, resulting in multiple linearization possibilities.
This increases the data variation and further introduces ambiguity. 
Despite the drawbacks that mentioned above, these approaches received great success with the aid of large-scale training.
For example, \citet{konstas-EtAl:2017:Long} leveraged 20 million sentences paired with automatically parsed AMR graphs using a multi-layer LSTM encoder, which demonstrates dramatic improvements (5.0+ BLEU points) over the existing statistical models \citep{pourdamghani-knight-hermjakob:2016:INLG,jeff2016amrgen,song-emnlp16,song-acl17}.
This demonstrates the strong learning power of RNNs, but there is still room for improvement due to the above mentioned drawbacks.

\subparagraph{Encoding with DAG networks}
One better alternative than RNNs are DAG networks \cite{zhu-sobhani-guo:2016:N16-1,su2017lattice,P18-1144}, which extend RNN on directed acyclic graphs (DAGs).
Comparing with sentences, where each word has exactly one preceding word and one succeeding word, a node in a DAG can have multiple preceding and succeeding nodes, respectively.
To adapt RNNs on DAGs, the hidden states of multiple preceding nodes are first merged before being applied to calculate the hidden state of the current node.
One popular way for merging preceding hidden states is called ``child-sum'' \citep{tai-socher-manning:2015:ACL-IJCNLP}, which simply sum up their states to product one vector.
Comparing RNNs, DAG networks have the advantage of preserving the original graph structures.
Recently, \citet{takase-EtAl:2016:EMNLP2016} applied a DAG network on encoding AMRs for headline generation, but no comparisons were made to contrast their model with any RNN-based models.
Still, DAG networks are intuitively more suitable for encoding graphs than RNNs.

However, DAG networks suffer from two major problems. 
First, they fail on cyclic graphs, as they require an exact and finite node order to be executed on.
Second, sibling nodes can not incorporate the information of each other, as the encoding procedure can either be bottom-up or top-down, but not both at the same time.
For the first problem, previous work introduces two solutions to adapt DAG networks on cyclic graphs, but to my knowledge, no solution has been available for the second problem.
Still, both solutions for the first problem have their own drawbacks, which I will be discussing here.

One solution \citep{TACL1028} first splits a cyclic graph into two DAGs, which are then encoded with separate DAG networks.
Note that legal splits always exist, one can first decide an order over graph nodes, before separating left-to-right edges from right-to-left ones to make a split.
An obvious drawback is that structural information is lost by splitting a graph into two DAGs.
Chapter \ref{chap:nary} mainly studies this problem and gives our solution.
The other solution \citep{liang2016semantic} is to leverage another model to pick a node order from an undirected and unrooted graph.
Since there are exponential numbers of node orders, more ambiguity is introduced, even through they use a model to pick the order.
Also, preceding nodes cannot incorporate the information from their succeeding nodes.

\section{Encoding graphs with Graph Neural Network}
\label{sec:2_gnn_comp}

Since being introduced, graph neural networks (GNNs) \citep{scarselli2009graph} have long been neglected until recently \citep{li2015gated,kipf2017semi,P18-1030}.
To update node states within a graph, GNNs rely on a message passing mechanism that iteratively updates the node states in parallel.
During an iteration, a message is first \textbf{aggregated} for each node from its neighbors, then the message is \textbf{applied} to update the node state. 
To be more specific, updating the hidden state $\boldsymbol{h}_i$ for node $v_i$ for iteration $t$ can be formalized as the following equations:
\begin{align} 
        \boldsymbol{m}^t_i &= \text{Aggregate}(\boldsymbol{x}_i, \boldsymbol{H}^{t-1}_{\boldsymbol{N_i}}, \boldsymbol{X}_{\boldsymbol{N_i}}) \label{eq:2_aggre} \\
        \boldsymbol{h}^{t}_i &= \text{Apply}(\boldsymbol{h}^{t-1}_i, \boldsymbol{m}^t_i) \text{,} \label{eq:2_apply}
\end{align}
where $\boldsymbol{H}^{t-1}_{\boldsymbol{N_i}}$ and $\boldsymbol{X}_{\boldsymbol{N_i}}$ represent the hidden states and embeddings of the neighbors for $v_i$, and $\boldsymbol{N_i}$ corresponds to the set of neighbors for $v_i$.
We can see that each node gradually absorbs larger context through this message passing framework.
This framework is remotely related to loopy belief propagation (LBP) \citep{murphy1999loopy}, but the main difference is that LBP propagates probabilities, while GNNs propagate hidden-state units.
Besides, the message passing process is only executed for a certain number of times for GNNs, while LBP is usually executed until convergence.
The reason is that GNNs are optimized for end-to-end task performance, not for a joint probability.

\subsection{Main difference between GNNs and RNNs}

The main difference is that GNNs do not require a node order for the input, such as a left-to-right order for sentences and a bottom-up order for trees.
In contrast, having an order is crucial for RNNs and their DAG extensions.
In fact, GNNs only require the local neighborhood information, thus they are agnostic of the input structures and are very general for being applied on any types of graphs, trees and even sentences.

This is a fundamental difference that leads to many superior properties of GNNs comparing with RNNs and DAG networks.
First, GNNs update node states in parallel within an iteration, thus they can be much faster than RNNs.
We will give more discussions and analysis in Chapters \ref{chap:nary} and \ref{chap:amrgen}.
Second, sibling nodes can easily incorporate the information of each other with GNNs. 
This can be achieved by simply executing a GNN for 2 iterations, so that the information of one sibling can go up then down to reach the other.

\subsection{Different types of GNNs}

So far there have been several types of GNNs, and their main differences lay in the way for updating node states from aggregated messages (Equation \ref{eq:2_apply}). 
We list several existing approaches in Table \ref{tab:2_gnns} and give detailed introduction below:

\subparagraph{Convolution} 
The first type is named graph convolutional network (GCN) \citep{kipf2017semi}.
It borrows the idea of convolutional neural network (CNN) \citep{krizhevsky2012imagenet}, which gathers larger contextual information through each convolution operation.
To deal with the problem where a graph node can have arbitrary number of neighbors, GCN and its later variants first sum up the hidden states of all neighbors, before applying the result of summation as messages to update the graph node state:
\begin{equation} \label{eq:2_aggre_sum}
    m^t_i = \sum_{v_j \in \boldsymbol{N_i}} \boldsymbol{h}^{t-1}_j
\end{equation}
Note that some GCN variations try to distinguish different types of neighbors before the summation:
\begin{equation}
    m^t_i = \sum_{v_j \in \boldsymbol{N_i}} \boldsymbol{W}_{L(i,j)} \boldsymbol{h}^{t-1}_j
\end{equation}
But they actually choose $\boldsymbol{W}_{L(i,j)}$s to be identical, and thus this is equivalent to Equation \ref{eq:2_aggre_sum}.
The underlying reason is that making $\boldsymbol{W}_{L(i,j)}$s to be different will introduce a lot of parameters, especially when the number of neighbor types is large.
The summation operation can be considered as the message aggregation process first mentioned in Equation \ref{eq:2_aggre}.
In fact, most existing GNNs use sum to aggregate message.
There are also other ways, which I will introduce later in this chapter.

After messages are calculated, they are applied to update graph node states.
GCNs use the simple linear transformation with ReLU \citep{nair2010rectified} as the activation function.

\subparagraph{Attention}
Another type of GNNs are called graph attention network (GAN) \citep{velivckovic2017graph}.
In general, it adopts multi-head self attention \citep{NIPS2017_7181} to calculate messages:
\begin{align}
    \boldsymbol{m}^{t}_i &= \sigma\left( \sum_{j \in \boldsymbol{N}_i} a_{i,j} \boldsymbol{h}^{t-1}_j\right) \\
    a_{i,j} &= \frac{\exp(e_{i,j})}{ \sum_{j' \in \boldsymbol{N}_i} \exp(e_{i,j'})} \\
    e_{i,j} &= MHA_t(\boldsymbol{h}^{t-1}_i, \boldsymbol{h}^{t-1}_j)
\end{align}
where $MHA_t$ corresponds to the $t$-th multi-head self attention layer.
For the next step, GAN diretly use the newly calculated message $\boldsymbol{m}^{t}_i$ as the new node state: $\boldsymbol{h}^{t}_i = \boldsymbol{m}^{t}_i$.

\begin{table}
    \centering
    \begin{tabular}{lll}
    \toprule
        Type & Model & Ways for applying messages \\
    \midrule
        Convolution & GCN & $\boldsymbol{h}^{t}_i = \text{ReLU}(\boldsymbol{W}^{t} \boldsymbol{m}^{t}_i + \boldsymbol{b}^{t}$)\\
    \midrule
        Attention & GAN & $\boldsymbol{h}^{t}_i = \boldsymbol{m}^{t}_i$ \\
    \midrule
        \multirow{2}{*}{Gated} & GGNN & $\boldsymbol{h}^{t}_i = \text{GRU}(\boldsymbol{m}^{t}_i, \boldsymbol{h}^{t-1}_i)$ \\
                               & GRN & $\boldsymbol{h}^{t}_i, \boldsymbol{c}^{t}_i = \text{LSTM}(\boldsymbol{m}^{t}_i, [\boldsymbol{h}^{t-1}_i, \boldsymbol{c}^{t-1}_i])$ \\
    \bottomrule
    \end{tabular}
    \caption{Comparison of several kinds of graph neural networks, where $\boldsymbol{m}$s and $\boldsymbol{h}$s have the same meaning as Equation \ref{eq:2_apply}. $\boldsymbol{W}$s and $\boldsymbol{b}$s represent model parameters.}
    \label{tab:2_gnns}
\end{table}

\subparagraph{Gated}
The massage propagation method based on linear transformations in GCN may suffer from long-range dependency problem, when dealing with very large and complex graphs.
To alleviate the long-range dependency problem, recent work has proposed to use gated mechanisms to process messages.
In particular, graph recurrent networks (GRN) \citep{P18-1030,P18-1150} leverage the gated operations of an LSTM \citep{hochreiter1997long} step to apply messages for node state updates.
On the other hand, gated graph neural networks (GGNN) \citep{li2015gated,P18-1026} adopt a GRU \citep{cho-EtAl:2014:EMNLP2014} step to conduct the update.
The message propagation mechanism for both models are shown in the last group of Table \ref{tab:2_gnns}. 
To generate messages, they also simply sum up the hidden states of all neighbors.
This is the same as GCNs.

\subparagraph{Discussion on message aggregation}
The models mentioned above either use summations or an attention mechanism to calculate messages.
As a result, these models have the same property: they are invariant to the permutations of their inputs, which result in different orders of neighbors.
This property are also called ``symmetric'', and \citet{hamilton2017inductive} introduce several other ``symmetric'' and ``asymmetric'' message aggregators.
In addition to summation and attention mechanisms, mean pooling and max pooling operations are also ``symmetric''.
This is intuitive, as both pooling operations are obviously invariant to input orders.

On the other hand, they mention an LSTM aggregator, which generates messages by simply applying an LSTM to a random permutation of a node’s neighbors.
The LSTM aggregator is not permutation invariant, and thus is ``asymmetric''.

\subsection{Discussion on the memory usage of GNNs}

So far we have discussed several advantages of GNNs.
Comparing with RNNs and DAG networks, GNNs are more flexible for handling any types of graphs.
Besides, they allow better parallelization, and thus are more efficient on GPUs.
However, they also suffer from limitations, and the most severe one is the large-scale memory usage.

As mentioned above, GNNs update every graph node state within an iteration, and all node states are updated for $T$ times if a GNN executes for $T$ message passing steps.
As a result, the increasing computation causes more memory usage.
In general, the amount of memory usage is highly related to the density and scale of the input graph.

To alleviate the memory issue, FastGCN \citep{chen2018fastgcn} adopts importance sampling to remove edges, making graphs less dense.
In contrast to fixed sampling methods above, \citet{huang2018adaptive} introduce
a parameterized and trainable sampler to perform layerwise sampling conditioned on the former layer. Furthermore, this adaptive sampler could find optimal sampling importance and reduce variance simultaneously.

%% file: 3-chap-MHQA.tex
In this chapter, we introduce a graph-based model for tackling multi-hop reading comprehension.
Multi-hop reading comprehension focuses on one type of factoid question, where a system needs to properly integrate multiple pieces of evidence to correctly answer a question. 
Previous work approximates global evidence with local coreference information, encoding coreference chains with DAG-styled GRU layers within a gated-attention reader.
However, coreference is limited in providing information for rich inference.
We introduce a new method for better connecting global evidence, which forms more complex graphs compared to DAGs.
To perform evidence integration on our graphs, we investigate our graph recurrent network (GRN).
Experiments on two standard datasets show that richer global information leads to better answers.
Our approach shows highly competitive performances on these datasets without deep language models (such as ELMo).

\section{Introduction}

Recent years have witnessed a growing interest in the task of machine reading comprehension.
Most existing work \citep{hermann2015teaching,wang2016machine,seo2016bidirectional,wang2016multi,weissenborn2017making,dhingra-EtAl:2017:Long2,shen2017reasonet,xiong2016dynamic} focuses on a factoid scenario where the questions can be answered by simply considering very local information, such as one or two sentences.
For example, to correctly answer a question ``What causes precipitation to fall?'',
a QA system only needs to refer to one sentence in a passage: ``... In meteorology, precipitation is any product of the condensation of atmospheric water vapor that falls under gravity. ...'', and the final answer ``gravity'' is indicated key words of ``precipitation'' and ``falls''.

A more challenging yet practical extension is multi-hop reading comprehension (MHRC) \citep{welbl2018constructing}, where a system needs to properly integrate multiple pieces of evidence to correctly answer a question.
Figure \ref{fig:3_example} shows an example, which contains three associated passages, a question and several candidate choices. 
In order to correctly answer the question, a system has to integrate the facts ``The Hanging Gardens are in Mumbai'' and ``Mumbai is a city in India''.
There are also some irrelevant facts, such as ``The Hanging Gardens provide sunset views over the Arabian Sea'' and ``The Arabian Sea is bounded by Pakistan and Iran'', which make the task more challenging, as an MHRC model has to distinguish the relevant facts from the irrelevant ones.

\begin{figure}
\begin{tabularx}{\textwidth}{|X|}
\hline
[\textbf{The Hanging Gardens}], in [\textbf{Mumbai}], also known as Pherozeshah Mehta Gardens, are terraced gardens ... [\textbf{They}] provide sunset views over the [\textbf{Arabian Sea}] ... \\
\hline
\hline
[\textbf{Mumbai}] (also known as Bombay, the official name until 1995) is the capital city of the Indian state of Maharashtra. [\textbf{It}] is the most populous city in [\textbf{India}] ... \\
\hline
\hline
The [\textbf{Arabian Sea}] is a region of the northern Indian Ocean bounded 
on the north by [\textbf{Pakistan}] and [\textbf{Iran}], on the west by northeastern [\textbf{Somalia}] and the Arabian Peninsula, and on the east by ... \\
\hline
\hline
\textbf{Q}: (The Hanging gardens, country, ?)  \\
\textbf{Candidate answers}:  {Iran, India, Pakistan, Somalia, ...} \\
\hline
\end{tabularx}
\caption{An example from WikiHop \citep{welbl2018constructing}, where some relevant entity mentions and their coreferences are highlighted.}
\label{fig:3_example}
\end{figure}

\begin{figure}
\centering
\includegraphics[width=0.8\linewidth]{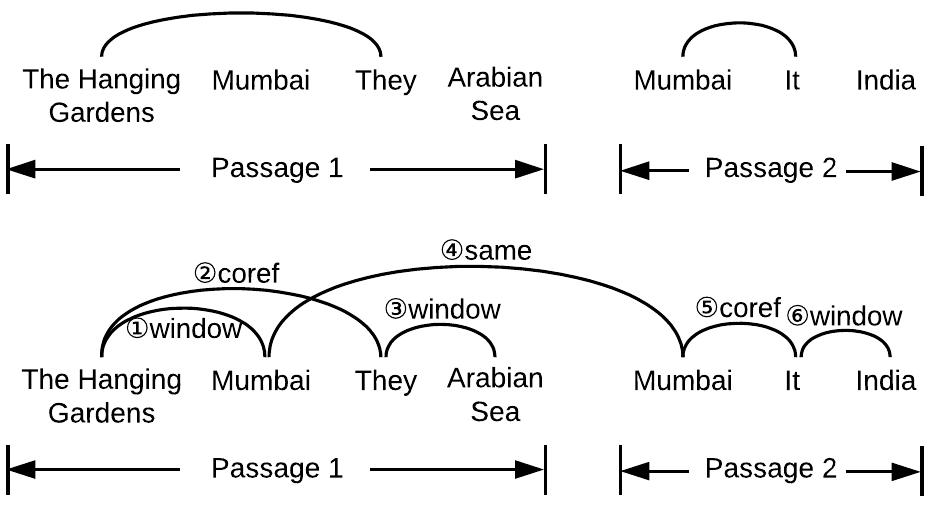}
\caption{A DAG generated by \citet{N18-2007} (top) and a graph by considering all three types of edges (bottom) on the example in Figure \ref{fig:3_example}.}
\label{fig:3_coref_vs_graph}
\end{figure}

Despite being a practical task, so far MHRC has received little research attention.
One notable method, Coref-GRU \citep{N18-2007}, uses coreference information to gather richer context for each candidate.
However, one main disadvantage of Coref-GRU is that the coreferences it considers are usually local to a sentence, neglecting other useful global information.
In addition, the resulting DAGs are usually very sparse, thus few new facts can be inferred.
The top part of Figure \ref{fig:3_coref_vs_graph} shows a directed acyclic graph (DAG) with only coreference edges. 
In particular, the two coreference edges infer two facts: ``The Hanging Gardens provide views over the Arabian Sea'' and ``Mumbai is a city in India'', from which we cannot infer the ultimate fact, ``The Hanging Gardens are in India'', for correctly answering this instance.

We propose a general graph scheme for evidence integration, which allows information exchange beyond co-reference nodes, by allowing arbitrary degrees of the connectivity of the reference graphs.
In general, we want the resulting graphs to be more densely connected so that more useful facts can be inferred.
For example each edge can connect two related entity mentions, while unrelated mentions, such as ``the Arabian Sea'' and ``India'', may not be connected.
In this paper, we consider three types of relations 
as shown in the bottom part of Figure \ref{fig:3_coref_vs_graph}.

The first type of edges connect the mentions of the \emph{same} entity appearing across passages or further apart in the same passage.
Shown in Figure \ref{fig:3_coref_vs_graph}, one instance connects the two ``Mumbai'' across the two passages.
Intuitively, \emph{same}-typed edges help to integrate global evidence related to the same entity, which are not covered by pronouns.
The second type of edges connect two mentions of different entities within a context \emph{window}.
They help to pass useful evidence further across entities.
For example, in the bottom graph of Figure \ref{fig:3_coref_vs_graph}, both \emph{window}-typed edges of \textcircled{1} and \textcircled{6} help to pass evidence from ``The Hanging Gardens'' to ``India'', the answer of this instance.
Besides, \emph{window}-typed edges enhance the relations between local mentions that can be missed by the sequential encoding baseline.
Finally, \emph{coreference}-typed edges are further complementary to the previous two types, and thus we also include them.

Our generated graphs are complex and can have cycles, making it difficult to directly apply a DAG network (e.g. the structure of Coref-GRU). 
So we adopt graph recurrent network (GRN), as it has been shown successful on encoding various types of graphs, including semantic graphs \citep{P18-1150}, dependency graphs \citep{song2018nary} and even chain graphs created by raw texts \citep{P18-1030}.

Given an instance containing several passages and a list of candidates, we first use NER and coreference resolution tools to obtain entity mentions, and then create a graph out of the mentions and relevant pronouns.
As the next step, evidence integration is executed on the graph by adopting a graph neural network on top of a sequential layer.
The sequential layer learns local representation for each mention, while the graph network learns a global representation.
The answer is decided by matching the representations of the mentions against the question representation.

Experiments on WikiHop \citep{welbl2018constructing} show that our created graphs are highly useful for MHRC.
On the hold-out testset, it achieves an accuracy of 65.4\%, which is highly competitive on the leaderboard\footnote{http://qangaroo.cs.ucl.ac.uk/leaderboard.html} as of the paper submission time.
In addition, our experiments show that the questions and answers are dramatically better connected on our graphs than on the coreference DAGs, if we map the questions on graphs using the question subject.
Our experiments also show a positive relation between graph connectivity and end-to-end accuracy.

On the testset of ComplexWebQuestions \citep{N18-1059}, our method also achieves better results than all published numbers.
To our knowledge, we are among the first to investigate graph neural networks on reading comprehension.

\section{Baseline}

\begin{figure}
\centering
\includegraphics[width=0.85\linewidth]{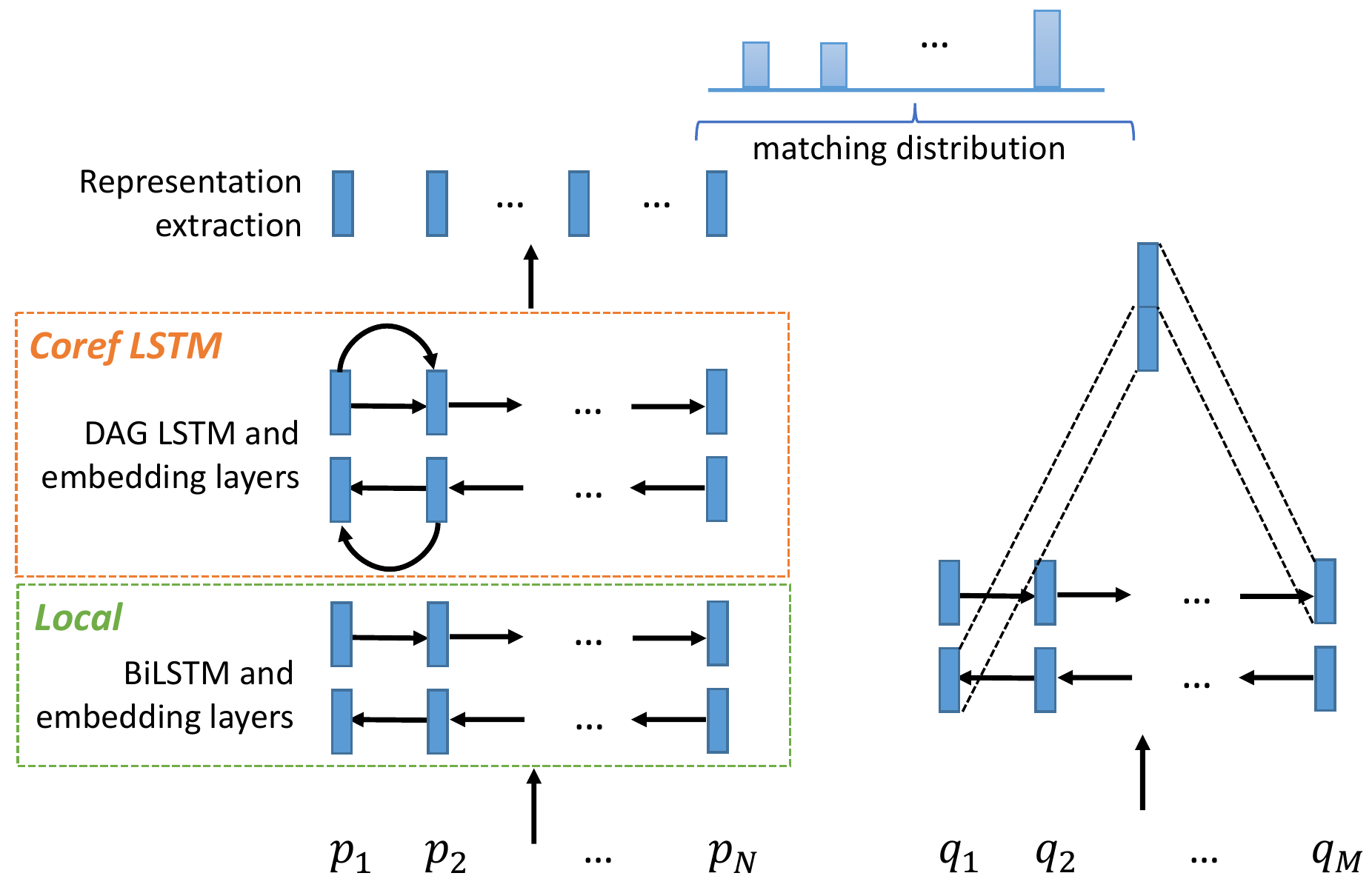}
\caption{Baselines. The upper dotted box is a DAG LSTM layer with addition coreference links, while the bottom one is a typical BiLSTM layer. Either layer is used.}
\label{fig:3_baseline}
\end{figure}

As shown in Figure \ref{fig:3_baseline}, we introduce two baselines, which are inspired by \citet{N18-2007}.
The first baseline, \emph{Local}, uses a standard BiLSTM layer (shown in the green dotted box), where inputs are first encoded with a BiLSTM layer, and then the representation vectors for the mentions in the passages are extracted, before being matched against the question for selecting an answer.
The second baseline, \emph{Coref LSTM}, differs from \emph{Local} by replacing the BiLSTM layer with a DAG LSTM layer (shown in the orange dotted box) for encoding additional coreference information, as proposed by \citet{N18-2007}.


\subsection{\emph{Local}: BiLSTM encoding}

Given a list of relevant passages, we first concatenate them into one large passage $p_1, p_2 \dots p_N$, where each $p_i$ is a passage word and $\boldsymbol{x}_{p_i}$ is the embedding of it.
The \emph{Local} baseline adopts a Bi-LSTM to encode the passage:
\begin{align}
\overleftarrow{\boldsymbol{h}}_p^i &= \textrm{LSTM}(\overleftarrow{\boldsymbol{h}}_p^{i+1}, \boldsymbol{x}_{p_i}) \\
\overrightarrow{\boldsymbol{h}}_p^i &= \textrm{LSTM}(\overrightarrow{\boldsymbol{h}}_p^{i-1}, \boldsymbol{x}_{p_i})
\end{align}
Each hidden state contains the information of its local context.
Similarly, the question words $q_1, q_2 \dots q_M$ are first converted into embeddings $\boldsymbol{x}_{q_1}, \boldsymbol{x}_{q_2} \dots \boldsymbol{x}_{q_M}$ before being encoded by another BiLSTM:
\begin{align}
\overleftarrow{\boldsymbol{h}}_q^j &= \textrm{LSTM}(\overleftarrow{\boldsymbol{h}}_q^{j+1}, \boldsymbol{x}_{q_j}) \\
\overrightarrow{\boldsymbol{h}}_q^j &= \textrm{LSTM}(\overrightarrow{\boldsymbol{h}}_q^{j-1}, \boldsymbol{x}_{q_j})
\end{align}

\subsection{\emph{Coref LSTM}: DAG LSTM encoding with conference}
\label{sec:3_coref_lstm}

Taking the passage word embeddings $\boldsymbol{x}_{p_1}, \dots \boldsymbol{x}_{p_N}$ and coreference information as the input, 
the DAG LSTM layer encodes each input word embedding (such as $\boldsymbol{x}_{p_i}$) with the following gated operations\footnote{Only the forward direction is shown for space consideration}:
\begin{equation}
\begin{split}
&\boldsymbol{i}_i = \sigma(\boldsymbol{W}_i \boldsymbol{x}_{p_i} + \boldsymbol{U}_i \sum_{i'\in \boldsymbol{\Omega}(i)}\overrightarrow{\boldsymbol{h}}_p^{i'} + \boldsymbol{b}_i) \\
&\boldsymbol{o}_i = \sigma(\boldsymbol{W}_o \boldsymbol{x}_{p_i} + \boldsymbol{U}_o \sum_{i'\in \boldsymbol{\Omega}(i)}\overrightarrow{\boldsymbol{h}}_p^{i'} + \boldsymbol{b}_o) \\
&\boldsymbol{f}_{i',i} = \sigma(\boldsymbol{W}_f \boldsymbol{x}_{p_i} + \boldsymbol{U}_f \overrightarrow{\boldsymbol{h}}_p^{i'} + \boldsymbol{b}_f) \\
&\boldsymbol{u}_i = \sigma(\boldsymbol{W}_u \boldsymbol{x}_{p_i} + \boldsymbol{U}_u \sum_{i'\in \boldsymbol{\Omega}(i)}\overrightarrow{\boldsymbol{h}}_p^{i'} + \boldsymbol{b}_u) \\
&\overrightarrow{\boldsymbol{c}}_p^i = \boldsymbol{i}_i \odot \boldsymbol{u}_i + \sum_{i'\in \boldsymbol{\Omega}(i)} \boldsymbol{f}_{i',i} \odot \overrightarrow{\boldsymbol{c}}_p^{i'} \\
&\overrightarrow{\boldsymbol{h}}_p^i = \boldsymbol{o}_i \odot \tanh (\overrightarrow{\boldsymbol{c}}_p^i) \\
\end{split}
\end{equation}
$\boldsymbol{\Omega}(i)$ represents all preceding words of $p_i$ in the DAG, $\boldsymbol{i}_i$, $\boldsymbol{o}_i$ and $\boldsymbol{f}_{i',i}$ are the input, output and forget gates, respectively. $\boldsymbol{W}_x$, $\boldsymbol{U}_x$ and $\boldsymbol{b}_x$ ($x \in \{i,o,f,u\}$) are model parameters.

\subsection{Representation extraction}

After encoding both the passage and the question, we obtain a representation vector for each entity mention $\epsilon_k \in \boldsymbol{E}$ ($\boldsymbol{E}$ represents all entities), spanning from $k_i$ to $k_j$, by concatenating the hidden states of its start and end positions, before they are correlated with a fully connected layer:
\begin{equation} \label{eq:3_base_rep}
\boldsymbol{h}_\epsilon^k = \boldsymbol{W}_1 [\overleftarrow{\boldsymbol{h}}_p^{k_i};\overrightarrow{\boldsymbol{h}}_p^{k_i};\overleftarrow{\boldsymbol{h}}_p^{k_j};\overrightarrow{\boldsymbol{h}}_p^{k_j}]+\boldsymbol{b}_1 \textrm{,}
\end{equation}
where $\boldsymbol{W}_1$ and $\boldsymbol{b}_1$ are model parameters for compressing the concatenated vector.
Note that the current multi-hop reading comprehension datasets all focus on the situation where the answer is a named entity.
Similarly, the representation vector for the question is generated by concatenating the hidden states of its first and last positions:
\begin{equation}
\boldsymbol{h}_q = \boldsymbol{W}_2 [\overleftarrow{\boldsymbol{h}}_q^1;\overrightarrow{\boldsymbol{h}}_q^1;\overleftarrow{\boldsymbol{h}}_q^M;\overrightarrow{\boldsymbol{h}}_q^M]+\boldsymbol{b}_2
\end{equation}
where $\boldsymbol{W}_2$ and $\boldsymbol{b}_2$ are also model parameters.

\subsection{Attention-based matching}

Given the representation vectors for the question and the entity mentions in the passages, an additive attention model \citep{bahdanau2015neural}\footnote{We adopt a standard matching method, as our focus is the effectiveness of evidence integration. We leave investigating other approaches \citep{luong-pham-manning:2015:EMNLP,Wang:2017:BMM:3171837.3171865} as future work.}
is adopted by treating all entity mention representations and the question representation as the memory and the query, respectively.
In particular, the probability for a candidate $c$ being the answer given input $X$ is calculated by summing up all the occurrences of $c$ across the input passages:\footnote{All candidates form a subset of all entities ($\boldsymbol{E}$).}
\begin{equation} \label{eq:3_merge}
p(c|X) = \frac{\sum_{k\in \mathcal{N}_c} \alpha_k}{\sum_{k'\in \mathcal{N}} \alpha_{k'}} \textrm{,}
\end{equation}
where $\mathcal{N}_c$ and $\mathcal{N}$ represent all occurrences of the candidate $c$ and all occurrences of all candidates, respectively. 
Previous work \citep{wang2018r3} shows that summing the probabilities over all occurrences of the same entity mention is important for the multi-passage scenario.
$\alpha_k$ is the attention score for the entity mention $\epsilon_k$, calculated by an additive attention model shown below:
\begin{align}
e_0^k &= \boldsymbol{v}_{\alpha}^T \tanh(\boldsymbol{W}_{\alpha} \boldsymbol{h}_\epsilon^k + \boldsymbol{U}_{\alpha} \boldsymbol{h}_q + \boldsymbol{b}_{\alpha}) \\
\alpha_k &= \frac{\exp(e_0^k)}{\sum_{k'\in \mathcal{N}} \exp(e_0^{k'})} \label{eq:softmax}
\end{align}
where $\boldsymbol{v}_{\alpha}$, $\boldsymbol{W}_{\alpha}$, $\boldsymbol{U}_{\alpha}$ and $\boldsymbol{b}_{\alpha}$ are model parameters.

\paragraph{Comparison with \citet{N18-2007}}

The Coref-GRU model \citep{N18-2007} is based on the gated-attention reader (GA reader) \citep{dhingra-EtAl:2017:Long2}.
GA reader is designed for the cloze-style reading comprehension task \citep{hermann2015teaching}, where \emph{one} token is selected from the input passages as the answer for each instance.
To adapt their model for the WikiHop benchmark, where an answer candidate can contain multiple tokens, they first generate a probability distribution over the passage tokens with GA reader, and then compute the probability for each candidate $c$ by aggregating the probabilities of all passage tokens that appear in $c$ and renormalizing over the candidates.

In addition to using LSTM instead of GRU\footnote{Model architectures are selected according to dev results.}, the main difference between our two baselines and \citet{N18-2007} is that our baselines consider each candidate as a whole unit no matter whether it contains multiple tokens or not. 
This makes our models more effective on the datasets containing phrasal answer candidates.

\begin{figure}
\centering
\includegraphics[width=0.9\linewidth]{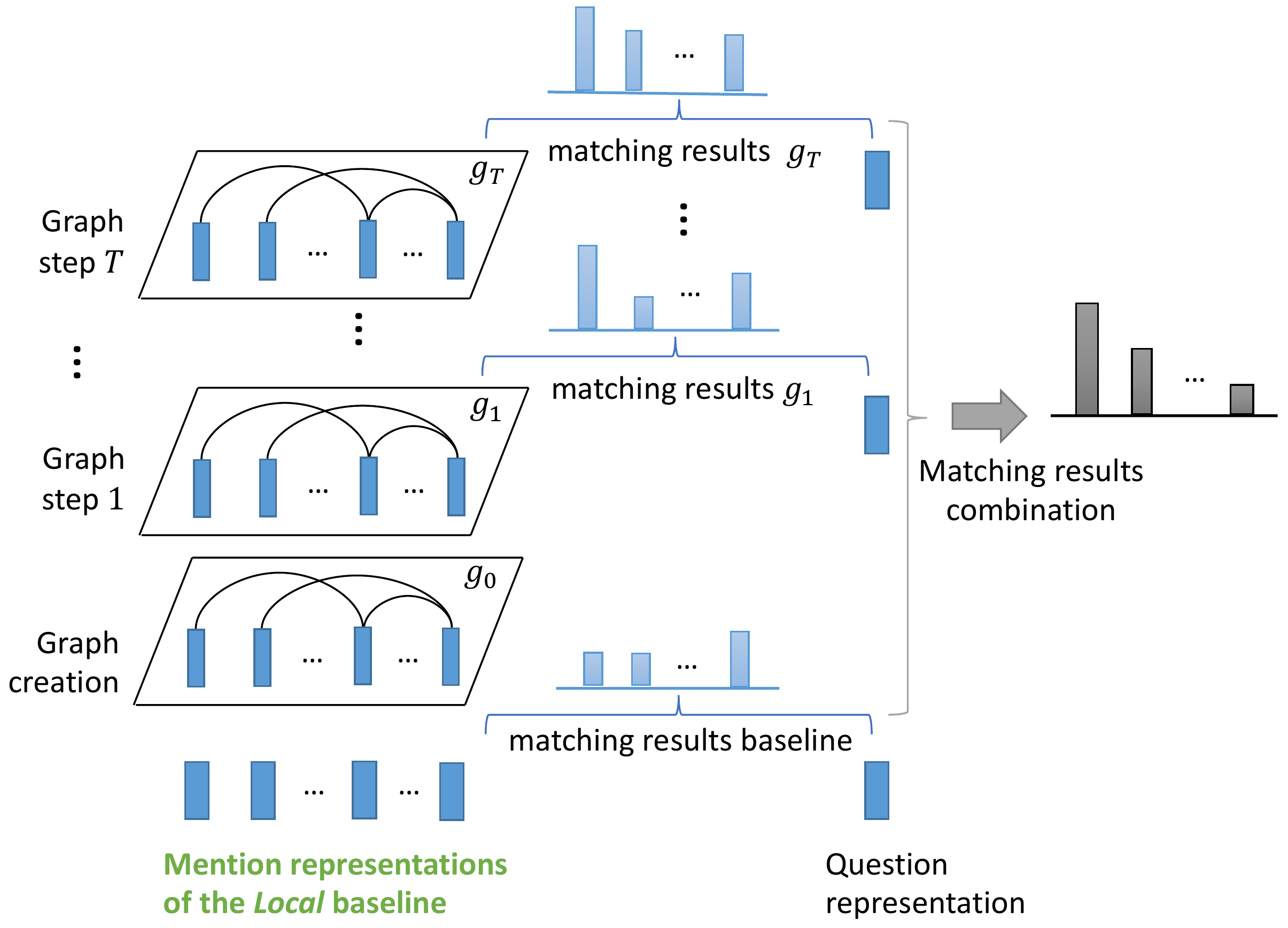}
\caption{Model framework.}
\label{fig:3_model_overview}
\end{figure}

\section{Evidence integration with GRN encoding}

Over the representation vectors for a question and the corresponding entity mentions, we build an evidence integration graph of the entity mentions by connecting relevant mentions with edges, and then integrating relevant information for each graph node (entity mention) with a graph recurrent network (GRN) \citep{P18-1030,P18-1150}.
Figure \ref{fig:3_model_overview} shows the overall procedure of our approach.

\subsection{Evidence graph construction}
\label{sec:3_graph_cons}

As a first step, we create an evidence graph from a list of input passages to represent interrelations among entities within the passages.
The entity mentions within the passages are taken as the graph nodes. 
They are automatically generated by NER and coreference annotators, so that each graph node is either an entity mention or a pronoun representing an entity.
We then create a graph by ensuring that edges between two nodes follow the situations below:
\begin{itemize}
\item They are occurrences of the \textbf{same} entity mention across passages or with a distance larger than a threshold $\tau_L$ when being in the same passage.
\item One is an entity mention and the other is its \textbf{coreference}. Our coreference information is automatically generated by a coreference annotator.
\item Between two mentions of different entities in the same passage within a \textbf{window} threshold of $\tau_S$.
\end{itemize}
Between every two entities that satisfy the situations above, we make two edges in opposite directions. 
As a result, each generated graph can also be considered as an undirected graph.

\subsection{Evidence integration with graph encoding}

Tackling multi-hop reading comprehension requires inferring on global context.
As the next step, we merge related information through the three types of edges just created by applying GRN on our graphs.

\begin{figure}
    \centering
    \includegraphics[width=0.5\textwidth]{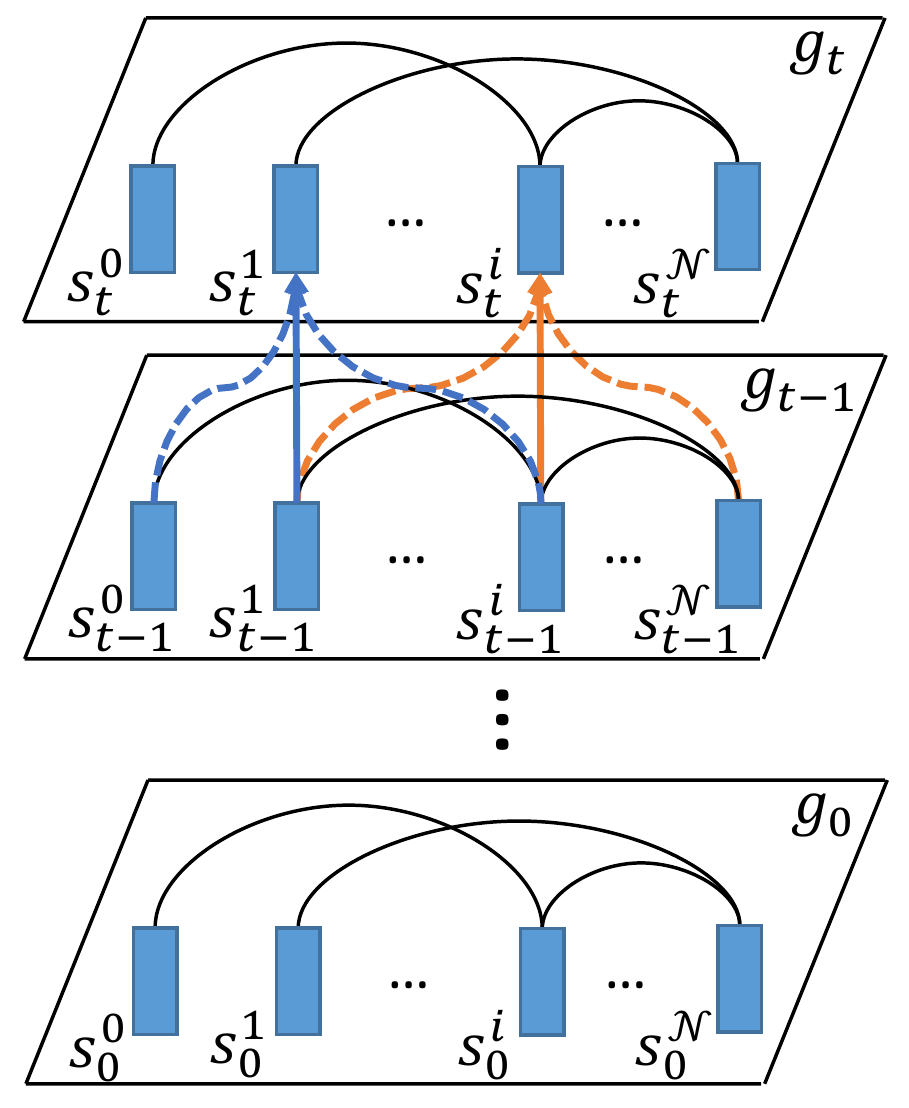}
    \caption{Graph recurrent network for evidence graph encoding.}
    \label{fig:3_grn}
\end{figure}

Figure \ref{fig:3_grn} shows the overall structure of our graph encoder.
Formally, given a graph $G=(V,E)$, a hidden state vector $\boldsymbol{s}^k$ is created to represent each entity mention $\epsilon_k \in V$.
The state of the graph can thus be represented as:
\begin{equation}
\boldsymbol{g}=\{\boldsymbol{s}^k\}|\epsilon_k \in V
\end{equation}
In order to integrate non-local evidence among nodes, information exchange between neighborhooding nodes is performed through recurrent state transitions, leading to a sequence of graph states $\boldsymbol{g}_0, \boldsymbol{g}_1, \dots, \boldsymbol{g}_T$, where $\boldsymbol{g}_T=\{\boldsymbol{s}^k_T\}|\epsilon_k \in V$ and $T$ is a hyperparameter representing the number of graph state transition decided by a development experiment.
For initial state $\boldsymbol{g}_0=\{\boldsymbol{s}^k_0\}|\epsilon_k \in V$, we initialize each $\boldsymbol{s}^k_0$ by:
\begin{equation} \label{eq:3_init}
\boldsymbol{s}^k_0 = \boldsymbol{W}_3 [\boldsymbol{h}_\epsilon^k; \boldsymbol{h}_q] + \boldsymbol{b}_3 \textrm{,}
\end{equation}
where $\boldsymbol{h}_\epsilon^k$ is the corresponding representation vector of entity mention $\epsilon_k$, calculated by Equation \ref{eq:3_base_rep}. $\boldsymbol{h}_q$ is the question representation. $\boldsymbol{W}_3$ and $\boldsymbol{b}_3$ are model parameters.

\subparagraph{State transition}
A gated recurrent neural network is used to model the state transition process. 
In particular, the transition from $\boldsymbol{g}_{t-1}$ to $\boldsymbol{g}_t$ consists of a hidden state transition for each node, as shown in Figure \ref{fig:3_grn}.
At each step $t$, direct information exchange is conducted between a node and all its neighbors.
To avoid gradient diminishing or bursting, LSTM \citep{hochreiter1997long} is adopted, where a cell vector $\boldsymbol{c}^k_t$ is taken to record memory for hidden state $\boldsymbol{s}^k_t$:
\begin{equation} \label{eq:3_grn_mp}
\begin{split}
\boldsymbol{i}_t^k &= \sigma (\boldsymbol{W}_i \boldsymbol{m}_{t}^k + \boldsymbol{b}_i) \\
\boldsymbol{o}_t^k &= \sigma (\boldsymbol{W}_o \boldsymbol{m}_{t}^k + \boldsymbol{b}_o) \\
\boldsymbol{f}_t^k &= \sigma (\boldsymbol{W}_f \boldsymbol{m}_{t}^k + \boldsymbol{b}_f) \\
\boldsymbol{u}_t^k &= \sigma (\boldsymbol{W}_u \boldsymbol{m}_{t}^k + \boldsymbol{b}_u) \\
\boldsymbol{c}_t^k &= \boldsymbol{f}_t^k \odot \boldsymbol{c}_{t-1}^k + \boldsymbol{i}_t^k \odot \boldsymbol{u}_t^k \\
\boldsymbol{s}_t^k &= \boldsymbol{o}_t^k \odot \tanh(\boldsymbol{c}_t^k) \textrm{,}
\end{split}
\end{equation}
where $\boldsymbol{c}_t^k$ is the cell vector to record memory for $\boldsymbol{s}_t^k$, and $\boldsymbol{i}_t^k$, $\boldsymbol{o}_t^k$ and $\boldsymbol{f}_t^k$ are the input, output and forget gates, respectively. 
$\boldsymbol{W}_x$ and $\boldsymbol{b}_x$ ($x\in \{i,o,f,u\}$) are model parameters.
In the remaining of this thesis, I will use the symbol \emph{LSTM} to represent Equation \ref{eq:3_grn_mp}.
$\boldsymbol{m}_{t}^k$ is the sum of the neighborhood hidden states for the node $\epsilon_k$\footnote{We tried distinguishing neighbors by different types of edges, but it does not improve the performance.}:
\begin{equation} \label{eq:3_aggre}
\boldsymbol{m}_{t}^k = \sum_{i \in \boldsymbol{\Omega}(k)} \boldsymbol{s}_{t-1}^i
\end{equation}
$\boldsymbol{\Omega}(k)$ represents the set of all neighbors of $\epsilon_k$.

\subparagraph{Message passing}
To describe GRN using the message parsing framework shown in Equations \ref{eq:2_aggre} and \ref{eq:2_apply} (Section \ref{sec:2_gnn_comp}), 
$\boldsymbol{m}_t^k$ represents the message for node $\epsilon_k$ at step $t$.
It is first aggregated by summing up the hidden states of all neighbors of $\epsilon_k$, before being applied to update $\boldsymbol{s}_t^k$ with an LSTM step.

\subparagraph{Recurrent steps}
Using the above state transition mechanism, information from each node propagates to all its neighboring nodes after each step.
Therefore, for the worst case where the input graph is a chain of nodes, the maximum number of steps necessary for information from one arbitrary node to reach another is equal to the
size of the graph.
We experiment with different transition steps to study the effectiveness of global encoding.

Note that unlike the sequence LSTM encoder, our graph encoder allows parallelization in node-state updates, and thus can be highly efficient using a GPU. 
It is general and can be potentially applied to other tasks, including sequences, syntactic trees and cyclic structures.

\subsection{Matching and combination}

After evidence integration, we match the hidden states at each graph encoding step with the question representation using the same additive attention mechanism introduced in the Baseline section.
In particular, for each entity $\epsilon_k$, the matching results for the baseline and each graph encoding step $t$ are first generated, before being combined using a weighted sum to obtain the overall matching result:
\begin{align}
e_t^k &= \boldsymbol{v}_{a_t}^T \tanh(\boldsymbol{s}_t^k \boldsymbol{W}_{a_t} + \boldsymbol{h}_q \boldsymbol{U}_{a_t} + \boldsymbol{b}_{a_t}) \\
e^k &= \boldsymbol{w}_c \odot [e_0^k, e_1^k, \dots, e_T^k] + b_c \textrm{,}
\end{align}
where $e_0^k$ is the baseline matching result for $\epsilon_k$, $e_t^k$ is the matching results after $t$ steps, and $T$ is the total number of graph encoding steps. 
$\boldsymbol{W}_{a_t}$, $\boldsymbol{U}_{a_t}$, $\boldsymbol{v}_{a_t}$, $\boldsymbol{b}_{a_t}$, $\boldsymbol{w}_c$ and $b_c$ are model parameters.
In addition, a probability distribution is calculated from the overall matching results using softmax, similar to Equations \ref{eq:softmax}. 
Finally, probabilities that belong to the same entity mention are merged to obtain the final distribution, as shown in Equation \ref{eq:3_merge}.

\section{Training} 
We train both the baseline and our models using the cross-entropy loss:
\begin{equation}
l = -\log p(c^*|\boldsymbol{X};\boldsymbol{\theta}) \textrm{,}
\end{equation}
where $c^*$ is ground-truth answer, $\boldsymbol{X}$ and $\boldsymbol{\theta}$ are the input and model parameters, respectively.
Adam \citep{kingma2014adam} with a learning rate of 0.001 is used as the optimizer. 
Dropout with rate 0.1 and a $l$2 normalization weight of $10^{-8}$ are used during training.

\section{Experiments on WikiHop}

In this section, we study the effectiveness of rich types of edges and the graph encoders using the WikiHop \citep{welbl2018constructing} dataset.
It is designed for multi-evidence reasoning, as its construction process makes sure that multiple evidence are required for inducing the answer for each instance.

\subsection{Data}

The dataset contains around 51K instances, including 44K for training, 5K for development and 2.5K for held-out testing.
Each instance consists of a question, a list of associated passages, a list of candidate answers and a correct answer.
One example is shown in Figure \ref{fig:3_example}.
On average each instance has around 19 candidates, all of which are the same category.
For example, if the answer is a country, all other candidates are also countries. 
We use Stanford CoreNLP \citep{manning-EtAl:2014:P14-5} to obtain coreference and NER annotations. 
Then the entity mentions, pronoun coreferences and the provided candidates are taken as graph nodes to create an evidence graph.
The distance thresholds ($\tau_L$ and $\tau_S$, in Section \ref{sec:3_graph_cons}) for making \emph{same} and \emph{window} typed edges are set to 200 and 20, respectively.

\begin{figure}
\centering
\includegraphics[width=0.85\linewidth]{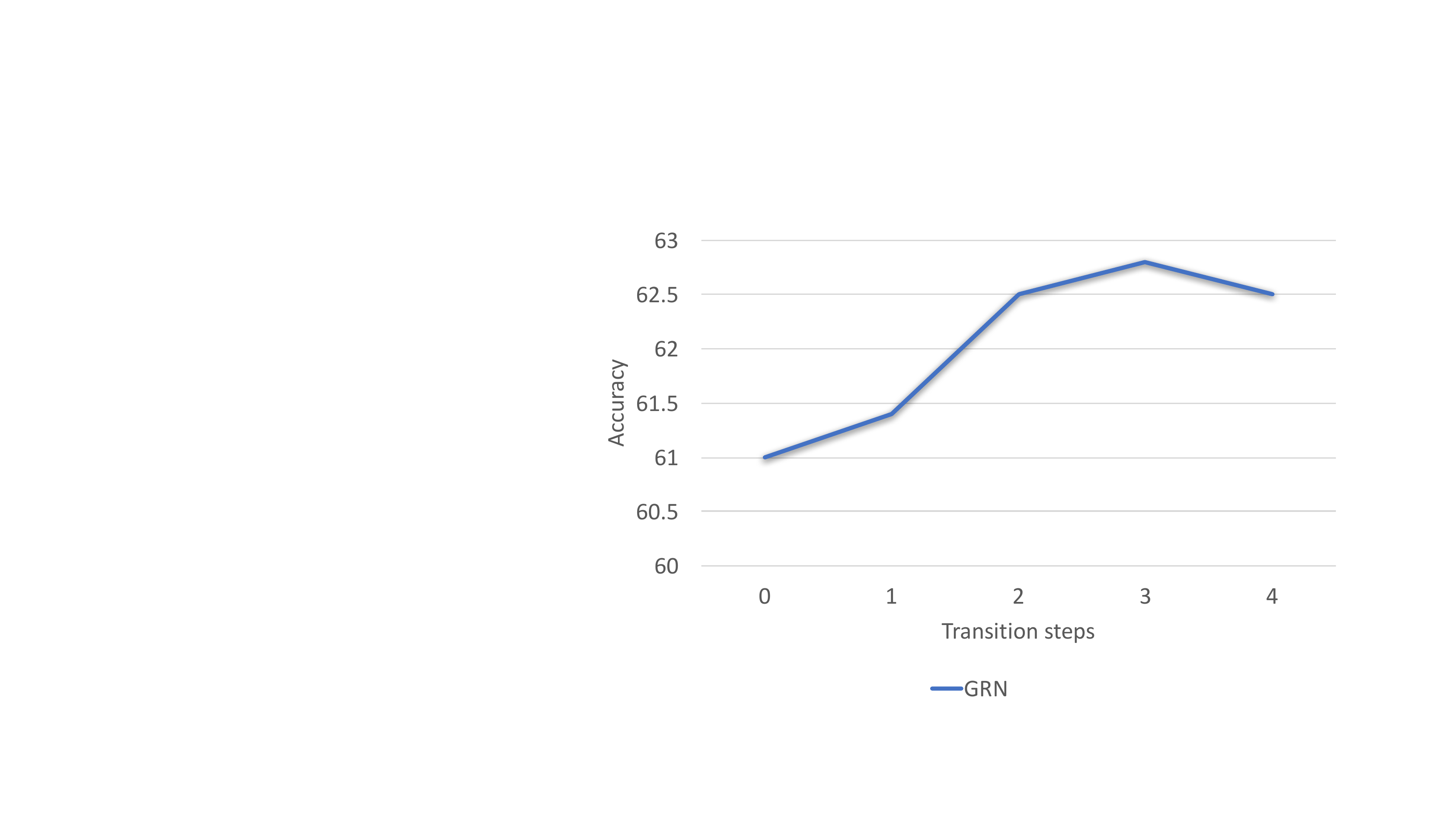}
\caption{\textsc{Dev} performances of different transition steps.}
\label{fig:3_dev_steps}
\end{figure}

\subsection{Settings}
We study the model behavior on the WikiHop devset, choosing the best hyperparameters for online system evaluation on the final holdout testset.
Our word embeddings are initialized from the 300-dimensional pretrained Glove word embeddings \citep{pennington2014glove} on Common Crawl, and are not updated during training.

For model hyperparameters, we set the graph state transition number as 3 according to development experiments.
Each node takes information from at most 200 neighbors, where \emph{same} and \emph{coref} typed neighbors are kept first. 
The hidden vector sizes for both bidirectional LSTM and GRN layers are set to 300.

\subsection{Development experiments}

Figure \ref{fig:3_dev_steps} shows the devset performance of our GRN-based model with different transition steps.
It shows the baseline performances when transition step is 0.
The accuracy goes up when increasing the transition step to 3.
Further increasing the transition step leads to a slight performance decrease.
One reason can be that executing more transition steps may also introduce more noise through richly connected edges.
We set the transition step to 3 for all remaining experiments.

\begin{table}
\centering
\begin{tabular}{l|c|c}
Model & Dev & Test \\
\hline
GA w/ GRU \citep{N18-2007} & 54.9 & -- \\
GA w/ Coref-GRU \citep{N18-2007} & 56.0 & 59.3 \\
\hline
Local & 61.0 & -- \\
Local-2L & 61.3 & -- \\
Coref LSTM & 61.4 & -- \\
Coref GRN & 61.4 & -- \\
Fully-Connect-GRN & 61.3 & -- \\
MHQA-GRN & \bf 62.8 & \bf 65.4 \\
\hline
\hline
Leaderboard 1st [anonymized]$\dagger$ & -- & 70.6 \\
Leaderboard 2nd [anonymized]$\dagger$ & -- & 67.6 \\
3rd, \citet{gcnwiki}$\dagger$ & -- & 67.6 \\
\end{tabular}
\caption{
Main results (unmasked) on WikiHop, where systems with $\dagger$ use ELMo \cite{peters2018deep}.}
\label{tab:3_wikihop}
\end{table}


\subsection{Main results}

Table \ref{tab:3_wikihop} shows the main comparison results\footnote{At paper-writing time, we observe a recent short arXiv paper \citep{gcnwiki} and two anonymous papers submitted to ICLR, showing better results with ELMo \citep{peters2018deep}. Our main contribution is studying an evidence integration approach, which is orthogonal to the contribution of ELMo on large-scale training. We will investigate ELMo in a future version.} with existing work.
\emph{GA w/ GRU} and \emph{GA w/ Coref-GRU} correspond to \citet{N18-2007}, and their reported numbers are copied.
The former is their baseline, a gated-attention reader \citep{dhingra-EtAl:2017:Long2}, and the latter is their proposed method.

For our baselines, \emph{Local} and \emph{Local-2L} encode passages with a BiLSTM and a 2-layer BiLSTM, respectively, both only capture local information for each mention.
We introduce \emph{Local-2L} for better comparison, as our models have more parameters than \emph{Local}.
\emph{Coref LSTM} is another baseline, encoding passages with coreference annotations by a DAG LSTM (Section \ref{sec:3_coref_lstm}).
This is a reimplementation of \citet{N18-2007} based on our framework.
\emph{Coref GRN} is another baseline that encodes coreferences with GRN.
It is for contrasting coreference DAGs with our evidence integration graphs.
\emph{MHQA-GRN} corresponds to our evidence integration approaches via graph encoding, adopting GRN for graph encoding.

\begin{table}
\centering
\begin{tabular}{l|c}
Edge type & Dev \\
\hline
all types & 62.8 \\
~~~~~~~~~~w/o same & 61.9 \\
~~~~~~~~~~w/o coref & 61.7 \\
~~~~~~~~~~w/o window & 62.4 \\
\hline
only same &  61.6 \\
only coref & 61.4 \\
only window & 61.1 \\
\end{tabular}
\caption{Ablation study on different types of edges using GRN as the graph encoder.}
\label{tab:3_dev_edges}
\end{table}

First, even our \emph{Local} show much higher accuracies compared with \emph{GA w/ GRU} and \emph{GA w/ Coref-GRU}. 
This is because our models are more compatible with the evaluated dataset.
In particular, \emph{GA w/ GRU} and \emph{GA w/ Coref-GRU} calculate the probability for each candidate by summing up the probabilities of all tokens within the candidate.
As a result, they cannot handle phrasal candidates very well, especially for the overlapping candidates, such as ``New York'' and ``New York City''.
On the other hand, we consider each candidate answer as a single unit, and does not suffer from this issue.
As a reimplementation of their idea, \emph{Coref LSTM} only shows 0.4 points gains over \emph{Local}, a stronger baseline than \emph{GA w/ GRU}.
On the other hand, \emph{MHQA-GRN} is 1.8 points more accurate than \emph{Local}.

The comparisons below help to further pinpoint the advantage of our approach:
\emph{MHQA-GRN} is 1.4 points better than \emph{Coref GRN} , while \emph{Coref GRN} gives a comparable performance with \emph{Coref LSTM}.
Both comparisons show that our evidence graphs are the main reason for achieving the 1.8-points improvement, and it is mainly because our evidence graphs are better connected than coreference DAGs and are more suitable for integrating relevant evidence.
\emph{Local-2L} is not significantly better than \emph{Local}, meaning that simply introducing more parameters does not help.

In addition to the systems above, we introduce \emph{Fully-Connect-GRN} for demonstrating the effectiveness of our evidence graph creating approach.
\emph{Fully-Connect-GRN} creates fully connected graphs out of the entity mentions, before encoding them with GRN.
Within each fully connected graph, the question is directly connected with the answer.
However, fully connected graphs are brute-force connections, and are not representative for integrating related evidence. 
\emph{MHQA-GRN} is 1.5 points better than \emph{Fully-Connect-GRN}, while questions and answers are more directly connected (with distance 1 for all cases) by \emph{Fully-Connect-GRN}.
The main reason can be that our evidence graphs only connect related entity mentions, making our models easier to learn how to integrate evidence.
On the other hand, there are barely learnable patterns within fully connected graphs.
More analyses on the relation between graph connectivity and end-to-end performance will be shown in later paragraphs.

We observe some unpublished papers showing better results with ELMo \citep{peters2018deep}, which is orthogonal to our contribution. 

\subsection{Analysis}

\subparagraph{Effectiveness of edge types}
Table \ref{tab:3_dev_edges} shows the ablation study of different types of edges that we introduce for evidence integration.
The first group shows the situations where one type of edges are removed.
In general, there is a large performance drop by removing any type of edges.
The reason can be that the connectivity of the resulting graphs is reduced, thus fewer facts can be inferred.
Among all these types, removing \emph{window}-typed edges causes the least performance drop.
One possible reason is that some information captured by them has been well captured by sequential encoding.
However, \emph{window}-typed edges are still useful, as they can help passing evidence through to further nodes.
Take Figure \ref{fig:3_coref_vs_graph} as an example, two \emph{window}-typed edges help to pass information from ``The Hanging Gardens'' to ``India''.
The other two types of edges are slightly more important than \emph{window}-typed edges.
Intuitively, they help to gather more global information than \emph{window}-typed edges, thus learn better representations for entities by integrating contexts from their occurrences and co-references.

The second group of Table \ref{tab:3_dev_edges} shows the model performances when only one type of edges are used.
None of the performances with single-typed edges are significantly better than the \emph{Local} baseline, whereas the combination of all types of edges achieves a much better accuracy (1.8 points) than \emph{Local}. 
This indicates the importance of evidence integration over better connected graphs.
We show more detailed quantitative analyses later on.
The numbers generally demonstrate the same patterns as the first group.
In addition, \emph{only same} is slightly better than \emph{only coref}.
It is likely because some coreference information can also be captured by sequential encoding.

\begin{figure}
\centering
\includegraphics[width=0.85\linewidth]{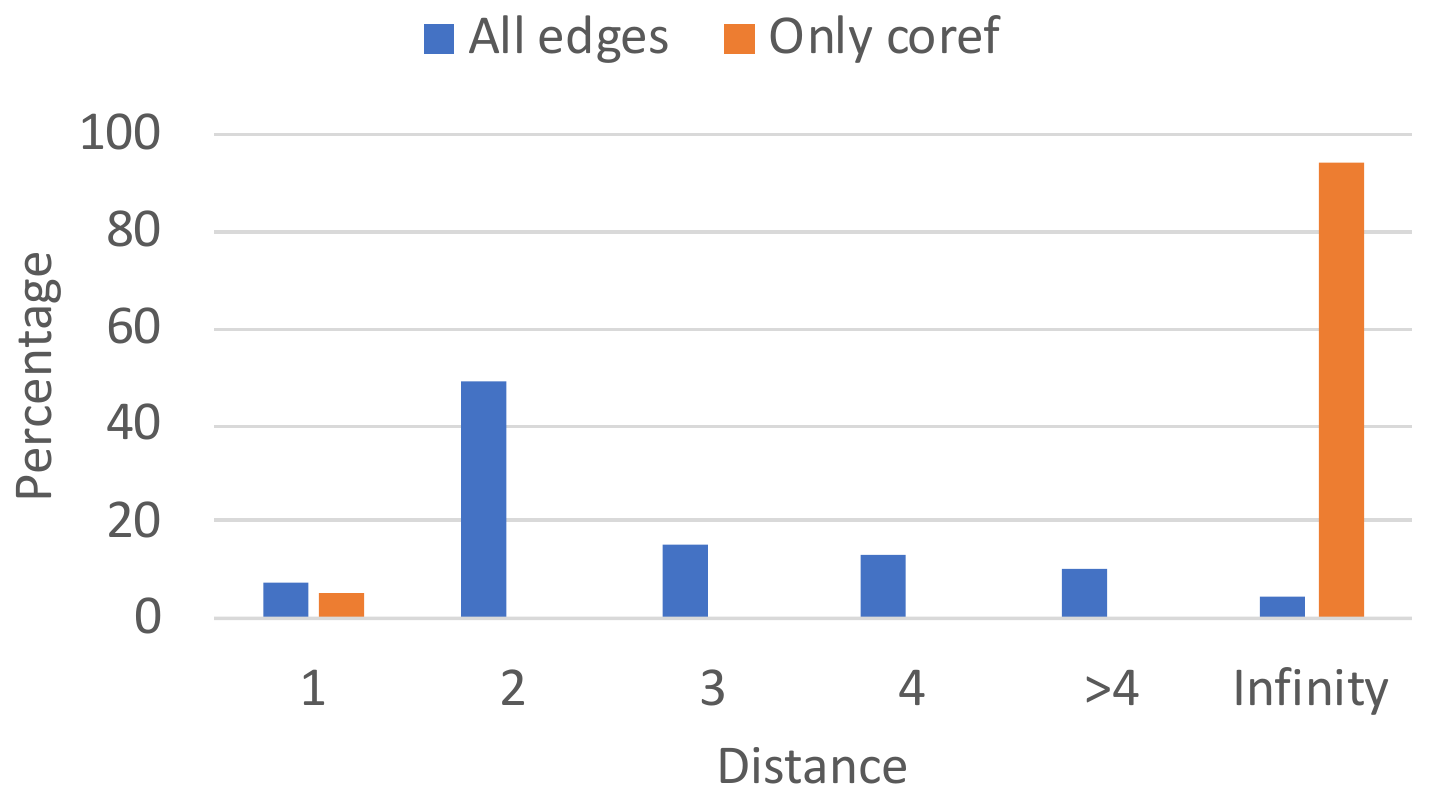}
\caption{Distribution of distances between a question and an answer on the \textsc{Devset}.}
\label{fig:3_dist}
\end{figure}

\vspace{0.5em}
\textbf{Distance}
Figure \ref{fig:3_dist} shows the percentage distribution of distances between a question and its closest answer when either all types of edges are adopted or only coreference edges are used.
The subject of each question\footnote{As shown in Figure \ref{fig:3_example}, each question has a subject, a relation and asks for the object.} is used to locate the question on the corresponding graph.

When all types of edges are adopted, the questions and the answers for more than 90\% of the development instances are connected, and the question-and-answer distances for more than 70\% are within 3.
On the other hand, the instances with distances longer than 4 only count for 10\%.
This can be the reason why performances do not increase when more than 3 transition steps are performed in our model.
The advantage of our approach can be shown by contrasting the distance distributions over graphs generated either by the baseline or by our approach.

We further evaluate both approaches on a subset of the development instances, where the answer-and-question distance is at most 3 in our graph.
The accuracies of \emph{Coref LSTM} and \emph{MHQA-GRN} on this subset are 61.1 and 63.8, respectively.
Comparing with the performances on the whole devset (61.4 vs 62.8), the performance gap on this subset is increased by 1.3 points.
This indicates that our approach can better handle these ``relatively easy'' reasoning tasks.
However, as shown in Figure \ref{fig:3_dev_steps}, instances that require large reasoning steps are still challenging to our approach.

\section{Experiments on ComplexWebQuestions}

In this section, we conduct experiments on the newly released ComplexWebQuestions version 1.1 \citep{N18-1059} for better evaluating our approach.
Compared with WikiHop, where the complexity is implicitly specified in the passages, the complexity of this dataset is explicitly specified on the question side.
One example question is ``What city is the birthplace of the author of `Without end'''.
A two-step reasoning is involved, with the first step being ``the author of `Without end''' and the second being ``the birthplace of $x$''. 
$x$ is the answer of the first step.

\begin{table}
\centering
\begin{tabular}{l|c|c}
Model & Dev & Test \\
\hline
SimpQA & 30.6 & -- \\
SplitQA & 31.1 & -- \\
\hline
Local & 31.2 & 28.1 \\
MHQA-GRN w/ only same & 32.2 & -- \\
MHQA-GRN & \bf 33.2 & \bf 30.1 \\
\hline
\hline
SplitQA w/ additional labeled data & 35.6 & 34.2 \\
\end{tabular}
\caption{Results on the ComplexWebQuestions dataset.}
\label{tab:3_compqa}
\end{table}

In this dataset, web snippets (instead of passages as in WikiHop) are used for extracting answers.
The baseline of \citet{N18-1059} (\emph{SimpQA}) only uses a full question to query the web for obtaining relevant snippets, while their model (\emph{SplitQA}) obtains snippets for both the full question and its sub-questions.
With all the snippets, \emph{SplitQA} models the QA process based on a computation tree\footnote{A computation tree is a special type of semantic parse, which has two levels. The first level contains sub-questions and the second level is a composition operation.} of the full question. 
In particular, they first obtain the answers for the sub-questions, and then integrate those answers based on the computation tree.
In contrast, our approach creates a graph from all the snippets, thus the succeeding evidence integration process can join all associated evidence.

\textbf{Main results}
As shown in Table \ref{tab:3_compqa}, similar to the observations in WikiHop, \emph{MHQA-GRN} achieves large improvements over \emph{Local}.
Both the baselines and our models use all web snippets, but \emph{MHQA-GRN} further considers the structural relations among entity mentions.
\emph{SplitQA} achieves 0.5\% improvement over \emph{SimpQA}\footnote{Upon the submission time, the authors of ComplexWebQuestions have not reported testing results for the two methods. To make a fair comparison we compare the devset accuracy.}. 
Our \emph{Local} baseline is comparable with \emph{SplitQA} and our graph-based models contribute a further 2\% improvement over \emph{Local}. 
This indicates that considering structural information on passages is important for the dataset.

\vspace{0.5em}
\textbf{Analysis}~~
To deal with complex questions that require evidence from multiple passages to answer, previous work \citep{wang2018evidence,lin2018denoising,wang2018joint} 
collect evidence from occurrences of an entity in different passages.
The above methods correspond to a special case of our method, i.e. MHQA with only the \emph{same}-typed edges.
From Table \ref{tab:3_compqa}, our method gives 1 point increase over \emph{MHQA-GRN w/ only same}, and it gives more increase in WikiHop (comparing \emph{all types} with \emph{only same} in Table \ref{tab:3_dev_edges}).
Both results indicate that our method could capture more useful information for multi-hop QA tasks, compared to the methods developed for previous multi-passage QA tasks.
This is likely because our method integrates not only evidences for an entity but also these for other related entities.

The leaderboard reports \emph{SplitQA} with additional sub-question annotations and gold answers for sub-questions.
These pairs of sub-questions and answers are used as additional data for training \emph{SplitQA}. 
The above approach relies on annotations of ground-truth answers for sub-questions and semantic parses, thus is not practically useful in general. 
However, the results have additional value since it can be viewed as an upper bound of \emph{SplitQA}. 
Note that the gap between this upper bound and our \emph{MHQA-GRN} is small, which further proves that larger improvement can be achieved by introducing structural connections on the passage side to facilitate evidence integration.

\section{Related Work}

\textbf{Question answering with multi-hop reasoning}~~
Multi-hop reasoning is an important ability for dealing with difficult cases in question answering~\citep{rajpurkar-EtAl:2016:EMNLP2016,boratko2018systematic}.
Most existing work on multi-hop QA focuses on hopping over knowledge bases or tables \citep{jain2016question,neelakantan2016neural,yin2016neural}, thus the problem is reduced to deduction on a readily-defined structure with known relations.
In contrast, we study multi-hop QA on textual data and we introduce an effective approach for creating evidence integration graph structures over the textual input for solving our problems.
Previous work \citep{hill2015goldilocks,shen2017reasonet} studying multi-hop QA on text does not create reference structures.
In addition, they only evaluate their models on a simple task \citep{weston2015towards} with a very limited vocabulary and passage length.
Our work is fundamentally different from theirs by modeling structures over the input, and we evaluate our models on more challenging tasks.

Recent work starts to exploit ways for creating structures from inputs.
\citet{N18-1059} build a two-level computation tree over each question, where the first-level nodes are sub-questions and the second-level node is a composition operation.
The answers for the sub-questions are first generated, and then combined with the composition operation.
They predefine two composition operations, which makes it not general enough for other QA problems.
\citet{N18-2007} create DAGs over passages with coreference. 
The DAGs are then encoded using a DAG recurrent network.
Our work follows the second direction by creating reasoning graphs on the passage side.
However, we consider more types of relations than coreference, making a thorough study on evidence integration.
Besides, we also investigate a recent graph neural network (namely GRN) on this problem.

\subparagraph{Question answering over multiple passages}
Recent efforts in open-domain QA start to generate answers from multiple passages instead of a single passage.
However, most existing work on multi-passage QA selects the most relevant passage for answering the given question, thus reducing the problem to single-passage reading comprehension \citep{chen-EtAl:2017:Long4,dunn2017searchqa,dhingra2017quasar,wang2018r3,clark2018simple}.
Our method is fundamentally different by truly leveraging multiple passages.

A few multi-passage QA approaches merge evidence from multiple passages before selecting an answer \citep{wang2018evidence,lin2018denoising,wang2018joint}.
Similar to our work, they combine evidences from multiple passages, thus fully utilizing input passages.
The key difference is that their approaches focus on how the contexts of a single answer candidate from different passages could cover different aspects of a complex question, while our approach studies how to properly integrate the related evidence of an answer candidate, some of which come from the contexts of different entity mentions.
This increases the difficulty, since those contexts do not co-occur with the candidate answer nor the question.
When a piece of evidence does not co-occur with the answer candidate, it is usually difficult for these methods to integrate the evidence. 
This is also demonstrated by our empirical comparison, where our approach shows much better performance than combining only the evidence of the same entity mentions.

\section{Conclusion}

We have introduced a new approach for tackling multi-hop reading comprehension (MHRC), with a graph-based evidence integration process.
Given a question and a list of passages, we first connect related evidence in reference passages into a graph, and then adopt recent graph neural networks to encode resulted graphs for performing evidence integration.
Results show that the three types of edges are useful on combining global evidence and that the graph neural networks are effective on encoding complex graphs resulted by the first step.
Our approach shows highly competitive performances on two standard MHRC datasets.

%% file: 4-chap-nary.tex
In this chapter, we propose to tackle cross-sentence $n$-ary relation extraction, which aims at detecting relations among $n$ entities across multiple sentences.
Typical methods formulate an input as a \textit{document graph}, integrating various intra-sentential and inter-sentential dependencies.
The current state-of-the-art method splits the input graph into two DAGs, adopting a DAG-structured LSTM for each.
Though being able to model rich linguistic knowledge by leveraging graph edges, important information can be lost in the splitting procedure.
We propose a graph model by extending our graph recurrent network (GRN) with additional edge labels.
In particular, it uses a parallel state to model each word, recurrently enriching state values via message passing with its neighbors.
Compared with DAG LSTMs, our graph model keeps the original graph structure, and speeds up computation by allowing more parallelization.
On a standard benchmark, our model shows the best result in the literature.

\section{Introduction}

As a central task in natural language processing, relation extraction has been investigated on news, web text and biomedical domains.
It has been shown to be useful for detecting explicit facts, such as cause-effect \cite{hendrickx2009semeval}, and predicting the effectiveness of a medicine on a cancer caused by mutation of a certain gene in the biomedical domain \cite{quirk-poon:2017:EACLlong,TACL1028}.
While most existing work extracts relations within a sentence \cite{zelenko2003kernel,palmer2005proposition,zhao-grishman:2005:ACL,jiang-zhai:2007:main,plank-moschitti:2013:ACL2013,li-ji:2014:P14-1,gormley-yu-dredze:2015:EMNLP,miwa-bansal:2016:P16-1,zhang-zhang-fu:2017:EMNLP2017}, 
the task of cross-sentence relation extraction has received increasing attention \cite{gerber-chai:2010:ACL,yoshikawa2011coreference}. 
Recently, \citet{TACL1028} extend cross-sentence relation extraction by further detecting relations among several entity mentions ($n$-ary relation).
Table \ref{tab:4_task_example} shows an example, which conveys the fact that cancers caused by the \emph{858E} mutation on \emph{EGFR} gene can respond to the \emph{gefitinib} medicine.
The three entity mentions form a ternary relation yet appear in distinct sentences.

\begin{table}
\centering
\begin{tabularx}{0.99\textwidth}{|X|}
\hline
The deletion mutation on exon-19 of \textbf{EGFR} gene was present in 16 patients, while the \textbf{858E} point mutation on exon-21 was noted in 10. \\
All patients were treated with \textbf{gefitinib} and showed a partial response. \\
\hline
\end{tabularx}
\caption{An example showing that tumors with \textit{L858E} mutation in \textit{EGFR} gene respond to \textit{gefitinib} treatment.}
\label{tab:4_task_example}
\end{table}

\begin{figure}
\centering
\includegraphics[width=0.9\textwidth]{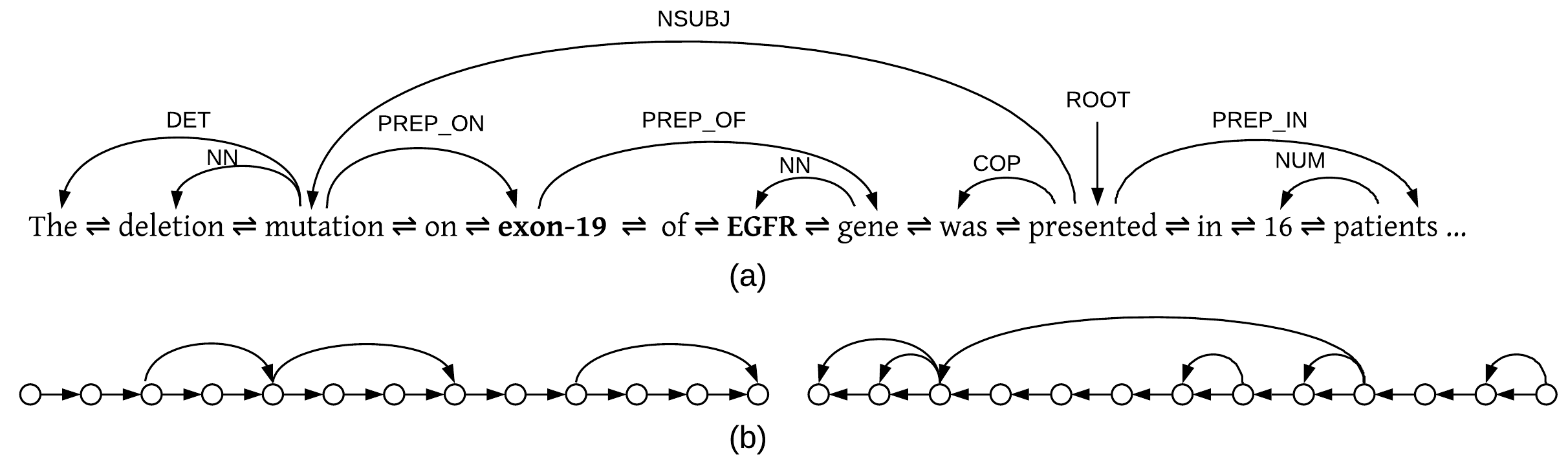}
\caption{(a) A fraction of the dependency graph of the example in Table \ref{tab:4_task_example}. For simplicity, we omit edges of discourse relations. (b) Results after splitting the graph into two DAGs.}
\label{fig:4_example_bidir}
\end{figure}

\citet{TACL1028} proposed a graph-structured LSTM for $n$-ary relation extraction.
As shown in Figure \ref{fig:4_example_bidir} (a), graphs are constructed from input sentences with dependency edges, links between adjacent words, and inter-sentence relations, so that syntactic and discourse information can be used for relation extraction.
To calculate a hidden state encoding for each word, \citet{TACL1028} first split the input graph into two directed acyclic graphs (DAGs) by separating left-to-right edges from right-to-left edges (Figure \ref{fig:4_example_bidir} (b)).
Then, two separate gated recurrent neural networks, which extend tree LSTM \cite{tai-socher-manning:2015:ACL-IJCNLP}, were adopted for each single-directional DAG, respectively.
Finally, for each word, the hidden states of both directions are concatenated as the final state.
The bi-directional DAG LSTM model showed superior performance 
over several strong baselines, such as tree-structured LSTM \cite{miwa-bansal:2016:P16-1}, 
on a biomedical-domain benchmark.

However, the bidirectional DAG LSTM model suffers from several limitations.
First, important information can be lost when converting a graph into two separate DAGs.
For the example in Figure \ref{fig:4_example_bidir}, the conversion breaks the inner structure of ``exon-19 of EGFR gene'', 
where the relation between ``exon-19'' and ``EGFR'' via the dependency path ``exon-19 $\xrightarrow{\text{PREP\_OF}}$ gene $\xrightarrow{\text{NN}}$ EGFR'' is lost from the original subgraph.
Second, using LSTMs on both DAGs, information of only ancestors and descendants can be incorporated for each word.
Sibling information, which may also be important, is not included.

A potential solution to the problems above is to model a graph as a whole, learning its representation without breaking it into two DAGs. 
Due to the existence of cycles, naive extension of tree LSTMs cannot serve this goal.
Recently, graph convolutional networks (GCN) \citep{kipf2017semi,marcheggiani-titov:2017:EMNLP2017,bastings-EtAl:2017:EMNLP2017} and graph recurrent networks (GRN) \cite{P18-1030,P18-1150} have been proposed for representing graph structures for NLP tasks.
Such methods encode a given graph by hierarchically learning representations of neighboring nodes in the graphs via their connecting edges.
While GCNs use CNN for information exchange, GRNs take gated recurrent steps to this end. 
For fair comparison with DAG LSTMs, we build a graph model by extending \citet{P18-1150}, which strictly follow the configurations of \citet{TACL1028} such as the source of features and hyper parameter settings.
In particular, the full input graph is modeled as a single state, with words in the graph being its sub states. 
State transitions are performed on the graph recurrently, allowing word-level states to exchange information through dependency and discourse edges. 
At each recurrent step, each word advances its current state by receiving information from the current states of its adjacent words. 
Thus with increasing numbers of recurrent steps each word receives information from a larger context.
Figure \ref{fig:4_transition} shows the recurrent transition steps where each node works simultaneously within each transition step.

Compared with bidirectional DAG LSTM, our method has several advantages.
First, it keeps the original graph structure, and therefore no information is lost.
Second, sibling information can be easily incorporated by passing information up and then down from a parent.
Third, information exchange allows more parallelization, and thus can be very efficient in computation.

Results show that our model outperforms a bidirectional DAG LSTM baseline by 5.9\% in accuracy,
overtaking the state-of-the-art system of \citet{TACL1028} 
by 1.2\%.
Our code is available at \url{https://github.com/freesunshine0316/nary-grn}.

Our contributions are summarized as follows. 
\begin{itemize}
\item We empirically compared our GRN with DAG LSTM for $n$-ary relation extraction tasks, showing that the former is better by more effective use of structural information;
\item To our knowledge, we are the first to investigate a graph recurrent network for modeling dependency and discourse relations.
\end{itemize}

\section{Task Definition}
\label{sec:4_task}

Formally, the input for cross-sentence $n$-ary relation extraction can be represented as a pair $(\boldsymbol{\mathcal{E}}, \boldsymbol{\mathcal{T}})$, where $\boldsymbol{\mathcal{E}} = (\epsilon_1, \dots, \epsilon_N)$ is the set of entity mentions, and $\boldsymbol{\mathcal{T}} = [S_1;\dots;S_M]$ is a text consisting of multiple sentences.
Each entity mention $\epsilon_i$ belongs to one sentence in $\boldsymbol{\mathcal{T}}$.
There is a predefined relation set $\boldsymbol{\mathcal{R}} = (r_1, \dots, r_L, \textit{None})$, where \textit{None} represents that no relation holds for the entities.
This task can be formulated as a binary classification problem of determining whether $\epsilon_1, \dots, \epsilon_N$ together form a relation \cite{TACL1028}, or a multi-class classification problem of detecting which relation holds for the entity mentions.  
Take Table \ref{tab:4_task_example} as an example. 
The binary classification task is to determine whether \emph{gefitinib} would have an effect on this type of cancer, given a cancer patient with \emph{858E} mutation on gene \emph{EGFR}.
The multi-class classification task is to detect the exact drug effect: response, resistance, sensitivity, etc.

\section{Baseline: Bi-directional DAG LSTM}
\label{sec:4_baseline}

\citet{TACL1028} formulate the task as a graph-structured problem in order to adopt rich dependency and discourse features.
In particular, Stanford parser \cite{manning-EtAl:2014:P14-5} is used to assign syntactic structure to input sentences, and heads of two consecutive sentences are connected to represent discourse information, resulting in a graph structure.
For each input graph $\boldsymbol{G}=(\boldsymbol{V},\boldsymbol{E})$, the nodes $\boldsymbol{V}$ are words within input sentences, and each edge $e \in \boldsymbol{E}$  connects two words that either have a relation or are adjacent to each other.
Each edge is denoted as a triple $(i,j,l)$, where $i$ and $j$ are the indices of the source and target words, respectively, and the edge label $l$ indicates either a dependency or discourse relation (such as ``nsubj'') or a relative position (such as ``next\_tok'' or ``prev\_tok'').
Throughout this paper, we use $\boldsymbol{E}_{in}(j)$ and $\boldsymbol{E}_{out}(j)$ to denote the sets of incoming and outgoing edges for word $j$.

For a bi-directional DAG LSTM baseline, we follow \citet{TACL1028}, splitting each input graph into two separate DAGs by separating left-to-right edges from right-to-left edges (Figure \ref{fig:4_example_bidir}).
Each DAG is encoded by using a DAG LSTM (Section \ref{sec:4_baseline_gated}), which takes both source words and edge labels as inputs (Section \ref{sec:4_baseline_input}).
Finally, the hidden states of entity mentions from both LSTMs are taken as inputs to a logistic regression classifier to make a prediction:
\begin{equation}
\hat{y} = \text{softmax}(\boldsymbol{W}_0 [\boldsymbol{h}_{\epsilon_1}; \dots; \boldsymbol{h}_{\epsilon_N}] + \boldsymbol{b}_0) \text{,}
\end{equation}
where $\boldsymbol{h}_{\epsilon_j}$ is the hidden state of entity $\epsilon_j$.
$\boldsymbol{W}_0$ and $\boldsymbol{b}_0$ are parameters.

\subsection{Input Representation}
\label{sec:4_baseline_input}

Both nodes and edge labels are useful for modeling a syntactic graph.
As the input to our DAG LSTM, we first calculate the representation for each edge $(i,j,l)$ by:
\begin{equation} \label{eq:4_baseline_edge}
\boldsymbol{x}_{i,j}^l = \boldsymbol{W}_1 \Big( [\boldsymbol{e}_l; \boldsymbol{e}_i] \Big) + \boldsymbol{b}_1 \text{,}
\end{equation}
where $\boldsymbol{W}_1$ and $\boldsymbol{b}_1$ are model parameters, $\boldsymbol{e}_i$ is the embedding of the source word indexed by $i$, and $\boldsymbol{e}_l$ is the embedding of the edge label $l$.

\subsection{Encoding process}
\label{sec:4_baseline_gated}

The baseline LSTM model learns DAG representations sequentially, following word orders.
Taking the edge representations (such as $\boldsymbol{x}_{i,j}^l$) as input, gated state transition operations are executed on both the forward and backward DAGs.
For each word $j$, the representations of its incoming edges $\boldsymbol{E}_{in}(j)$ are summed up as one vector:
\begin{equation} \label{eq:4_baseline_input}
\boldsymbol{x}_j^{in} = \sum_{(i,j,l)\in \boldsymbol{E}_{in}(j)} \boldsymbol{x}_{i,j}^l \\
\end{equation}
Similarly, for each word $j$, the states of all incoming nodes are summed to a single vector before being passed to the gated operations:
\begin{equation}
\boldsymbol{h}_j^{in} = \sum_{(i,j,l)\in \boldsymbol{E}_{in}(j)} \boldsymbol{h}_{i} \\
\end{equation}
Finally, the gated state transition operation for the hidden state $\boldsymbol{h}_j$ of the $j$-th word can be defined as:
\begin{equation} \label{eq:4_baseline_gated}
\begin{split}
\boldsymbol{i}_j &= \sigma(\boldsymbol{W}_i \boldsymbol{x}_j^{in} + \boldsymbol{U}_i \boldsymbol{h}_j^{in} + \boldsymbol{b}_i) \\
\boldsymbol{o}_j &= \sigma(\boldsymbol{W}_o \boldsymbol{x}_j^{in} + \boldsymbol{U}_o \boldsymbol{h}_j^{in} + \boldsymbol{b}_o) \\
\boldsymbol{f}_{i,j} &= \sigma(\boldsymbol{W}_f \boldsymbol{x}_{i,j}^l + \boldsymbol{U}_f \boldsymbol{h}_i + \boldsymbol{b}_f) \\
\boldsymbol{u}_j &= \sigma(\boldsymbol{W}_u \boldsymbol{x}_j^{in} + \boldsymbol{U}_u \boldsymbol{h}_j^{in} + \boldsymbol{b}_u) \\
\boldsymbol{c}_j &= \boldsymbol{i}_j \odot \boldsymbol{u}_j + \sum_{(i,j,l)\in \boldsymbol{E}_{in}(j)} \boldsymbol{f}_{i,j} \odot \boldsymbol{c}_i \\
h_j &= o_j \odot \tanh (c_j) \text{,} \\
\end{split}
\end{equation}
where $\boldsymbol{i}_j$, $\boldsymbol{o}_j$ and $\boldsymbol{f}_{i,j}$ are a set of input, output and forget gates, respectively, and $\boldsymbol{W}_x$, $\boldsymbol{U}_x$ and $\boldsymbol{b}_x$ ($x \in \{i,o,f,u\}$) are model parameters.

\subsection{Comparison with \citet{TACL1028}}
\label{sec:4_baseline_comp}

Our baseline is computationally similar to \citet{TACL1028}, but different on how to utilize edge labels in the gated network.
In particular, \citet{TACL1028} make model parameters specific to edge labels. 
They consider two model variations, namely \emph{Full Parametrization (FULL)} and \emph{Edge-Type Embedding (EMBED)}.
\emph{FULL} assigns distinct $U$s (in Equation \ref{eq:4_baseline_gated}) to different edge types, so that each edge label is associated with a 2D weight matrix to be tuned in training.
On the other hand, \emph{EMBED} assigns each edge label to an embedding vector, but complicates the gated operations by changing the $U$s to be 3D tensors.\footnote{For more information please refer Section 3.3 of \citet{TACL1028}.}

In contrast, we take edge labels as part of the input to the gated network.
In general, the edge labels are first represented as embeddings, before being concatenated with the node representation vectors (Equation \ref{eq:4_baseline_edge}).
We choose this setting for both the baseline and our GRN in Section \ref{sec:4_model}, since it requires fewer parameters compared with \emph{FULL} and \emph{EMBED}, thus being less exposed to overfitting on small-scaled data.

\section{Encoding with Graph Recurrent Network}
\label{sec:4_model}

Our input graph formulation strictly follows Section \ref{sec:4_baseline}. 
In particular, our model adopts the same methods for calculating input representation (as in Section \ref{sec:4_baseline_input}) and performing classification as the baseline model.
However, different from the baseline bidirectional DAG LSTM model, we leverage GRN to directly model the input graph, without splitting it into two DAGs.
Comparing with the evidence graphs shown in Chapter \ref{chap:mhqa}, the dependency graphs are directed and contain edge labels that provide important information.
Here we adapt GRN to further incorporate this information.

\begin{figure}
\centering
\includegraphics[width=0.75\textwidth]{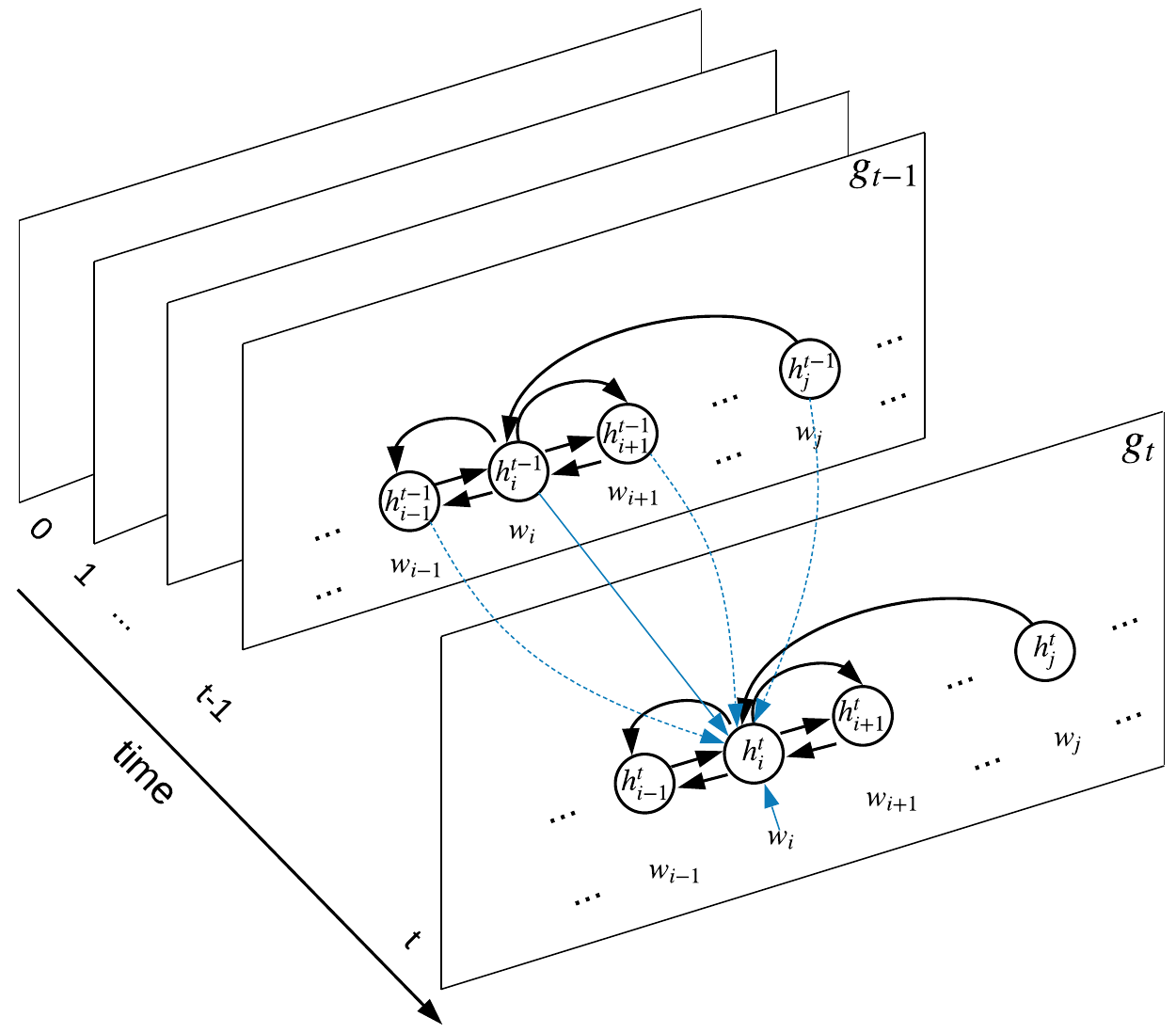}
\caption{GRN encoding for a dependency graph, where each $w_i$ is a word.}
\label{fig:4_transition}
\end{figure}

Figure \ref{fig:4_transition} shows an overview of the GRN encoder for dependency graphs.
Formally, given an input graph $\boldsymbol{G}=(\boldsymbol{V}, \boldsymbol{E})$, we define a state vector $\boldsymbol{h}^j$ for each word $v_j \in \boldsymbol{V}$. 
The state of the graph consists of all word states, and thus can be represented as:
\begin{equation}
\boldsymbol{g} = \{\boldsymbol{h}^j\}|_{v_j \in \boldsymbol{V}}
\end{equation}
Same as Chapter \ref{chap:mhqa}, the GRN-based encoder performs information exchange between neighboring words through a recurrent state transition process, resulting in a sequence of graph states $\boldsymbol{g}_0, \boldsymbol{g}_1, \dots, \boldsymbol{g}_t$, where $\boldsymbol{g}_t = \{\boldsymbol{h}_t^j\}|_{v_j \in \boldsymbol{V}}$, and the initial graph state $\boldsymbol{g}_0$ consists of a set of initial word states $\boldsymbol{h}_0^j=\boldsymbol{h}_0$, where $\boldsymbol{h}_0$ is a zero vector.
The main change is on message aggregation, where we further distinguish incoming neighbors and outgoing neighbors, and edge labels are also incorporated.


For each time step $t$, the message to a word $v_j$ includes the representations of the edges that are connected to $v_j$, where $v_j$ can be either the source or the target of the edge.
Similar to Section \ref{sec:4_baseline_input}, we define each edge as a triple $(i,j,l)$, where $i$ and $j$ are indices of the source and target words, respectively, and $l$ is the edge label.
$\boldsymbol{x}_{i,j}^l$ is the representation of edge $(i,j,l)$.
The inputs for $v_j$ are distinguished by incoming and outgoing directions, where:
\begin{equation}
\begin{split}
\boldsymbol{x}_j^{in} &= \sum_{(i,j,l)\in \boldsymbol{E}_{in}(j)} \boldsymbol{x}_{i,j}^l \\
\boldsymbol{x}_j^{out} &= \sum_{(j,k,l)\in \boldsymbol{E}_{out}(j)} \boldsymbol{x}_{j,k}^l \\
\end{split}
\end{equation}
Here $\boldsymbol{E}_{in}(j)$ and $\boldsymbol{E}_{out}(j)$ denote the sets of incoming and outgoing edges of $v_j$, respectively.

In addition to edge inputs, the message also contains the hidden states of its incoming and outgoing words during a state transition. 
In particular, the states of all incoming words and outgoing words are summed up, respectively:
\begin{equation}
\begin{split}
\boldsymbol{h}_j^{in} &= \sum_{(i,j,l)\in \boldsymbol{E}_{in}(j)} \boldsymbol{h}_{t-1}^{i} \\
\boldsymbol{h}_j^{out} &= \sum_{(j,k,l)\in \boldsymbol{E}_{out}(j)} \boldsymbol{h}_{t-1}^{k} \text{,} \\
\end{split}
\end{equation} 
Based on the above definitions of $\boldsymbol{x}_j^{in}$, $\boldsymbol{x}_j^{out}$, $\boldsymbol{h}_j^{in}$ and $\boldsymbol{h}_j^{out}$, the message $\boldsymbol{m}_t^j$ is aggregated by their concatenation:
\begin{equation}
    \boldsymbol{m}_t^j = [\boldsymbol{x}_j^{in}; \boldsymbol{x}_j^{out}; \boldsymbol{h}_j^{in}; \boldsymbol{h}_j^{out}] \text{,}
\end{equation}
before being applied with an LSTM step (defined in Equation \ref{eq:3_grn_mp}) to update the node hidden state $\boldsymbol{h}_{t-1}^j$:
\begin{equation}
    \boldsymbol{h}_t^j, \boldsymbol{c}_t^j = \text{LSTM}(\boldsymbol{m}_t^j, [\boldsymbol{h}_{t-1}^j, \boldsymbol{c}_{t-1}^j]) \text{,}
\end{equation}
where $\boldsymbol{c}^j$ is the cell memory for hidden state $\boldsymbol{h}^j$.



\subparagraph{GRN vs bidirectional DAG LSTM}
A contrast between the baseline DAG LSTM and our graph LSTM can be made from the perspective of information flow. 
For the baseline, information flow follows the natural word order in the input sentence, with the two DAG components propagating information from left to right and from right to left, respectively. 
In contrast, information flow in our GRN is relatively more concentrated at individual words, with each word exchanging information with all its graph neighbors simultaneously at each sate transition. 
As a result, wholistic contextual information can be leveraged for extracting features for each word, as compared to separated handling of bi-directional information flow in DAG LSTM. 
In addition, arbitrary structures, including arbitrary cyclic graphs, can be handled.

From an initial state with isolated words, information of each word propagates to its graph neighbors after each step. 
Information exchange between non-neighboring words can be achieved through multiple transition steps.
We experiment with different transition step numbers to study the effectiveness of global encoding. 
Unlike the baseline DAG LSTM encoder, our model allows parallelization in node-state updates, and thus can be highly efficient using a GPU.

\section{Training}

We train our models with a cross-entropy loss over a set of gold standard data:
\begin{equation}
l = -\log p(y_i|\boldsymbol{X}_i;\boldsymbol{\theta})\text{,}
\end{equation}
where $\boldsymbol{X}_i$ is an input graph, $y_i$ is the gold class label of $\boldsymbol{X}_i$, and $\boldsymbol{\theta}$ is the model parameters.
Adam \cite{kingma2014adam} with a learning rate of 0.001 is used as the optimizer, and the model that yields the best devset performance is selected to evaluate on the test set.
Dropout with rate 0.3 is used during training.
Both training and evaluation are conducted using a Tesla K20X GPU.

\section{Experiments}

We conduct experiments for the binary relation detection task and the multi-class relation extraction task discussed in Section \ref{sec:4_task}.

\subsection{Data}
\label{sec:4_data}

\begin{table} 
\centering
\begin{tabular} {l|c|c|c}
\hline
Data & Avg. Tok. & Avg. Sent. & Cross (\%) \\
\hline
\textsc{Ternary} & 73.9 & 2.0 & 70.1\% \\ 
\textsc{Binary}  & 61.0 & 1.8 & 55.2\% \\
\hline
\end{tabular}
\caption{Dataset statistics. \emph{Avg.} \emph{Tok.} and \emph{Avg.} \emph{Sent.} are the average number of tokens and sentences, respectively. \emph{Cross} is the percentage of instances that contain multiple sentences.}
\label{tab:4_stat}
\end{table}

We use the dataset of \citet{TACL1028}, which is a biomedical-domain dataset focusing on drug-gene-mutation ternary relations,\footnote{The dataset is available at \url{http://hanover.azurewebsites.net}.} extracted from PubMed.
It contains 6987 ternary instances about drug-gene-mutation relations, and 6087 binary instances about drug-mutation sub-relations.
Table \ref{tab:4_stat} shows statistics of the dataset. 
Most instances of ternary data contain multiple sentences, and the average number of sentences is around 2.
There are five classification labels: ``resistance or non-response'', ``sensitivity'', ``response'', ``resistance'' and ``None''.
We follow \citet{TACL1028} and binarize multi-class labels by grouping all relation classes as ``Yes'' and treat ``None'' as ``No''.

\subsection{Settings}

Following \citet{TACL1028}, five-fold cross-validation is used for evaluating the models,\footnote{The released data has been separated into 5 portions, and we follow the exact split.}
and the final test accuracy is calculated by averaging the test accuracies over all five folds.
For each fold, we randomly separate 200 instances from the training set for development.
The batch size is set as 8 for all experiments. 
Word embeddings are initialized with the 100-dimensional GloVe \cite{pennington2014glove} vectors, pretrained on 6 billion words from Wikipedia and web text.
The edge label embeddings are 3-dimensional and randomly initialized.
Pretrained word embeddings are not updated during training.
The dimension of hidden vectors in LSTM units is set to 150.

\subsection{Development Experiments}

\begin{figure}
\centering
\includegraphics[width=0.75\textwidth]{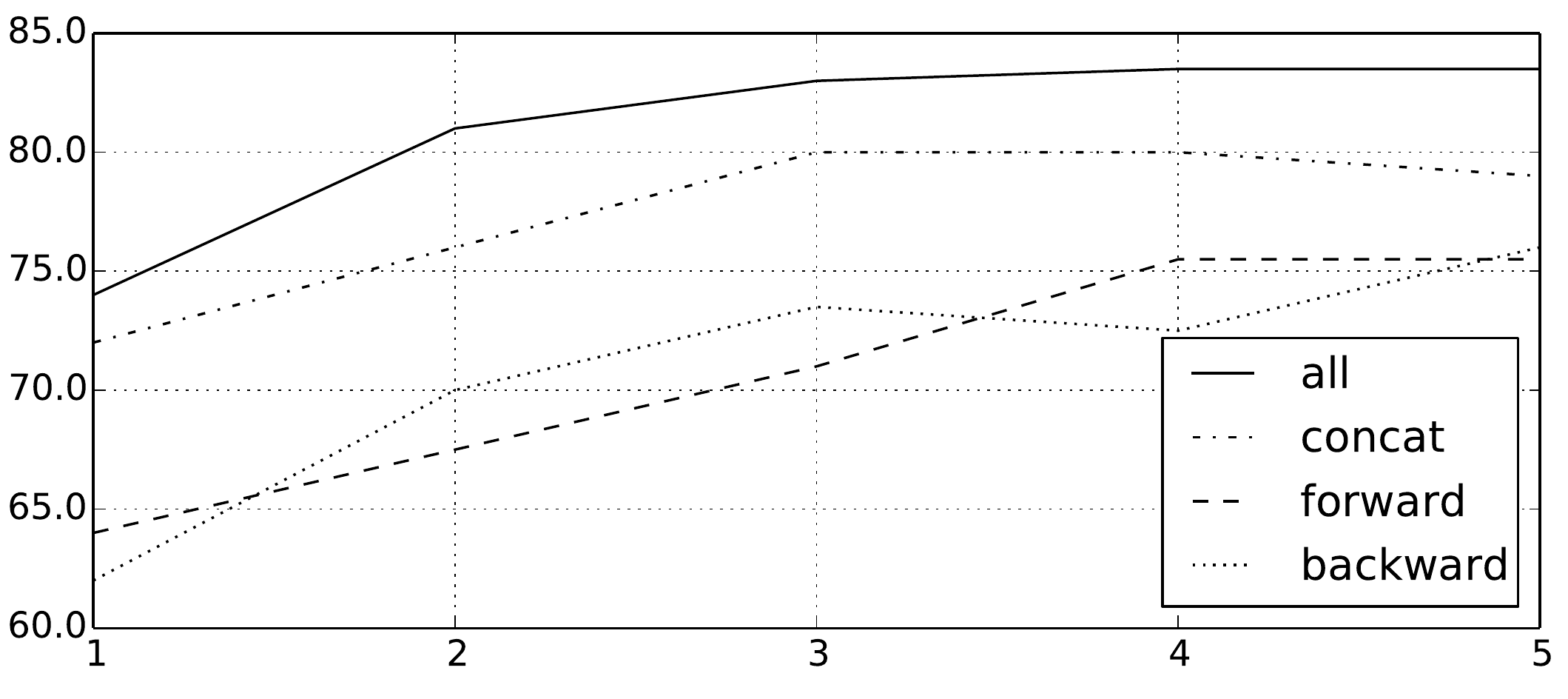}
\caption{Dev accuracies against transition steps for GRN.}
\label{fig:4_dev}
\end{figure}

We first analyze our model on the drug-gene-mutation ternary relation dataset, taking the first among 5-fold cross validation settings for our data setting.
Figure \ref{fig:4_dev} shows the devset accuracies of different state transition numbers, where \emph{forward} and \emph{backward} execute our graph state model only on the forward or backward DAG, respectively.
\emph{Concat} concatenates the hidden states of \emph{forward} and \emph{backward}.
\emph{All} executes our graph state model on original graphs.

The performance of \emph{forward} and \emph{backward} lag behind \emph{concat}, which is consistent with the intuition that both forward and backward relations are useful \cite{TACL1028}.
In addition, \emph{all} gives better accuracies compared with \emph{concat}, demonstrating the advantage of simultaneously considering forward and backward relations during representation learning.
For all the models, more state transition steps result in better accuracies, where larger contexts can be integrated in the representations of graphs.
The performance of \emph{all} starts to converge after 4 and 5 state transitions, so we set the number of state transitions to 5 in the remaining experiments.

\subsection{Final results}
\label{sec:4_results}

\begin{table}
\centering
\begin{tabular}{lcc}
\hline
Model & Single & Cross  \\
\hline
\citet{quirk-poon:2017:EACLlong} & 74.7 & 77.7 \\ 
\citet{TACL1028} - EMBED & 76.5 &  80.6 \\ 
\citet{TACL1028} - FULL & 77.9 & 80.7 \\ 
~~~~~~~~~~~+ multi-task & -- & 82.0 \\ 
\hline
Bidir DAG LSTM & 75.6 & 77.3 \\
GRN  & \textbf{80.3*} & \textbf{83.2*}  \\
\hline
\end{tabular}
\caption{Average test accuracies for \textsc{Ternary} drug-gene-mutation interactions. \emph{Single} represents experiments only on instances within single sentences, while \emph{Cross} represents experiments on all instances. *: significant at $p<0.01$}
\label{tab:4_ternary}
\end{table}

Table \ref{tab:4_ternary} compares our model with the bidirectional DAG baseline and the state-of-the-art results on this dataset, where \emph{EMBED} and \emph{FULL} have been briefly introduced in Section \ref{sec:4_baseline_comp}.
\emph{+multi-task} applies joint training of both ternary (drug-gene-mutation) relations and their binary (drug-mutation) sub-relations.
\citet{quirk-poon:2017:EACLlong} use a statistical method with a logistic regression classifier and features derived from shortest paths between all entity pairs.
\emph{Bidir DAG LSTM} is our bidirectional DAG LSTM baseline, and \emph{GRN} is our GRN-based model.

Using all instances (the \emph{Cross} column in Table \ref{tab:4_ternary}), our graph model shows the highest test accuracy among all methods, which is 5.9\% higher than our baseline.\footnote{$p<0.01$ using t-test. For the remaining of this paper, we use the same measure for statistical significance.}
The accuracy of our baseline is lower than \emph{EMBED} and \emph{FULL} of \citet{TACL1028}, which is likely due to the differences mentioned in Section \ref{sec:4_baseline_comp}.
Our final results are better than \citet{TACL1028}, despite the fact that we do not use multi-task learning.

We also report accuracies only on instances within single sentences (column \emph{Single} in Table \ref{tab:4_ternary}), which exhibit similar contrasts.
Note that all systems show performance drops when evaluated only on single-sentence relations, which are actually more challenging.
One reason may be that some single sentences cannot provide sufficient context for disambiguation, making it necessary to study cross-sentence context.
Another reason may be overfitting caused by relatively fewer training instances in this setting, as only 30\% instances are within a single sentence. 
One interesting observation is that our baseline shows the least performance drop of 1.7 points, in contrast to up to 4.1 for other neural systems.
This can be a supporting evidence for overfitting, as our baseline has fewer parameters at least than \emph{FULL} and \emph{EMBED}.

\subsection{Analysis}

\begin{table}
\centering
\begin{tabular}{lcc}
\hline
Model & Train & Decode \\
\hline
Bidir DAG LSTM & 281s & 27.3s \\
GRN  & 36.7s & 2.7s \\
\hline
\end{tabular}
\caption{The average times for training one epoch and decoding (seconds) over five folds on drug-gene-mutation \textsc{Ternary} cross sentence setting.}
\label{tab:time}
\end{table}

\subparagraph{Efficiency.} 
Table \ref{tab:time} shows the training and decoding time of both the baseline and our model.
Our model is 8 to 10 times faster than the baseline in training and decoding speeds, respectively.
By revisiting Table \ref{tab:4_stat}, we can see that the average number of tokens for the ternary-relation data is 74, which means that the baseline model has to execute 74 recurrent transition steps for calculating a hidden state for each input word. 
On the other hand, our model only performs 5 state transitions, and calculations between each pair of nodes for one transition are parallelizable.
This accounts for the better efficiency of our model.

\begin{figure}
\centering
\includegraphics[width=0.9\textwidth]{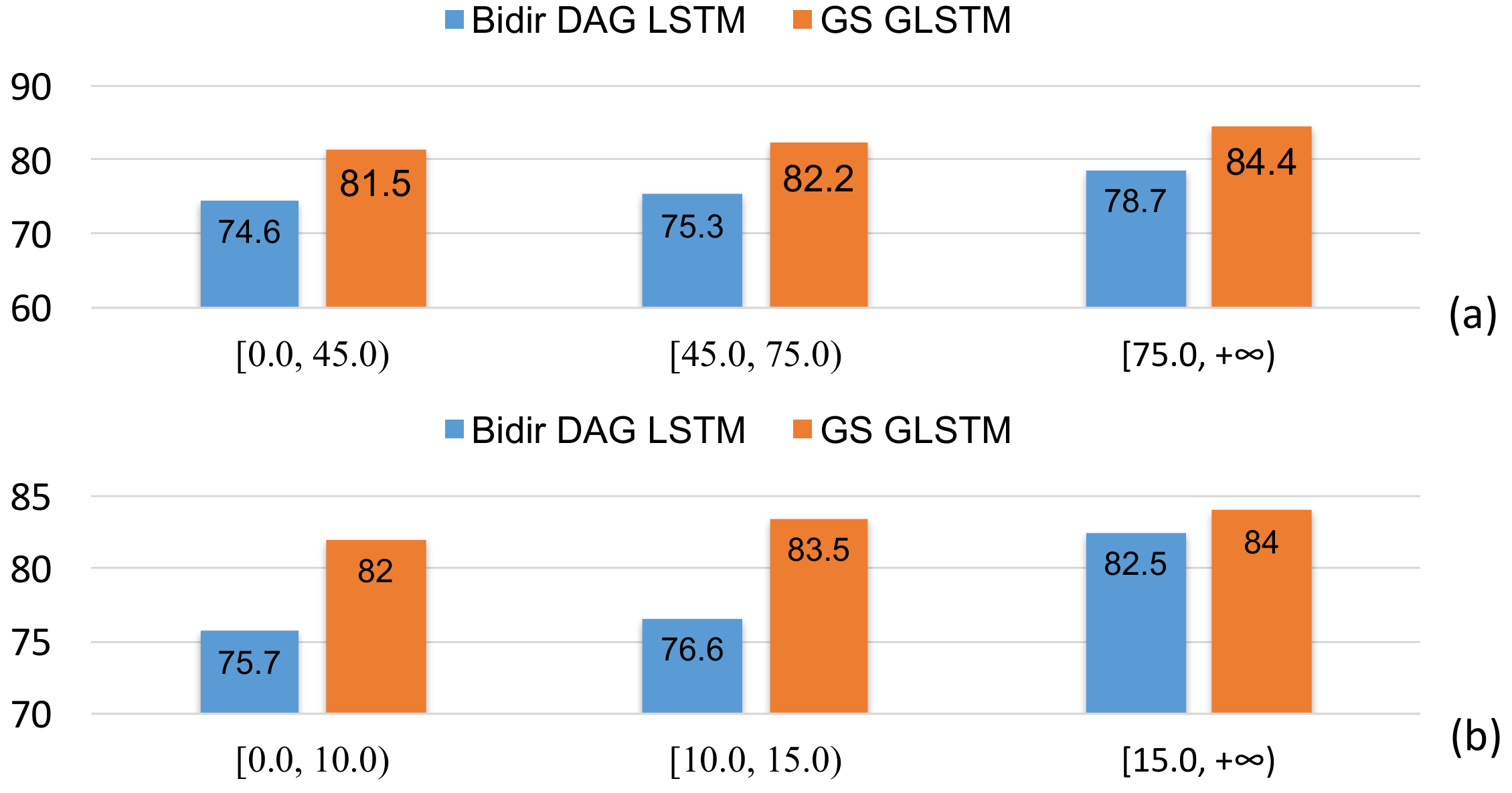}
\caption{Test set performances on (a) different sentence lengths, and (b) different maximal number of neighbors.}
\label{fig:4_by_length}
\end{figure}

\subparagraph{Accuracy against sentence length}
Figure \ref{fig:4_by_length} (a) shows the test accuracies on different sentence lengths.
We can see that \emph{GRN} and \emph{Bidir DAG LSTM} show performance increase along increasing input sentence lengths.
This is likely because longer contexts provide richer information for relation disambiguation.
\emph{GRN} is consistently better than \emph{Bidir DAG LSTM}, and the gap is larger on shorter instances.
This demonstrates that \emph{GRN} is more effective in utilizing a smaller context for disambiguation.

\subparagraph{Accuracy against the maximal number of neighbors}
Figure \ref{fig:4_by_length} (b) shows the test accuracies against the maximum number of neighbors.
Intuitively, it is easier to model graphs containing nodes with more neighbors, because these nodes can serve as a ``supernode'' that allow more efficient information exchange. 
The performances of both \emph{GRN} and \emph{Bidir DAG LSTM} increase with increasing maximal number of neighbors, which coincide with this intuition.
In addition, \emph{GRN} shows more advantage than \emph{Bidir DAG LSTM} under the inputs having lower maximal number of neighbors, which further demonstrates the superiority of \emph{GRN} over \emph{Bidir DAG LSTM} in utilizing context information.

\begin{figure}
\centering
\includegraphics[width=0.99\textwidth]{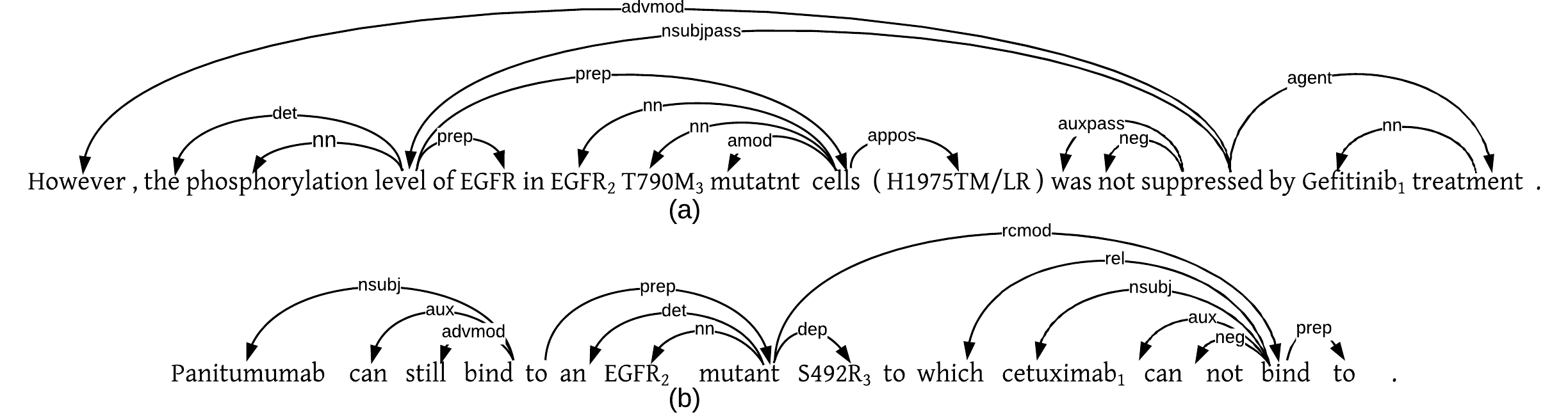}
\caption{Example cases. Words with subindices 1, 2 and 3 represent drugs, genes and mutations, respectively. References for both cases are ``No''. For both cases, \emph{GRN} makes the correct predictions, while \emph{Bidir DAG LSTM} does incorrectly.}
\label{fig:4_case_study}
\end{figure}

\subparagraph{Case study}
Figure \ref{fig:4_case_study} visualizes the merits of \emph{GRN} over \emph{Bidir DAG LSTM} using two examples. 
\emph{GRN} makes the correct predictions for both cases, while \emph{Bidir DAG LSTM} fails to.
The first case generally mentions that \emph{Gefitinib} does not have an effect on \emph{T790M} mutation on \emph{EGFR} gene.
Note that both ``However'' and ``was not'' serve as indicators; thus incorporating them into the contextual vectors of these entity mentions is important for making a correct prediction.
However, both indicators are leaves of the dependency tree, making it impossible for \emph{Bidir DAG LSTM} to incorporate them into the contextual vectors of entity mentions up the tree through dependency edges.\footnote{As shown in Figure \ref{fig:4_example_bidir}, a directional DAG LSTM propagates information according to the edge directions.}
On the other hand, it is easier for \emph{GRN}.
For instance, ``was not'' can be incorporated into ``Gefitinib'' through ``suppressed $\xrightarrow{\text{agent}}$ treatment $\xrightarrow{\text{nn}}$ Gefitinib''.

\begin{table}
\centering
\begin{tabular}{lcccc}
\hline
Model & Single & Cross \\
\hline
\citet{quirk-poon:2017:EACLlong} & 73.9 & 75.2 \\
\citet{miwa-bansal:2016:P16-1} & 75.9 & 75.9 \\ 
\citet{TACL1028} - EMBED & 74.3 &  76.5 \\ 
\citet{TACL1028} - FULL & 75.6 & 76.7 \\ 
~~~~~~~~~~~+ multi-task & -- & 78.5 \\ 
\hline
Bidir DAG LSTM & 76.9 & 76.4 \\
GRN  & \textbf{83.5*} & \textbf{83.6*} \\
\hline
\end{tabular}
\caption{Average test accuracies in five-fold cross-validation for \textsc{Binary} drug-mutation interactions.}
\label{tab:4_binary}
\end{table}

The second case is to detect the relation among ``cetuximab'' (drug), ``EGFR'' (gene) and ``S492R'' (mutation), which does not exist.
However, the context introduces further ambiguity by mentioning another drug ``Panitumumab'', which does have a  relation with ``EGFR'' and ``S492R''.
Being sibling nodes in the dependency tree, ``can not'' is an indicator for the relation of ``cetuximab''.
\emph{GRN} is correct, because ``can not'' can be easily included into the contextual vector of ``cetuximab'' in two steps via ``bind $\xrightarrow{\text{nsubj}}$cetuximab''.

\subsection{Results on Binary Sub-relations}

Following previous work, we also evaluate our model on drug-mutation binary relations.
Table \ref{tab:4_binary} shows the results, where \citet{miwa-bansal:2016:P16-1} is a state-of-the-art model using sequential and tree-structured LSTMs to jointly capture linear and dependency contexts for relation extraction.
Other models have been introduced in Section \ref{sec:4_results}.

Similar to the ternary relation extraction experiments, \emph{GRN} outperforms all the other systems with a large margin, which shows that the message passing graph LSTM is better at encoding rich linguistic knowledge within the input graphs.
Binary relations being easier, both \emph{GRN} and \emph{Bidir DAG LSTM} show increased or similar performances compared with the ternary relation experiments. 
On this set, our bidirectional DAG LSTM model is comparable to \emph{FULL} using all instances (``Cross'') and slightly better than \emph{FULL} using only single-sentence instances (``Single'').

\subsection{Fine-grained Classification}

Our dataset contains five classes as mentioned in Section \ref{sec:4_data}.
However, previous work only investigates binary relation detection.
Here we also study the multi-class classification task, which can be more informative for applications.

Table \ref{tab:4_multi_class} shows accuracies on multi-class relation extraction, which makes the task more ambiguous compared with binary relation extraction.
The results show similar comparisons with the binary relation extraction results.
However, the performance gaps between \emph{GRN} and \emph{Bidir DAG LSTM} dramatically increase, showing the superiority of \emph{GRN} over \emph{Bidir DAG LSTM} in utilizing context information.

\section{Related Work}

\subparagraph{$N$-ary relation extraction}
$N$-ary relation extractions can be traced back to MUC-7 \cite{chinchor1998overview}, which focuses on entity-attribution relations.
It has also been studied in biomedical domain \cite{mcdonald-EtAl:2005:ACL}, but only the instances within a single sentence are considered.
Previous work on cross-sentence relation extraction relies on either explicit co-reference annotation \cite{gerber-chai:2010:ACL,yoshikawa2011coreference}, or the assumption that the whole document refers to a single coherent event \cite{wick-culotta-mccallum:2006:EMNLP,swampillai-stevenson:2011:RANLP}.
Both simplify the problem and reduce the need for learning better contextual representation of entity mentions.
A notable exception is \citet{quirk-poon:2017:EACLlong}, who adopt distant supervision and integrated contextual evidence of diverse types without relying on these assumptions.
However, they only study binary relations.
We follow \citet{TACL1028} by studying ternary cross-sentence relations.

\begin{table}
\centering
\begin{tabular}{lcc}
\hline
Model & \textsc{Ternary} & \textsc{Binary} \\
\hline
Bidir DAG LSTM & 51.7 & 50.7 \\
GRN & \textbf{71.1*} & \textbf{71.7*} \\
\hline
\end{tabular}
\caption{Average test accuracies for multi-class relation extraction with all instances (``Cross'').}
\label{tab:4_multi_class}
\end{table}

\textbf{Graph encoder}
\citet{liang2016semantic} build a graph LSTM model for semantic object parsing, which aims to segment objects within an image into more fine-grained, semantically meaningful parts.
The nodes of an input graph come from image superpixels, and the edges are created by connecting spatially neighboring nodes.
Their model is similar as \citet{TACL1028} by calculating node states sequentially: for each input graph, a start node and a node sequence are chosen, which determines the order of recurrent state updates.
In contrast, our graph LSTM do not need ordering of graph nodes, and is highly parallelizable.

\section{Conclusion}

We explored graph recurrent network for cross-sentence $n$-ary relation extraction, which uses a recurrent state transition process to incrementally refine a neural graph state representation capturing graph structure contexts.
Compared with a bidirectional DAG LSTM baseline, our model has several advantages.
First, it does not change the input graph structure, so that no information can be lost.
For example, it can easily incorporate sibling information when calculating the contextual vector of a node.
Second, it is better parallelizable.
Experiments show significant improvements over the previously reported numbers, including that of the bidirectional graph LSTM model.

%% file: 5-chap-amrgen.tex
The problem of AMR-to-text generation is to recover a text representing the same meaning as an input AMR graph.
The current state-of-the-art method uses a sequence-to-sequence model, leveraging LSTM for encoding a linearized AMR structure. 
Although it is able to model non-local semantic information, a sequence LSTM can lose information from the AMR graph structure, and thus faces challenges with large graphs, which result in long sequences. 
We introduce a neural graph-to-sequence model, using a novel LSTM structure for directly encoding graph-level semantics.
On a standard benchmark, our model shows superior results to existing methods in the literature.

\section{Introduction}

\begin{figure}
\centering
\includegraphics[scale=0.9]{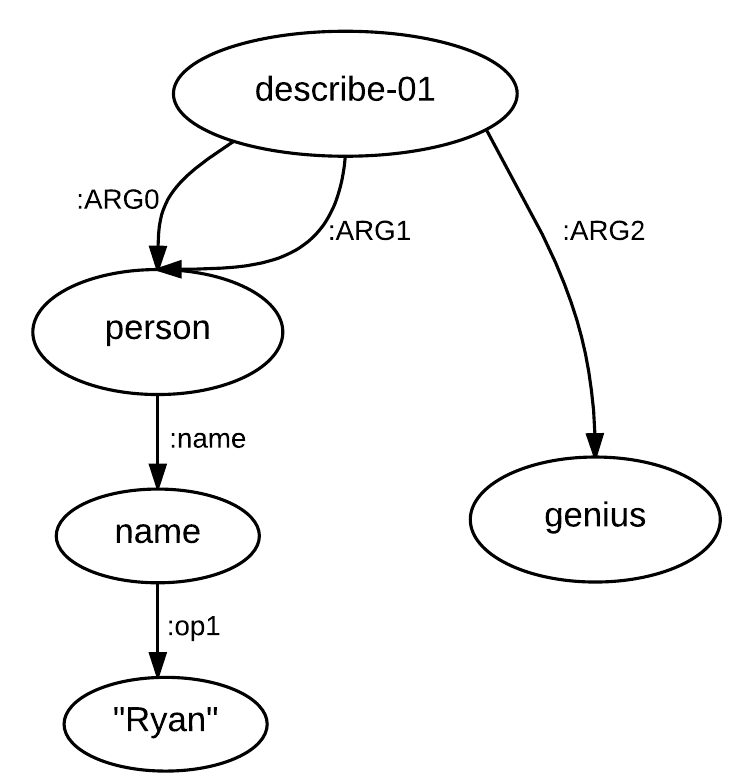}
\caption{An example of AMR graph meaning ``Ryan's description of himself: a genius.''}
\label{fig:5_example_amr}
\end{figure}

Abstract Meaning Representation (AMR) \cite{banarescu-EtAl:2013:LAW7-ID} is a semantic formalism that encodes the meaning of a sentence as a rooted, directed graph.
Figure \ref{fig:5_example_amr} shows an AMR graph in which the nodes (such as ``describe-01'' and ``person'') represent the concepts, and edges (such as ``:ARG0'' and ``:name'') represent the relations between concepts they connect.
AMR has been proven helpful on other NLP tasks, such as machine translation \cite{jones2012semantics,tamchyna-quirk-galley:2015:S2MT}, question answering \cite{mitra2015addressing}, summarization \cite{takase-EtAl:2016:EMNLP2016} and event detection \cite{li-EtAl:2015:CNewsStory}.

The task of AMR-to-text generation is to produce a text with the same meaning as a given input AMR graph. 
The task is challenging as word tenses and function words are abstracted away when constructing AMR graphs from texts.
The translation from AMR nodes to text phrases can be far from literal.
For example, shown in Figure \ref{fig:5_example_amr}, ``Ryan'' is represented as ``(p / person :name (n / name :op1 ``Ryan''))'', and ``description of'' is represented as ``(d / describe-01 :ARG1 )''.

While initial work used statistical approaches \cite{jeff2016amrgen,pourdamghani-knight-hermjakob:2016:INLG,song-acl17,lampouras-vlachos:2017:SemEval,mille-EtAl:2017:SemEval,gruzitis-gosko-barzdins:2017:SemEval}, recent research has demonstrated the success of deep learning, and in particular the sequence-to-sequence model \cite{sutskever2014sequence}, which has achieved the state-of-the-art results on AMR-to-text generation \cite{konstas-EtAl:2017:Long}. 
One limitation of sequence-to-sequence models,
however, is that they require serialization of input AMR graphs, which adds to the challenge of representing graph structure information, especially when the graph is large.
In particular, closely-related nodes, such as parents, children and siblings can be far away after serialization.
It can be difficult for a linear recurrent neural network to automatically induce their original connections from bracketed string forms.

To address this issue, we introduce a novel graph-to-sequence model, where a graph recurrent network (GRN) is used to encode AMR structures directly.
To capture non-local information, the encoder performs graph state transition by information exchange between connected nodes, with a graph state consisting of all node states.
Multiple recurrent transition steps are taken so that information can propagate non-locally, and LSTM \cite{hochreiter1997long} is used to avoid gradient diminishing and bursting in the recurrent process.
The decoder is an attention-based LSTM model with a 
copy mechanism \cite{gu-EtAl:2016:P16-1,gulcehre-EtAl:2016:P16-1}, which helps copy sparse tokens (such as numbers and named entities) from the input.

Trained on a standard dataset (LDC2015E86), our model surpasses a strong sequence-to-sequence baseline by 2.3 BLEU points, demonstrating the advantage of graph-to-sequence models for AMR-to-text generation compared to sequence-to-sequence models.
Our final model achieves a BLEU score of 23.3 on the test set, which is 1.3 points higher than the  existing state of the art \cite{konstas-EtAl:2017:Long} trained on the same dataset.
When using gigaword sentences as additional training data, our model is consistently better than \citet{konstas-EtAl:2017:Long} using the same amount of gigaword data, showing the effectiveness of our model on large-scale training set.
We release our code and models at \url{https://github.com/freesunshine0316/neural-graph-to-seq-mp}.

\section{Baseline: a seq-to-seq model}
\label{sec:5_base}

Our baseline is a sequence-to-sequence model, which follows the encoder-decoder framework of \citet{konstas-EtAl:2017:Long}.

\subsection{Input representation}
\label{sec:5_base_inp}

Given an AMR graph $\boldsymbol{G}=(\boldsymbol{V},\boldsymbol{E})$, where $\boldsymbol{V}$ and $\boldsymbol{E}$ denote the sets of nodes and edges, respectively, we use the depth-first traversal of \citet{konstas-EtAl:2017:Long} to linearize it to obtain a sequence of tokens $v_1, \dots, v_N$, where $N$ is the number of tokens.
For example, the AMR graph in Figure 1 is serialized as ``describe :arg0 ( person :name ( name :op1 ryan )  )  :arg1 person :arg2 genius''.
We can see that the distance between ``describe'' and ``genius'', which are directly connected in the original AMR,  becomes 14 in the serialization result.

A simple way to calculate the representation for each token $v_j$ is using its word embedding $\boldsymbol{e}_j$:
\begin{equation}
\boldsymbol{x}_j = \boldsymbol{W}_1 \boldsymbol{e}_{j} + \boldsymbol{b}_1 \text{,}
\label{eq:5_base_inp}
\end{equation}
where $\boldsymbol{W}_1$ and $\boldsymbol{b}_1$ are model parameters for compressing the input vector size.
To alleviate the data sparsity problem and obtain better word representation as the input, we also adopt a forward LSTM over the characters of the token, and concatenate the last hidden state $\boldsymbol{h}_{j}^c$ with the word embedding:
\begin{equation}
\boldsymbol{x}_j = \boldsymbol{W}_1 \Big( [\boldsymbol{e}_{j}; \boldsymbol{h}_{j}^c] \Big) + \boldsymbol{b}_1
\label{eq:5_base_inp_2}
\end{equation}

\subsection{Encoder}

The encoder is a bi-directional LSTM applied on the linearized graph by depth-first traversal, as in \citet{konstas-EtAl:2017:Long}.
At each step $j$, the current states $\boldsymbol{\overleftarrow{h}}_j$ and $\boldsymbol{\overrightarrow{h}}_j$ are generated given the previous states $\boldsymbol{\overleftarrow{h}}_{j+1}$ and $\boldsymbol{\overrightarrow{h}}_{j-1}$ and the current input $\boldsymbol{x}_j$:
\begin{align}
\boldsymbol{\overleftarrow{h}}_j &= \text{LSTM}(\boldsymbol{\overleftarrow{h}}_{j+1}, \boldsymbol{x}_j) \\
\boldsymbol{\overrightarrow{h}}_j &= \text{LSTM}(\boldsymbol{\overrightarrow{h}}_{j-1}, \boldsymbol{x}_j)
\end{align}

\subsection{Decoder}
\label{sec:5_base_dec}

We use an attention-based LSTM decoder \cite{bahdanau2015neural}, where the attention memory ($\boldsymbol{A}$) is the concatenation of the attention vectors among all input words. 
Each attention vector $\boldsymbol{a}_j$ is the concatenation of the encoder states of an
input token in both directions ($\boldsymbol{\overleftarrow{h}}_j$ and $\boldsymbol{\overrightarrow{h}}_j$) and its input vector ($\boldsymbol{x}_j$):
\begin{align}
\boldsymbol{a}_j &= [\boldsymbol{\overleftarrow{h}}_j; \boldsymbol{\overrightarrow{h}}_j; \boldsymbol{x}_j] \\
\boldsymbol{A} &= [\boldsymbol{a}_1; \boldsymbol{a}_2; \dots; \boldsymbol{a}_N]
\end{align}
where $N$ is the number of input tokens.

The decoder yields an output sequence $w_1, w_2, \dots, w_M$ by calculating a sequence of hidden states $\boldsymbol{s}_1, \boldsymbol{s}_2 \dots, \boldsymbol{s}_M$ recurrently.
While generating the $m$-th word, the decoder considers five factors: 
(1) the attention memory $\boldsymbol{A}$; 
(2) the previous hidden state of the LSTM decoder $\boldsymbol{s}_{m-1}$; 
(3) the embedding of the current input word (previously generated word) $\boldsymbol{e}_{m}$; 
(4) the previous context vector $\boldsymbol{\mu}_{m-1}$, which is calculated by an attention mechanism (will be shown in the next paragraph) from $\boldsymbol{A}$; 
and (5) the previous coverage vector $\boldsymbol{\gamma}_{m-1}$, which is the accumulation of all attention distributions so far \cite{tu-EtAl:2016:P16-1}. 
When $t=1$, we initialize $\boldsymbol{\mu}_{0}$ and $\boldsymbol{\gamma}_{0}$ as zero vectors, set $\boldsymbol{e}_{1}$ to the embedding of the start token ``$<$s$>$'', and calculate $\boldsymbol{s}_{0}$ by averaging all encoder states.

For each time-step $m$, the decoder feeds the concatenation of the embedding of the current input $\boldsymbol{e}_{m}$ and the previous context vector $\boldsymbol{\mu}_{m-1}$ into the LSTM model to update its hidden state.
Then the attention probability $\alpha_{m,i}$ on the attention vector $\boldsymbol{a}_i \in A$ for the time-step is calculated as:
\begin{align}
\epsilon_{m,i} &= \boldsymbol{v}_2^\intercal \tanh(\boldsymbol{W}_a \boldsymbol{a}_i + \boldsymbol{W}_s \boldsymbol{s}_t + \boldsymbol{W}_{\gamma} \boldsymbol{\gamma}_{m-1} + \boldsymbol{b}_2) \\
\alpha_{m,i} &= \frac{\exp(\epsilon_{m,i})}{\sum_{j=1}^N\exp(\epsilon_{m,j})} 
\end{align}
where $\boldsymbol{W}_a$, $\boldsymbol{W}_s$, $\boldsymbol{W}_{\gamma}$, $\boldsymbol{v}_2$ and $\boldsymbol{b}_2$ are model parameters.
The coverage vector $\boldsymbol{\gamma}_m$ is updated by $\boldsymbol{\gamma}_m = \boldsymbol{\gamma}_{m-1} + \boldsymbol{\alpha}_m$, and the new context vector $\boldsymbol{\mu}_m$ is calculated via $\boldsymbol{\mu}_m = \sum_{i=1}^N \alpha_{m,i} \boldsymbol{a}_{i}$.
The output probability distribution over a vocabulary at the current state is calculated by:
\begin{equation}
\boldsymbol{p}_{vocab} = \text{softmax}(\boldsymbol{V}_3[\boldsymbol{s}_t,\boldsymbol{\mu}_t]+\boldsymbol{b}_3)\text{,}
\label{eq:5_pvocab}
\end{equation}
where $\boldsymbol{V}_3$ and $\boldsymbol{b}_3$ are model parameters, and the number of rows in $\boldsymbol{V}_3$ represents the number of words in the vocabulary.

\section{The graph-to-sequence model}

Unlike the baseline sequence-to-sequence model,
we leverage our recurrent graph network (GRN) to represent each input AMR, which directly models the graph structure without serialization. 

\subsection{The graph encoder}
\label{sec:5_encoder}

\begin{figure}
\centering
\includegraphics[width=0.75\linewidth]{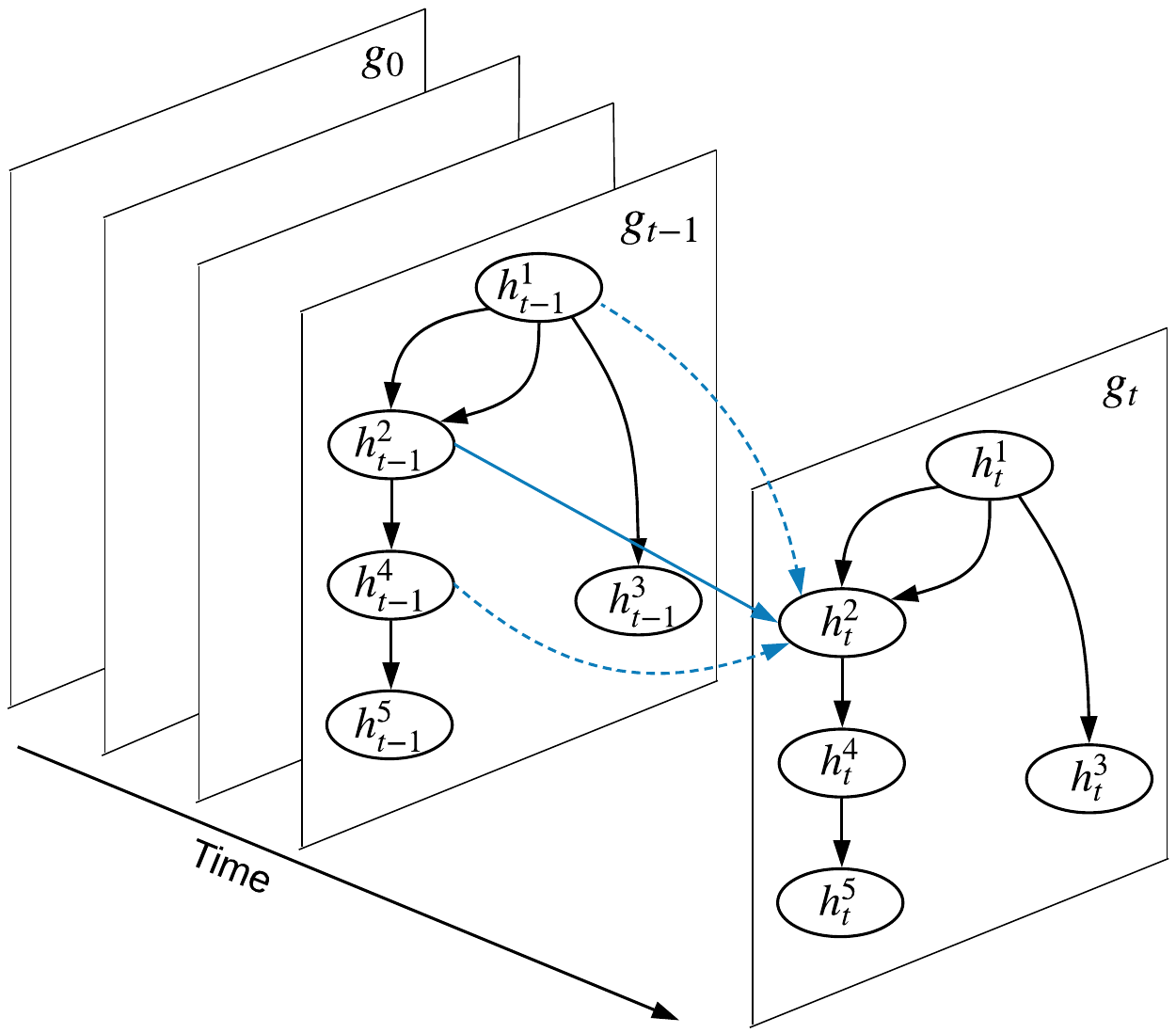}
\caption{GRN encoding for an AMR graph.}
\label{fig:5_encoder}
\end{figure}

Figure \ref{fig:5_encoder} shows the overall structure of our graph encoder. 
Formally, given an AMR graph $\boldsymbol{G}=(\boldsymbol{V}, \boldsymbol{E})$, we use a hidden state vector $\boldsymbol{h}^j$ to represent each node $v_j \in \boldsymbol{V}$. 
The state of the graph can thus be represented as:
\begin{equation}
\boldsymbol{g} = \{\boldsymbol{h}^j\}|_{v_j \in \boldsymbol{V}}
\end{equation}
Same as Chapter \ref{chap:nary}, our GRN-based graph encoder performs information exchange between nodes through a sequence of state transitions, leading to a sequence of states $\boldsymbol{g}_0, \boldsymbol{g}_1, \dots, \boldsymbol{g}_t, \dots$, where $\boldsymbol{g}_t = \{\boldsymbol{h}_t^j\}|_{v_j \in \boldsymbol{V}}$.
The initial state $\boldsymbol{g}_0$ consists of a set of node states $\boldsymbol{h}_0^j$ that contain all zeros.

The AMR graphs are similar with the dependency graphs (described in Chapter \ref{chap:nary}) in that both are directed and contain edge labels, so we simply adopt the GRN in Chapter \ref{chap:nary} as our AMR graph encoder.
Particularly, for node $v_j$, the inputs include representations of edges that are connected to it, where it can be either the source or the target of the edge.
We follow Chapter \ref{chap:nary} to define each edge as a triple $(i,j,l)$, where $i$ and $j$ are indices of the source and target nodes, respectively, and $l$ is the edge label.
$\boldsymbol{x}_{i,j}^l$ is the representation of edge $(i,j,l)$, detailed in Section \ref{sec:5_input}.
The inputs for $v_j$ are distinguished by incoming and outgoing edges, before being summed up:
\begin{equation}
\begin{split}
\boldsymbol{x}_j^{in} &= \sum_{(i,j,l)\in \boldsymbol{E}_{in}(j)} \boldsymbol{x}_{i,j}^l \\
\boldsymbol{x}_j^{out} &= \sum_{(j,k,l)\in \boldsymbol{E}_{out}(j)} \boldsymbol{x}_{j,k}^l \text{,} \\
\end{split}
\end{equation}
where $\boldsymbol{E}_{in}(j)$ and $\boldsymbol{E}_{out}(j)$ denote the sets of incoming and outgoing edges of $v_j$, respectively.
In addition to edge inputs, the encoder also considers the hidden states of its incoming nodes and outgoing nodes during a state transition. 
In particular, the states of all incoming nodes and outgoing nodes are summed up before being passed to the cell and gate nodes:
\begin{equation}
\begin{split}
\boldsymbol{h}_j^{in} &= \sum_{(i,j,l)\in \boldsymbol{E}_{in}(j)} \boldsymbol{h}_{t-1}^{i} \\
\boldsymbol{h}_j^{out} &= \sum_{(j,k,l)\in \boldsymbol{E}_{out}(j)} \boldsymbol{h}_{t-1}^{k} \text{,} \\
\end{split}
\end{equation} 
As the next step, the message $\boldsymbol{m}_t^j$ is aggregated by the concatenation:
\begin{equation}
    \boldsymbol{m}_t^j = [\boldsymbol{x}_j^{in}; \boldsymbol{x}_j^{out}; \boldsymbol{h}_j^{in}; \boldsymbol{h}_j^{out}]
\end{equation}
Then, it is applied with an LSTM step to update the node hidden state $\boldsymbol{h}_{t-1}^j$, the detailed equations are shown in Equation \ref{eq:3_grn_mp}.
\begin{equation}
    \boldsymbol{h}_t^j, \boldsymbol{c}_t^j = \text{LSTM}(\boldsymbol{m}_t^j, [\boldsymbol{h}_{t-1}^j, \boldsymbol{c}_{t-1}^j]) \text{,}
\end{equation}
where $\boldsymbol{c}^j$ is the cell memory for hidden state $\boldsymbol{h}^j$.

\subsection{Input Representation}
\label{sec:5_input}

Different from sequences, the edges of an AMR graph contain labels, which represent relations between the nodes they connect, and are thus important for modeling the graphs.
Similar with Section \ref{eq:5_base_inp_2}, we adopt two different ways for calculating the representation for each edge $(i,j,l)$: 
\begin{align}
\boldsymbol{x}_{i,j}^l &= \boldsymbol{W}_4 \Big( [\boldsymbol{e}_l; \boldsymbol{e}_i] \Big) + \boldsymbol{b}_4 \\
\boldsymbol{x}_{i,j}^l &= \boldsymbol{W}_4 \Big( [\boldsymbol{e}_l; \boldsymbol{e}_i; \boldsymbol{h}_i^c] \Big) + \boldsymbol{b}_4 \text{,}
\end{align}
where $\boldsymbol{e}_l$ and $\boldsymbol{e}_i$ are the embeddings of edge label $l$ and source node $v_i$, $\boldsymbol{h}_i^c$ denotes the last hidden state of the character LSTM over $v_i$, and $\boldsymbol{W}_4$ and $\boldsymbol{b}_4$ are trainable parameters.
The equations correspond to Equations \ref{eq:5_base_inp} and \ref{eq:5_base_inp_2} in Section \ref{sec:5_base_inp}, respectively.

\subsection{Decoder}

\begin{figure}
\centering
\includegraphics[scale=0.7]{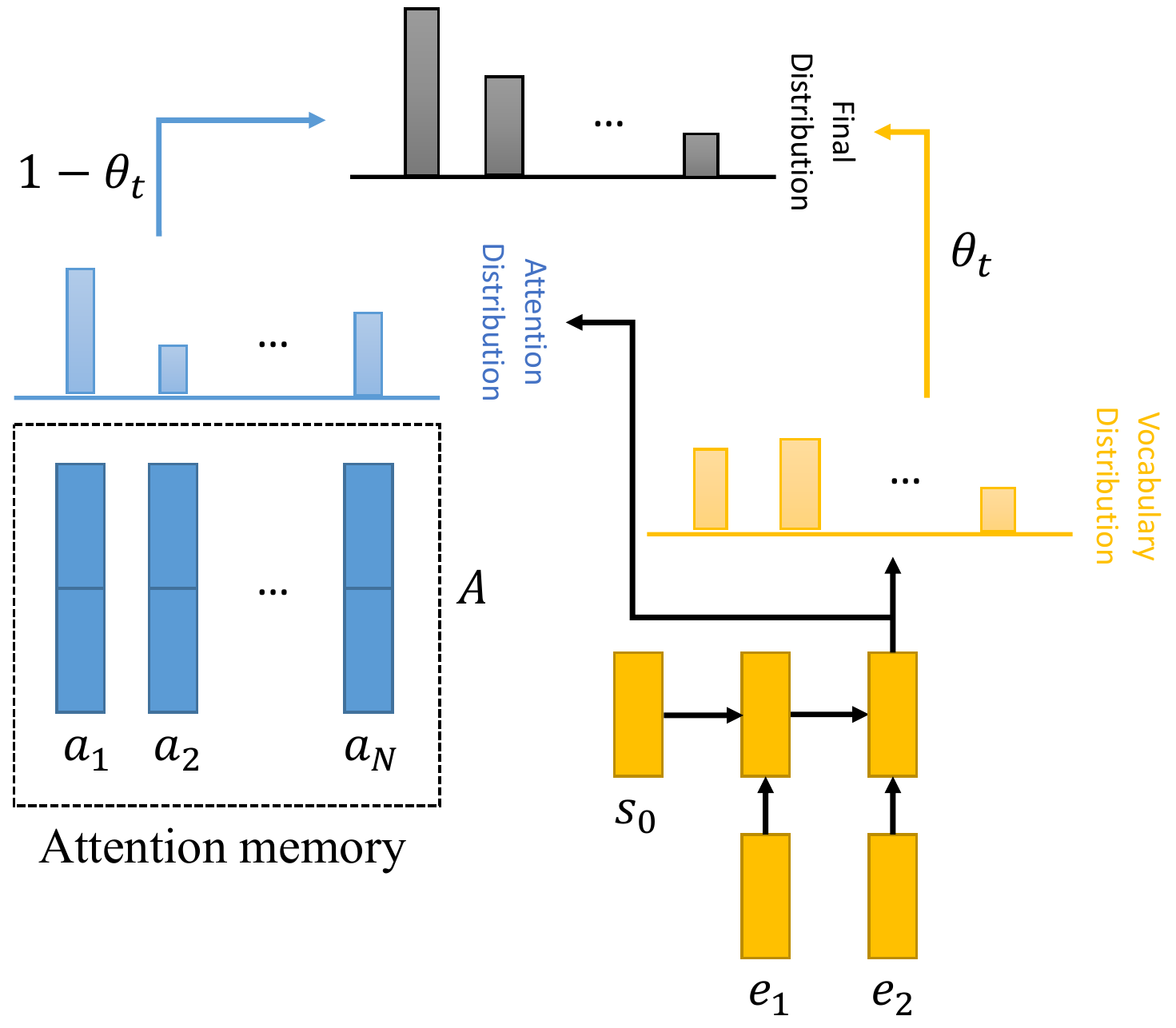}
\caption{The decoder with copy mechanism.}
\label{fig:5_decoder}
\end{figure}

As shown in Figure \ref{fig:5_decoder}, we adopt the attention-based LSTM decoder as described in Section \ref{sec:5_base_dec}.
Since our graph encoder generates a sequence of graph states, only the last graph state is adopted in the decoder.
In particular, we make the following changes to the decoder. First, each attention vector becomes $\boldsymbol{a}_j=[\boldsymbol{h}_T^j; \boldsymbol{x}_j]$, where $\boldsymbol{h}_T^j$ is the last state for node $v_j$. 
Second, the decoder initial state $\boldsymbol{s}_{-1}$ is the average of the last states of all nodes.

\subsection{Integrating the copy mechanism}
\label{sec:5_copy}

Open-class tokens, such as dates, numbers and named entities, account for a large portion in the AMR corpus. 
Most appear only a few times, resulting in a data sparsity problem.
To address this issue, \citet{konstas-EtAl:2017:Long} adopt anonymization for dealing with the data sparsity problem.
In particular, they first replace the subgraphs that represent dates, numbers and named entities (such as ``(q / quantity :quant 3)'' and ``(p / person :name (n / name :op1 ``Ryan''))'') with predefined placeholders (such as ``num\_0'' and ``person\_name\_0'') before decoding, and then recover the corresponding surface tokens (such as ``3'' and ``Ryan'') after decoding.
This method involves hand-crafted rules, which can be costly.

\subparagraph{Copy}
We find that most of the open-class tokens in a graph also appear in the corresponding sentence, and thus adopt the copy mechanism \cite{gulcehre-EtAl:2016:P16-1,gu-EtAl:2016:P16-1} to solve this problem.
The mechanism works on top of an attention-based RNN decoder by integrating the attention distribution into the final vocabulary distribution.
The final probability distribution is defined as the interpolation between two probability distributions:
\begin{equation}
\boldsymbol{p}_{final} = \theta_t \boldsymbol{p}_{vocab} + (1-\theta_t) \boldsymbol{p}_{attn}\text{,}
\end{equation}
where $\theta_t$ is a switch for controlling generating a word from the vocabulary or directly copying it from the input graph.
$\boldsymbol{p}_{vocab}$ is the probability distribution of directly generating the word, as defined in Equation \ref{eq:5_pvocab}, and 
$\boldsymbol{p}_{attn}$ is calculated based on the attention distribution $\boldsymbol{\alpha}_t$ by summing the probabilities of the graph nodes that contain identical concept.
Intuitively, $\theta_t$ is relevant to the current decoder input $\boldsymbol{e}_{t}$ and state $\boldsymbol{s}_t$, and the context vector $\boldsymbol{\mu}_t$.
Therefore, we define it as:
\begin{equation}
\theta_t = \sigma(\boldsymbol{w}_\mu^\intercal \boldsymbol{\mu}_t + \boldsymbol{w}_s^\intercal \boldsymbol{s}_t + \boldsymbol{w}_e^\intercal \boldsymbol{e}_{t} + b_5)\text{,}
\end{equation}
where vectors $\boldsymbol{w}_\mu$, $\boldsymbol{w}_s$, $\boldsymbol{w}_e$ and scalar $b_{5}$ are model parameters.
The copy mechanism favors generating words that appear in the input.
For AMR-to-text generation, it facilitates the generation of dates, numbers, and named entities that appear in AMR graphs.

\subparagraph{Copying vs anonymization}
Both copying and anonymization alleviate the data sparsity problem by handling the open-class tokens.
However, the copy mechanism has the following advantages over anonymization:
(1) anonymization requires significant manual work to define the placeholders and heuristic rules both from subgraphs to placeholders and from placeholders to the surface tokens, 
(2) the copy mechanism automatically learns what to copy, while anonymization relies on hard rules to cover all types of the open-class tokens, 
and (3) the copy mechanism is easier to adapt to new domains and languages than anonymization.

\section{Training and decoding}

We train our models using the cross-entropy loss over each gold-standard output sequence $\boldsymbol{W}^*=w_1^*, \dots, w_m^*, \dots, w_M^*$:
\begin{equation}
l = -\sum_{m=1}^M \log p(w_m^*|w_{m-1}^*,\dots,w_1^*,\boldsymbol{X};\boldsymbol{\theta})\text{,}
\end{equation}
where $\boldsymbol{X}$ is the input graph, and $\boldsymbol{\theta}$ is the model parameters.
Adam \cite{kingma2014adam} with a learning rate of 0.001 is used as the optimizer, and the model that yields the best devset performance is selected to evaluate on the test set.
Dropout with rate 0.1 is used during training.
Beam search with beam size to 5 is used for decoding.
Both training and decoding use Tesla K80 GPUs.

\section{Experiments}

\subsection{Data}

We use a standard AMR corpus (LDC2015E86) as our experimental dataset, which contains 16,833 instances for training, 1368 for development and 1371 for test. Each instance contains a sentence and an AMR graph.

Following \citet{konstas-EtAl:2017:Long}, we supplement the gold data with large-scale automatic data.
We take Gigaword as the external data to sample raw sentences, and train our model on both the sampled data and LDC2015E86.
We adopt \citet{konstas-EtAl:2017:Long}'s strategy for sampling sentences from Gigaword, and choose JAMR \cite{flanigan-EtAl:2016:SemEval} to parse selected sentences into AMRs, as the AMR parser of \citet{konstas-EtAl:2017:Long} only works on the anonymized data.
For training on both sampled data and LDC2015E86, we also follow the method of \citet{konstas-EtAl:2017:Long}, which is fine-tuning the model on the AMR corpus after every epoch of pretraining on the gigaword data.

\subsection{Settings}

We extract a vocabulary from the training set, which is shared by both the encoder and the decoder.
The word embeddings are initialized from Glove pretrained word embeddings \cite{pennington2014glove} on Common Crawl, and are not updated during training.
Following existing work, we evaluate the results with the BLEU metric \cite{papineni2002bleu}.

For model hyperparameters, we set the graph state transition number as 9 according to development experiments.
Each node takes information from at most 10 neighbors. 
The hidden vector sizes for both encoder and decoder are set to 300 (They are set to 600 for experiments using large-scale automatic data).
Both character embeddings and hidden layer sizes for character LSTMs are set 100, and at most 20 characters are taken for each graph node or linearized token.

\subsection{Development experiments}

\begin{table}
\centering
\begin{tabular}{l|c|c}
\hline
Model & BLEU & Time \\
\hline
\hline
Seq2seq & 18.8 & 35.4s \\
Seq2seq+copy & 19.9 & 37.4s \\
Seq2seq+charLSTM+copy & 20.6 & 39.7s \\
\hline
Graph2seq & 20.4 & 11.2s \\
Graph2seq+copy & 22.2 & 11.1s \\
Graph2seq+Anon & 22.1 & 9.2s \\
Graph2seq+charLSTM+copy & \textbf{22.8} & 16.3s \\
\hline
\end{tabular}
\caption{\textsc{Dev} BLEU scores and decoding times.}
\label{tab:5_dev_res}
\end{table}

As shown in Table \ref{tab:5_dev_res}, we compare our model with a set of baselines on the AMR devset to demonstrate how the graph encoder and the copy mechanism can be useful when training instances are not sufficient.
\emph{Seq2seq} is the sequence-to-sequence baseline described in Section \ref{sec:5_base}.
\emph{Seq2seq+copy} extends \emph{Seq2seq} with the copy mechanism, 
and \emph{Seq2seq+charLSTM+copy} further extends \emph{Seq2seq+copy} with character LSTM\@.
\emph{Graph2seq} is our graph-to-sequence model, \emph{Graph2seq+copy} extends \emph{Graph2seq} with the copy mechanism, and \emph{Graph2seq+charLSTM+copy} further extends \emph{Graph2seq+copy} with the character LSTM\@.
We also try \emph{Graph2seq+Anon}, which applies our graph-to-sequence model on the anonymized data from \citet{konstas-EtAl:2017:Long}.

\subparagraph{The graph encoder}
As can be seen from Table \ref{tab:5_dev_res}, the performance of \emph{Graph2seq} is 1.6 BLEU points higher than \emph{Seq2seq}, which shows that our graph encoder is effective when applied alone.
Adding the copy mechanism (\emph{Graph2seq+copy} vs \emph{Seq2seq+copy}), the gap becomes 2.3.
This shows that the graph encoder learns better node representations compared to the sequence encoder, which allows attention and copying to
function better.
Applying the graph encoder together with the copy mechanism gives a gain of 3.4 BLEU points over the baseline (\emph{Graph2seq+copy} vs \emph{Seq2seq}).
The graph encoder is consistently better than the sequence encoder no matter whether character LSTMs are used.

We also list the encoding part of decoding times on the devset, 
as the decoders of the seq2seq and the graph2seq models are similar,
so the time differences reflect efficiencies of the encoders.
Our graph encoder gives consistently better efficiency compared with the sequence encoder, showing the advantage of parallelization.

\subparagraph{The copy mechanism}
Table \ref{tab:5_dev_res} shows that the copy mechanism is effective on both the graph-to-sequence and the sequence-to-sequence models.
Anonymization gives comparable overall performance gains on our graph-to-sequence model as the copy mechanism (comparing \emph{Graph2seq+Anon} with \emph{Graph2seq+copy}).
However, the copy mechanism has several advantages over anonymization as discussed in Section \ref{sec:5_copy}.

\subparagraph{Character LSTM}
Character LSTM helps to increase the performances of both systems by roughly 0.6 BLEU points.
This is largely because it further alleviates the data sparsity problem by handling unseen words, which may share common substrings with in-vocabulary words.

\subsection{Effectiveness on graph state transitions}

We report a set of development experiments for understanding the graph LSTM encoder.

\begin{figure}[t]
\centering
\includegraphics[width=0.8\linewidth]{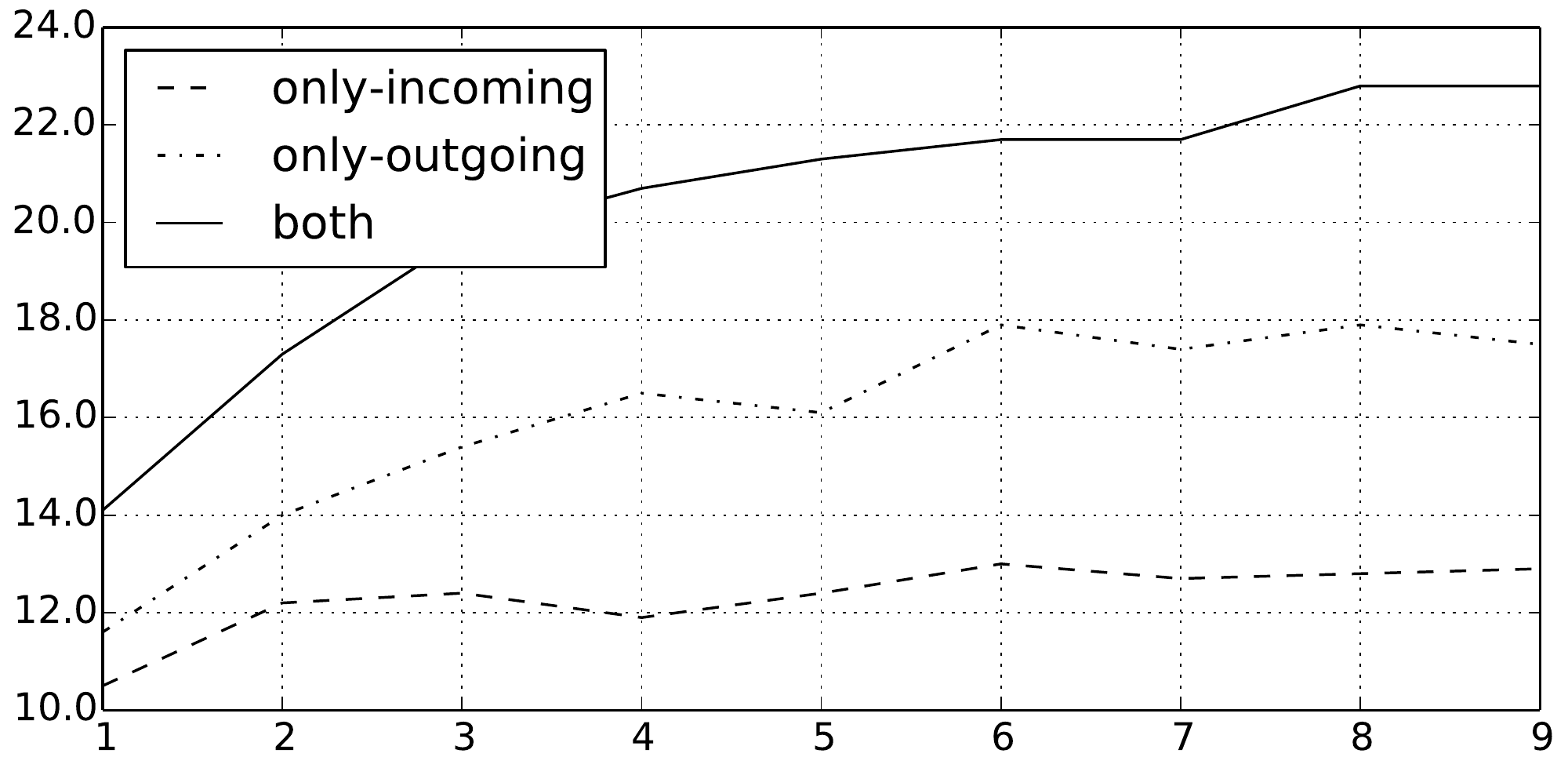}
\caption{\textsc{Dev} BLEU scores against transition steps for the graph encoder.}
\label{fig:5_iters}
\end{figure}

\begin{figure}[t]
\centering
\includegraphics[width=0.7\linewidth]{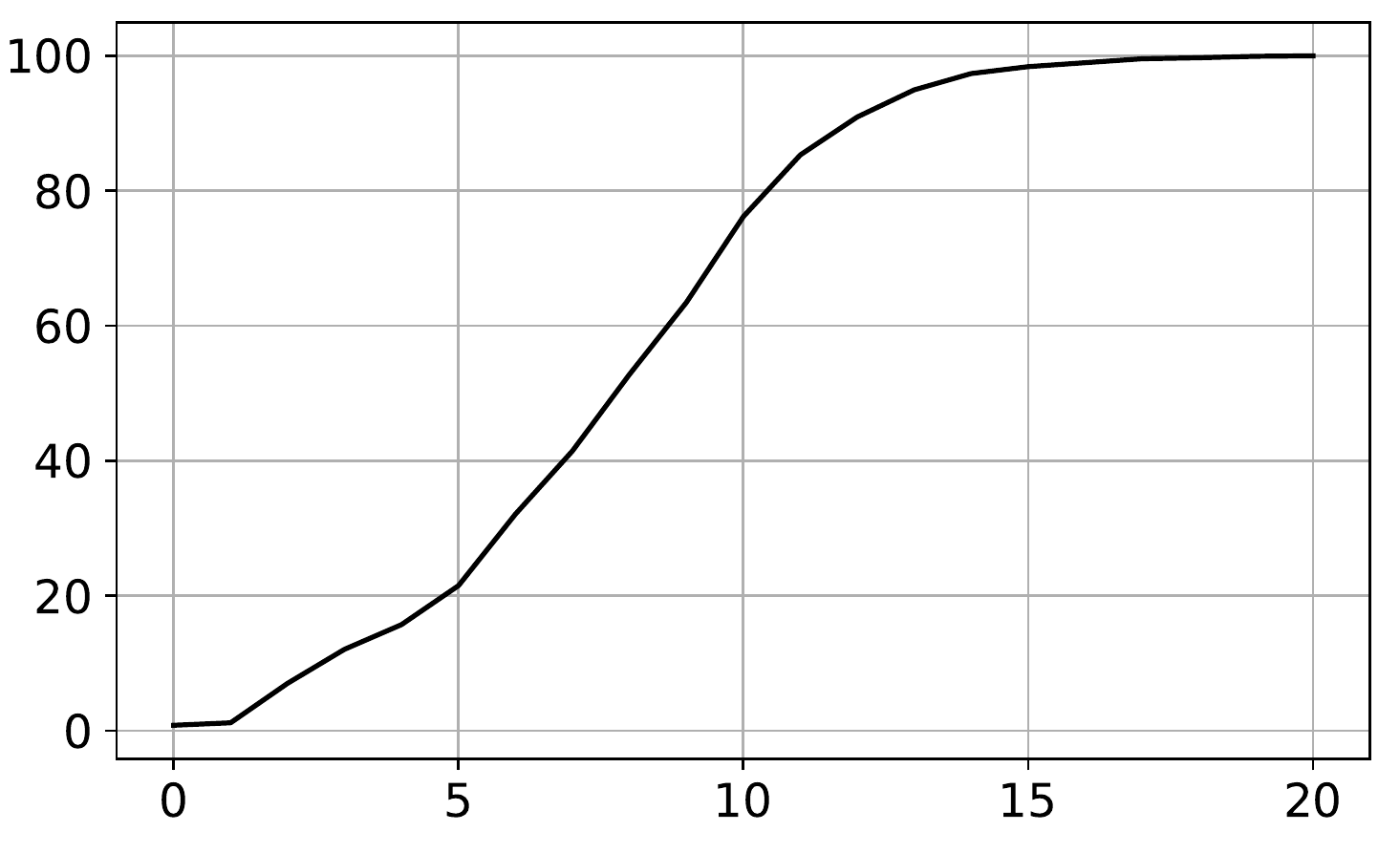}
\caption{Percentage of \textsc{Dev} AMRs with different diameters.}
\label{fig:5_depth}
\end{figure}

\subparagraph{Number of iterations}
We analyze the influence of the number of state transitions to the model performance on the devset.
Figure \ref{fig:5_iters} shows the BLEU scores of different state transition numbers, when both incoming and outgoing edges are taken for calculating the next state (as shown in Figure \ref{fig:5_encoder}).
The system is \emph{Graph2seq+charLSTM+copy}.
Executing only 1 iteration results in a poor BLEU score of 14.1.
In this case the state for each node only contains information about immediately adjacent nodes.
The performance goes up dramatically to 21.5 when increasing the iteration number to 5.
In this case, the state for each node contains information of all nodes within a distance of 5.
The performance further goes up to 22.8 when increasing the iteration number from 5 to 9,
where all nodes with a distance of less than 10 are incorporated in the state for each node.

\subparagraph{Graph diameter}
We analyze the percentage of the AMR graphs in the devset with different graph diameters and show the cumulative distribution in Figure \ref{fig:5_depth}.
The diameter of an AMR graph is defined as the longest distance between two AMR nodes.\footnote{The diameter of single-node graphs is 0.}
Even though the diameters for less than 80\% of the AMR graphs are less or equal than 10, our development experiments show that it is not necessary to incorporate the whole-graph information for each node.
Further increasing state transition number may lead to additional improvement. 
We do not perform exhaustive search for finding the optimal state transition number.

\subparagraph{Incoming and outgoing edges}
As shown in Figure \ref{fig:5_iters}, we analyze the efficiency of state transition when only incoming or outgoing edges are used.
From the results, we can see that there is a huge drop when state transition is performed only with incoming or outgoing edges.
Using edges of one direction, the node states only contain information of ancestors or descendants.
On the other hand, node states contain information of ancestors, descendants, and siblings if edges of both directions are used.
From the results, we can conclude that not only the ancestors and descendants, but also the siblings are important for modeling the AMR graphs.
This is similar to observations on syntactic parsing tasks \cite{mcdonald-crammer-pereira:2005:ACL}, where sibling features are adopted.

We perform a similar experiment for the \emph{Seq2seq+copy} baseline by only executing single-directional LSTM for the encoder.
We observe BLEU scores of 11.8 and 12.7 using only forward or backward LSTM, respectively.
This is consistent with our graph model in that execution using only one direction leads to a huge performance drop. 
The contrast is also reminiscent of using the normal input versus the reversed input in neural machine translation \citep{sutskever2014sequence}.

\subsection{Results}

\begin{table} 
\centering
\begin{tabular}{l|c}
\hline
Model & \textsc{Bleu} \\
\hline
\hline
PBMT  & 26.9 \\
SNRG  & 25.6 \\
Tree2Str  & 23.0 \\
MSeq2seq+Anon  & 22.0 \\
Graph2seq+copy & 22.7 \\
Graph2seq+charLSTM+copy & 23.3 \\
\hline
MSeq2seq+Anon (200K) & 27.4 \\
MSeq2seq+Anon (2M)  & 32.3 \\
MSeq2seq+Anon (20M)  & \textbf{33.8} \\
\hline
Seq2seq+charLSTM+copy (200K) & 27.4 \\
Seq2seq+charLSTM+copy (2M) & 31.7 \\
Graph2seq+charLSTM+copy (200K) & 28.2 \\
Graph2seq+charLSTM+copy (2M) & \textbf{33.6}\tablefootnote{It was 33.0 at submission, and has been improved.} \\
\hline
\end{tabular}
\caption{\textsc{Test} results. ``(200K)'', ``(2M)'' and ``(20M)'' represent training with the corresponding number of additional sentences from Gigaword.}
\label{tab:5_global_res}
\end{table}

Table \ref{tab:5_global_res} compares our final results with existing work.
\emph{MSeq2seq+Anon} \cite{konstas-EtAl:2017:Long} is an attentional multi-layer sequence-to-sequence model trained with the anonymized data.
\emph{PBMT} \cite{pourdamghani-knight-hermjakob:2016:INLG} adopts a phrase-based model for machine translation \cite{koehn2003statistical} on the input of linearized AMR graph, \emph{SNRG} \cite{song-acl17} uses synchronous node replacement grammar for parsing the AMR graph while generating the text, and \emph{Tree2Str} \cite{jeff2016amrgen} converts AMR graphs into trees by splitting the re-entrances before using a tree transducer to generate the results.

\emph{Graph2seq+charLSTM+copy} achieves a BLEU score of 23.3, which is 1.3 points better than \emph{MSeq2seq+Anon} trained on the same AMR corpus.
In addition, our model without character LSTM is still 0.7 BLEU points higher than \emph{MSeq2seq+Anon}.
Note that \emph{MSeq2seq+Anon} relies on anonymization, which requires additional manual work for defining mapping rules, thus limiting its usability on other languages and domains.
The neural models tend to underperform statistical models when trained on limited (16K) gold data, but performs better with scaled silver data \cite{konstas-EtAl:2017:Long}.

Following \citet{konstas-EtAl:2017:Long}, we also evaluate our model using both the AMR corpus and sampled sentences from Gigaword.
Using additional 200K or 2M gigaword sentences, \emph{Graph2seq+charLSTM+copy} achieves BLEU scores of 28.2 and 33.0, respectively, which are 0.8 and 0.7 BLEU points better than \emph{MSeq2seq+Anon} using the same amount of data, respectively.
The BLEU scores are 5.3 and 10.1 points better than the result when it is only trained with the AMR corpus, respectively.
This shows that our model can benefit from scaled data with automatically generated AMR graphs, and it is more effective than \emph{MSeq2seq+Anon} using the same amount of data.
Using 2M gigaword data, our model is better than all existing methods.
\citet{konstas-EtAl:2017:Long} also experimented with 20M external data, obtaining a BLEU of 33.8.
We did not try this setting due to hardware limitations.
The \emph{Seq2seq+charLSTM+copy} baseline trained on the large-scale data is close to \emph{MSeq2seq+Anon} using the same amount of training data, yet is much worse than our model.

\subsection{Case study}

We conduct case studies for better understanding the model performances.
Table \ref{tab:5_examples} shows example outputs of sequence-to-sequence  (\emph{S2S}), graph-to-sequence (\emph{G2S}) and graph-to-sequence with copy mechanism (\emph{G2S+CP}). 
\emph{Ref} denotes the reference output sentence, and \emph{Lin} shows the serialization results of input AMRs.
The best hyperparameter configuration is chosen for each model.

For the first example, \emph{S2S} fails to recognize the concept ``a / account'' as a noun and loses the concept ``o / old'' (both are underlined).
The fact that ``a / account'' is a noun is implied by ``a~/~account :mod (o~/~old)'' in the original AMR graph.
Though directly connected in the original graph, their distance in the serialization result (the input of \emph{S2S}) is 26, which 
may be why \emph{S2S} makes these mistakes.
In contrast, \emph{G2S} handles ``a~/~account'' and ``o~/~old'' correctly.
In addition, the copy mechanism helps to copy ``look-over'' from the input, which rarely appears in the training set.
In this case, \emph{G2S+CP} is incorrect only on hyphens and literal reference to ``anti-japanese war'', although the meaning is fully understandable.

For the second case, both \emph{G2S} and \emph{G2S+CP} correctly generate the noun ``agreement'' for ``a~/ agree'' in the input AMR, while \emph{S2S} fails to.
The fact that ``a~/~agree'' represents a noun can be determined by the original graph segment ``p / provide :ARG0 (a / agree)'', which indicates that ``a / agree'' is the subject of ``p / provide''.
In the serialization output, the two nodes are close to each other.
Nevertheless, \emph{S2S} still failed to capture this structural relation, which reflects the fact that a sequence encoder is not designed to explicitly model hierarchical information encoded in the serialized graph.
In the training instances, serialized nodes that are close to each other can originate from neighboring graph nodes, or distant graph nodes, which prevents the decoder from confidently deciding the correct relation between them.
In contrast, \emph{G2S} sends the node ``p / provide'' simultaneously with relation ``ARG0'' when calculating hidden states for ``a / agree'', which facilitates the yielding of ``the agreement provides''.

\begin{table}[t!] \footnotesize
\centering
\begin{tabularx}{\textwidth}{X}
\hline
(p / possible-01 :polarity - \\
~~~~:ARG1 (l / look-over-06 \\
~~~~~~~~:ARG0 (w / we) \\
~~~~~~~~:ARG1 (a / \underline{account}-01 \\
~~~~~~~~~~~~:ARG1 (w2 / war-01 \\
~~~~~~~~~~~~~~~~:ARG1 (c2 / country \\
~~~~~~~~~~~~~~~~~~~~:wiki ``Japan'' \\
~~~~~~~~~~~~~~~~~~~~:name (n2 / name :op1 ``Japan'')) \\
~~~~~~~~~~~~~~~~:time (p2 / previous) \\
~~~~~~~~~~~~~~~~:ARG1-of (c / call-01 \\
~~~~~~~~~~~~~~~~~~~~:mod (s / so))) \\
~~~~~~~~~~~~:mod (o / \underline{old})))) \\
\textbf{Lin}: possible :polarity - :arg1 ( look-over :arg0 we :arg1 ( \underline{account} :arg1 ( war :arg1 ( country :wiki japan :name ( name :op1 japan ) ) :time previous :arg1-of ( call :mod so ) ) :mod \underline{old} ) ) \\
\textbf{Ref}: we can n't look over the old accounts of the previous so-called anti-japanese war . \\
\textbf{S2S}: we can n't be able to account the past drawn out of japan 's entire war .\\
\textbf{G2S}: we can n't be able to do old accounts of the previous and so called japan war.\\
\textbf{G2S+CP}: we can n't look-over the old accounts of the previous so called war on japan . \\
\hline
(p / provide-01 \\
~~~~:ARG0 (a / \underline{agree}-01) \\
~~~~:ARG1 (a2 / and \\
~~~~~~~~:op1 (s / staff \\
~~~~~~~~~~~~:prep-for (c / center \\
~~~~~~~~~~~~~~~~:mod (r / research-01))) \\
~~~~~~~~:op2 (f / fund-01 \\
~~~~~~~~~~~~:prep-for c))) \\
\textbf{Lin}: provide :arg0 \underline{agree} :arg1 ( and :op1 ( staff :prep-for ( center :mod research ) ) :op2 ( fund :prep-for center ) ) \\
\textbf{Ref}: the agreement will provide staff and funding for the research center .\\
\textbf{S2S}: agreed to provide research and institutes in the center .\\
\textbf{G2S}: the agreement provides the staff of research centers and funding . \\
\textbf{G2S+CP}: the agreement provides the staff of the research center and the funding .\\
\hline
\end{tabularx}
\caption{Example system outputs.}
\label{tab:5_examples}
\end{table}

\section{Related work}

Among early statistical methods for AMR-to-text generation, \citet{jeff2016amrgen} convert input graphs to trees by splitting re-entrances, and then translate the trees into sentences 
with a tree-to-string transducer.
\citet{song-acl17} use a synchronous node replacement grammar to parse input AMRs and generate sentences at the same time.
\citet{pourdamghani-knight-hermjakob:2016:INLG} linearize input graphs by breadth-first traversal, and then use a phrase-based machine translation system\footnote{http://www.statmt.org/moses/} to generate results by translating linearized sequences.
Apparently, our neural graph-to-sequence model are significantly different from the previous work.
With addition automatic data, our model shows dramatic improvement over those methods, showing its effectiveness on benefiting from large-scale training.

In addition to NMT \cite{gulcehre-EtAl:2016:P16-1}, the copy mechanism has been shown effective on tasks such as dialogue \cite{gu-EtAl:2016:P16-1}, summarization \cite{see-liu-manning:2017:Long} and question generation \citep{song-naacl18}.
We investigate the copy mechanism on AMR-to-text generation.

\section{Conclusion}

We introduced a novel graph-to-sequence model for AMR-to-text generation.
Compared to sequence-to-sequence models, which require linearization of AMR before decoding, a graph LSTM is leveraged to directly model full AMR structure.
Allowing high parallelization, the graph encoder is more efficient than the sequence encoder.
In our experiments, the graph model outperforms a strong sequence-to-sequence model, achieving the best performance.

%% file: 6-chap-semnmt.tex
It is intuitive that semantic representations can be useful for machine translation, mainly because they can help in enforcing meaning preservation and handling data sparsity (many sentences correspond to one meaning) of machine translation models.
On the other hand, little work has been done on leveraging semantics for neural machine translation (NMT).
In this work, we study the usefulness of AMR (short for abstract meaning representation) on NMT.
Experiments on a standard English-to-German dataset show that incorporating AMR as additional knowledge can significantly improve a strong attention-based sequence-to-sequence neural translation model.

\section{Introduction}

It is intuitive that semantic representations ought to
be relevant to machine translation, given that the
task is to produce a target language sentence with the
same meaning as the source language input.
Semantic representations formed the core of the earliest symbolic
machine translation systems, and have been applied
to statistical but non-neural systems as well.

\begin{figure}
\centering
\includegraphics[width=0.7\linewidth]{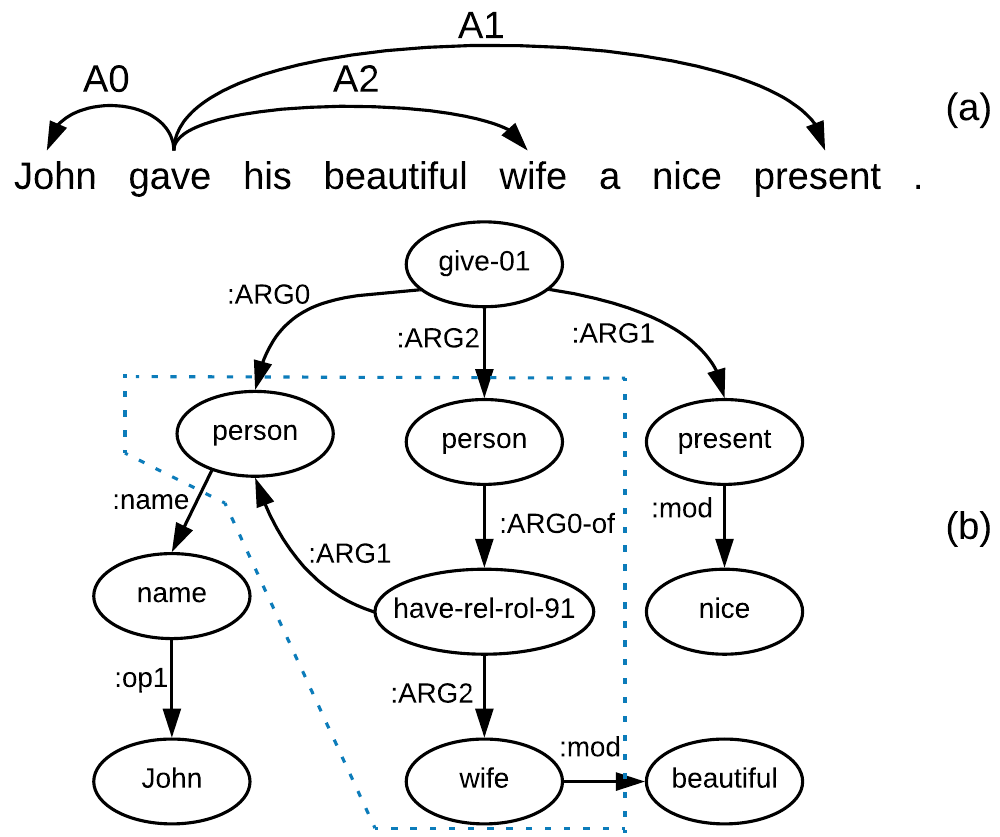}
\caption{(a) A sentence with semantic roles annotations, (b) the corresponding AMR graph of that sentence.}
\label{fig:6_example}
\end{figure}

Leveraging syntax for neural machine translation (NMT) has been an active research topic \cite{stahlberg-EtAl:2016:P16-2,aharoni-goldberg:2017:Short,li-EtAl:2017:Long,chen-EtAl:2017:Long6,bastings-EtAl:2017:EMNLP2017,wu2017improved,chen2017syntax}.
On the other hand, exploring semantics for NMT has so far received relatively little attention.
Recently, \citet{diego-EtAl:2018:PAPERS} exploited semantic role labeling (SRL) for NMT, showing that the predicate-argument information from SRL can improve the performance of an attention-based sequence-to-sequence model by alleviating the ``argument switching'' problem,\footnote{flipping arguments corresponding to different roles} one frequent and severe issue faced by NMT systems \cite{isabelle-cherry-foster:2017:EMNLP2017}.
Figure \ref{fig:6_example} (a) shows one example of semantic role information, which only captures the relations between a predicate ({\em gave}) and its arguments ({\em John}, {\em wife} and {\em present}).
Other important information, such as the relation between {\em John} and {\em wife}, can not be incorporated.

In this paper, we explore the usefulness of abstract meaning representation (AMR) \cite{banarescu-EtAl:2013:LAW7-ID} as a semantic representation for NMT\@.
AMR is a semantic formalism that encodes the meaning of a sentence as a rooted, directed graph.
Figure \ref{fig:6_example} (b) shows an AMR graph, in which the nodes (such as {\em give-01} and {\em John}) represent the concepts,
and edges (such as {\em :ARG0} and {\em :ARG1}) represent the relations between concepts they connect.
Comparing with semantic roles, AMRs capture more relations, such as the relation between {\em John} and {\em wife} (represented by the subgraph within dotted lines).
In addition, AMRs directly capture entity relations and abstract away inflections and function words.
As a result, they can serve as a source of knowledge for machine translation that is orthogonal to the textual input.
Furthermore, structural information from AMR graphs can help reduce data sparsity, when training data is not sufficient for large-scale training.

Recent advances in AMR parsing keep pushing the boundary of state-of-the-art performance \cite{flanigan2014discriminative,artzi-lee-zettlemoyer:2015:EMNLP,pust-EtAl:2015:EMNLP,peng2015conll,flanigan-EtAl:2016:SemEval,buys-blunsom:2017:Long,konstas-EtAl:2017:Long,wang-xue:2017:EMNLP2017,P18-1037,peng-acl18,P18-1170,D18-1198}, and have made it possible for automatically-generated AMRs to benefit down-stream tasks, such as question answering \cite{mitra2015addressing}, summarization \cite{takase-EtAl:2016:EMNLP2016}, and event detection \cite{li-EtAl:2015:CNewsStory}.
However, to our knowledge, no existing work has exploited AMR for enhancing NMT.

We fill in this gap, taking an attention-based sequence-to-sequence system as our baseline, which is similar to \citet{bahdanau2015neural}.
To leverage knowledge within an AMR graph, we adopt a graph recurrent network (GRN) \cite{P18-1030,P18-1150} as the AMR encoder.
In particular, a full AMR graph is considered as a single state, with nodes in the graph being its substates. 
State transitions are performed on the graph recurrently, allowing substates to exchange information through edges. 
At each recurrent step, each node advances its current state by receiving information from the current states of its adjacent nodes. 
Thus, with increasing numbers of recurrent steps, each word receives information from a larger context. Figure~\ref{fig:6_encoder} shows the recurrent transition, where each node works simultaneously.
Compared with other methods for encoding AMRs \cite{konstas-EtAl:2017:Long}, GRN keeps the original graph structure,
and thus no information is lost.
For the decoding stage, two separate attention mechanisms are adopted in the AMR encoder and sequential encoder, respectively.

Experiments on WMT16 English-German data (4.17M) show that adopting AMR significantly improves a strong attention-based sequence-to-sequence baseline (25.5 vs 23.7 BLEU).
When trained with small-scale (226K) data, the improvement increases (19.2 vs 16.0 BLEU), which shows that the structural information from AMR can alleviate data sparsity when training data are not sufficient.
To our knowledge, we are the first to investigate AMR for NMT.

Our code and parallel data with automatically parsed AMRs are available at \url{https://github.com/freesunshine0316/semantic-nmt.}

\section{Related work}

Most previous work on exploring semantics for statistical machine translation (SMT) studies the usefulness of predicate-argument structure from semantic role labeling \cite{wong-mooney:2006:HLT-NAACL06-Main,wu-fung:2009:NAACLHLT09-Short,liu-gildea:2010:PAPERS,baker2012modality}.
\citet{jones-EtAl:2012:PAPERS} first convert Prolog expressions into graphical meaning representations, leveraging synchronous hyperedge replacement grammar to parse the input graphs while generating the outputs.
Their graphical meaning representation is different from AMR under a strict definition, and their experimental data are limited to 880 sentences.
We are the first to investigate AMR on large-scale machine translation.

Recently, \citet{diego-EtAl:2018:PAPERS} investigate semantic role labeling (SRL) on neural machine translation (NMT).
The predicate-argument structures are encoded via graph convolutional network (GCN)  layers \cite{kipf2017semi}, which are laid on top of regular BiRNN or CNN layers. 
Our work is in line with exploring semantic information, but different in exploiting AMR rather than SRL for NMT\@.
In addition, we leverage a graph recurrent network (GRN) \cite{P18-1030,P18-1150} for modeling AMRs rather than GCN, which is formally consistent with the RNN sentence encoder.
Since there is no one-to-one correspondence between AMR nodes and source words, we adopt a doubly-attentive LSTM decoder, which is another major difference from \citet{diego-EtAl:2018:PAPERS}.



\section{Baseline: attention-based BiLSTM}
\label{sec:6_base}

We take the attention-based sequence-to-sequence model of \citet{bahdanau2015neural} as the baseline, but use LSTM cells \cite{hochreiter1997long} instead of GRU cells \cite{cho-EtAl:2014:EMNLP2014}.

\subsection{BiLSTM encoder}
\label{sec:6_base_enc}

The encoder is a bi-directional LSTM on the source side.
Given a sentence, two sequences of states $[\overleftarrow{\boldsymbol{h}}_1, \overleftarrow{\boldsymbol{h}}_2, \dots, \overleftarrow{\boldsymbol{h}}_N]$ and $[\overrightarrow{\boldsymbol{h}}_1, \overrightarrow{\boldsymbol{h}}_2, \dots \overrightarrow{\boldsymbol{h}}_N]$ are generated for representing the input word sequence $x_1, x_2, \dots, x_N$ in the right-to-left and left-to-right directions, respectively, where for each word $x_i$, 
\begin{align}
\overleftarrow{\boldsymbol{h}}_i &= \textrm{LSTM}(\overleftarrow{\boldsymbol{h}}_{i+1}, \boldsymbol{e}_{x_i}) \\
\overrightarrow{\boldsymbol{h}}_i &= \textrm{LSTM}(\overrightarrow{\boldsymbol{h}}_{i-1}, \boldsymbol{e}_{x_i})
\end{align}
$\boldsymbol{e}_{x_i}$ is the embedding of word $x_i$.

\subsection{Attention-based decoder}
\label{sec:6_base_dec}

The decoder yields a word sequence in the target language  $y_1, y_2, \dots, y_M$ by calculating a sequence of hidden states $\boldsymbol{s}_1, \boldsymbol{s}_2 \dots, \boldsymbol{s}_M$ recurrently.
We use an attention-based LSTM decoder \cite{bahdanau2015neural}, where the attention memory ($\boldsymbol{H}$) is the concatenation of the attention vectors among all source words. 
Each attention vector $\boldsymbol{h}_i$ is the concatenation of the encoder states of an input token in both directions ($\overleftarrow{\boldsymbol{h}}_i$ and $\overrightarrow{\boldsymbol{h}}_i$):
\begin{align}
\boldsymbol{h}_i &= [\overleftarrow{\boldsymbol{h}}_i; \overrightarrow{\boldsymbol{h}}_i] \\
\boldsymbol{H} &= [\boldsymbol{h}_1; \boldsymbol{h}_2; \dots; \boldsymbol{h}_N]
\end{align}
$N$ is the number of source words.

While generating the $m$-th word, the decoder considers four factors: 
(1) the attention memory $\boldsymbol{H}$; 
(2) the previous hidden state of the LSTM model $\boldsymbol{s}_{m-1}$; 
(3) the embedding of the current input (previously generated word) $\boldsymbol{e}_{y_m}$; 
and (4) the previous context vector $\boldsymbol{\zeta}_{m-1}$ from attention memory $\boldsymbol{H}$. 
When $m=1$, we initialize $\boldsymbol{\zeta}_{0}$ as a zero vector, set $\boldsymbol{e}_{y_1}$ to the embedding of sentence start token ``$<$s$>$'', and calculate $\boldsymbol{s}_{0}$ from the last step of the encoder states via a dense layer:
\begin{equation}
\boldsymbol{s}_{0} = \boldsymbol{W}_1 [\overleftarrow{\boldsymbol{h}}_0; \overrightarrow{\boldsymbol{h}}_N] + \boldsymbol{b}_1
\end{equation}
where $\boldsymbol{W}_1$ and $\boldsymbol{b}_1$ are model parameters.

For each decoding step $m$, the decoder feeds the concatenation of the embedding of the current input $\boldsymbol{e}_{y_m}$ and the previous context vector $\boldsymbol{\zeta}_{m-1}$ into the LSTM model to update its hidden state:
\begin{equation}
\boldsymbol{s}_m = \textrm{LSTM}(\boldsymbol{s}_{m-1}, [\boldsymbol{e}_{y_m};\boldsymbol{\zeta}_{m-1}])
\end{equation}
Then the attention probability $\alpha_{m,i}$ on the attention vector $\boldsymbol{h}_i \in \boldsymbol{H}$ for the current decode step is calculated as:
\begin{align}
\epsilon_{m,i} &= \boldsymbol{v}_2^\intercal \tanh(\boldsymbol{W}_h \boldsymbol{h}_i + \boldsymbol{W}_s \boldsymbol{s}_m + \boldsymbol{b}_2) \\
\alpha_{m,i} &= \frac{\exp(\epsilon_{m,i})}{\sum_{j=1}^N\exp(\epsilon_{m,j})} 
\end{align}
$\boldsymbol{W}_h$, $\boldsymbol{W}_s$, $\boldsymbol{v}_2$ and $\boldsymbol{b}_2$ are model parameters.
The new context vector $\boldsymbol{\zeta}_m$ is calculated via 
\begin{equation}
\boldsymbol{\zeta}_m = \sum_{i=1}^N \alpha_{m,i} \boldsymbol{h}_{i}
\end{equation}
The output probability distribution over the target vocabulary at the current state is calculated by:
\begin{equation}
\boldsymbol{p}_{vocab} = \textrm{softmax}(\boldsymbol{V}_3[\boldsymbol{s}_m,\boldsymbol{\zeta}_m]+\boldsymbol{b}_3)\textrm{,}
\label{eq:6_pvocab}
\end{equation}
where $\boldsymbol{V}_3$ and $\boldsymbol{b}_3$ are learnable parameters.

\section{Incorporating AMR}
\label{sec:amr}

\begin{figure}
\centering
\includegraphics[width=0.85\linewidth]{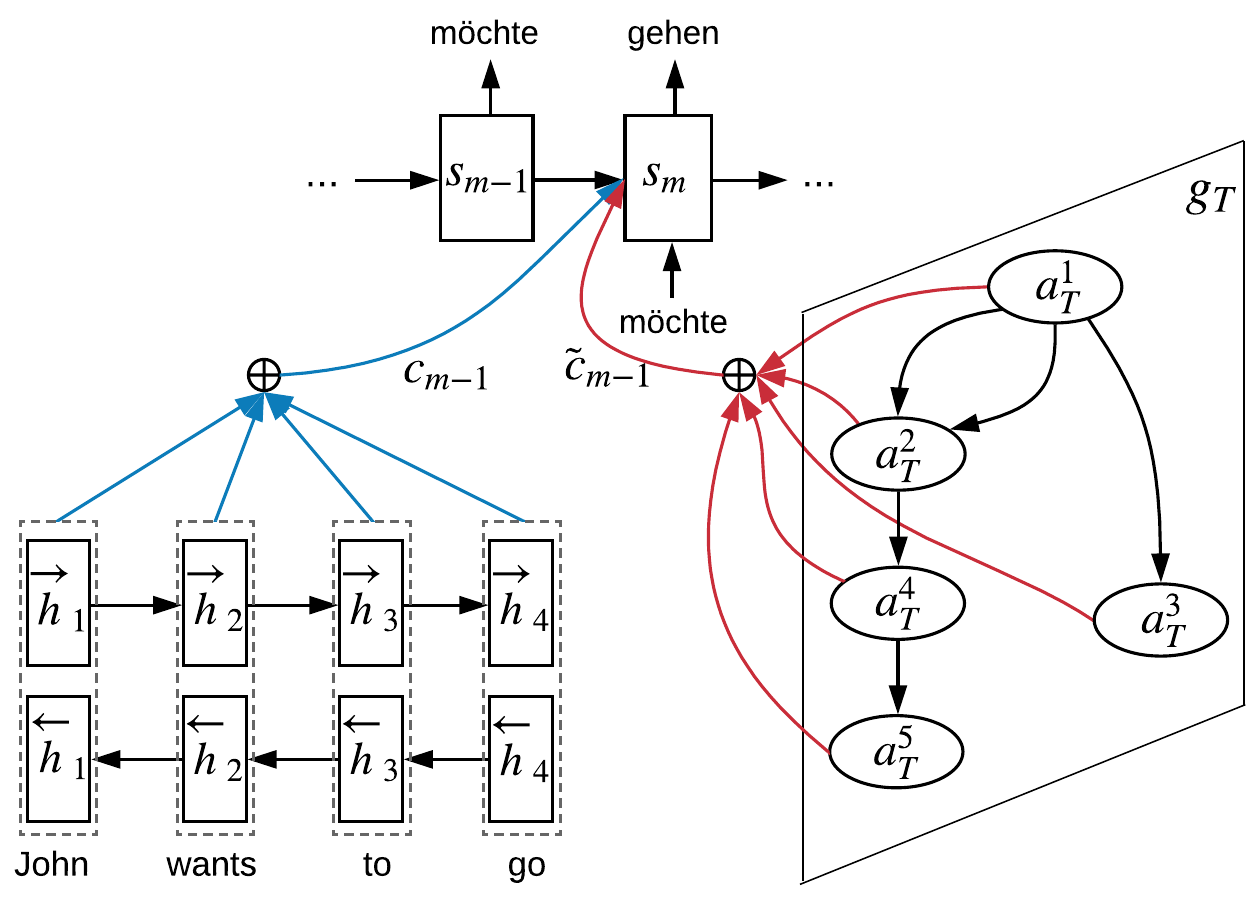}
\caption{Overall architecture of our model.}
\label{fig:6_dual2seq}
\end{figure}

Figure \ref{fig:6_dual2seq} shows the overall architecture of our model, which adopts a BiLSTM (bottom left) and our graph recurrent network (GRN)\footnote{We show the advantage of our graph encoder by comparing with another popular way for encoding AMRs in Section \ref{sec:main_res}.} (bottom right) for encoding the source sentence and AMR, respectively.
An attention-based LSTM decoder is used to generate the output sequence in the target language, with attention models over both the sequential encoder and the graph encoder.
The attention memory for the graph encoder is from the last step of the graph state transition process, which is shown in Figure \ref{fig:6_encoder}.

\subsection{Encoding AMR with GRN}

\begin{figure}
\centering
\includegraphics[width=0.75\linewidth]{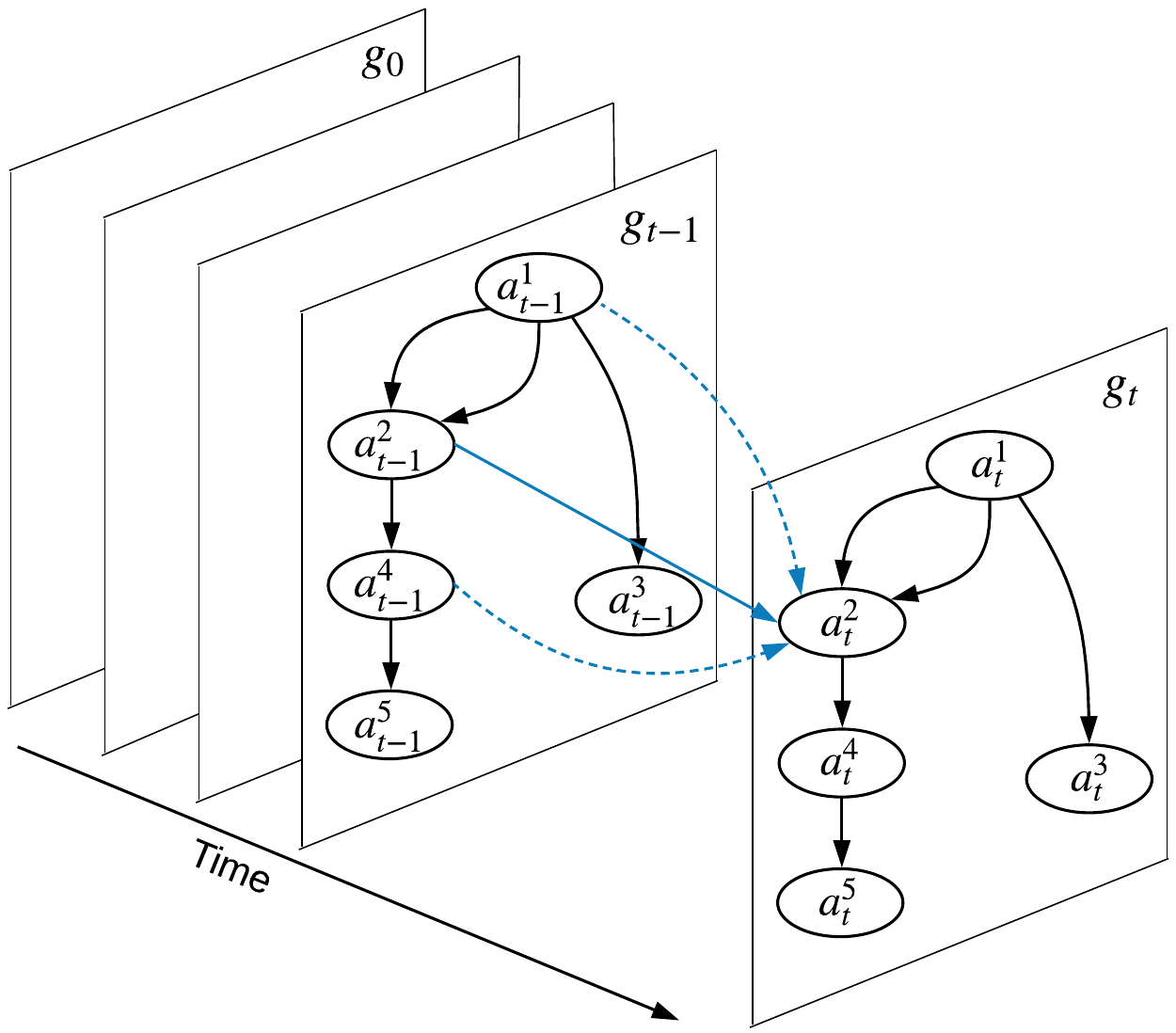}
\caption{Architecture of the graph recurrent network.}
\label{fig:6_encoder}
\end{figure}

Figure \ref{fig:6_encoder} shows the overall structure of our graph recurrent network for encoding AMR graphs, which follows Chapter \ref{chap:amrgen}.
Formally, given an AMR graph $\boldsymbol{G}=(\boldsymbol{V}, \boldsymbol{E})$, we use a hidden state vector $\boldsymbol{a}^j$ to represent each node $v_j \in \boldsymbol{V}$. 
The state of the graph can thus be represented as:
\begin{equation}
\boldsymbol{g} = \{\boldsymbol{a}^j\}|_{v_j \in \boldsymbol{V}}
\end{equation}
The GRN encoding exactly follows Chapter \ref{sec:5_encoder}, and it generates the final-state graph representation after $T$ message passing steps:
\begin{equation}
\boldsymbol{g}_T = \{\boldsymbol{a}_T^j\}|_{v_j \in \boldsymbol{V}}
\end{equation}

\subsection{Incorporating AMR information with a doubly-attentive decoder}

There is no one-to-one correspondence between AMR nodes and source words.
To incorporate additional knowledge from an AMR graph, an external attention model is adopted over the baseline model.
In particular, the attention memory from the AMR graph is the last graph state $\boldsymbol{g}_T = \{\boldsymbol{a}_T^j\}|_{v_j \in \boldsymbol{V}}$.
In addition, the contextual vector based on the graph state is calculated as:
\begin{align*}
\tilde{\epsilon}_{m,i} &= \tilde{\boldsymbol{v}}_2^\intercal \tanh(\boldsymbol{W}_a \boldsymbol{a}_T^i + \tilde{\boldsymbol{W}}_s \boldsymbol{s}_m + \tilde{\boldsymbol{b}}_2) \\
\tilde{\alpha}_{m,i} &= \frac{\exp(\tilde{\epsilon}_{m,i})}{\sum_{j=1}^N\exp(\tilde{\epsilon}_{m,j})} 
\end{align*}
$\boldsymbol{W}_a$, $\tilde{\boldsymbol{W}}_s$, $\tilde{\boldsymbol{v}}_2$ and $\tilde{\boldsymbol{b}}_2$ are model parameters.
The new context vector $\tilde{\boldsymbol{\zeta}}_m$ is calculated via $\sum_{i=1}^N \tilde{\alpha}_{m,i} \boldsymbol{a}_T^i$.
Finally, $\tilde{\boldsymbol{\zeta}}_m$ is incorporated into the calculation of the output probability distribution over the target vocabulary (previously defined in Equation \ref{eq:6_pvocab}):
\begin{equation}
\boldsymbol{P}_{vocab} = \textrm{softmax}(\boldsymbol{V}_3[\boldsymbol{s}_m,\boldsymbol{\zeta}_m,\tilde{\boldsymbol{\zeta}}_m]+\boldsymbol{b}_3)
\end{equation}

\section{Training}

Given a set of training instances $\{(\boldsymbol{X}^{(1)}, \boldsymbol{Y}^{(1)}),$ $(\boldsymbol{X}^{(2)}, \boldsymbol{Y}^{(2)}), \dots\}$,
we train our models using the cross-entropy loss over each gold-standard target sequence $\boldsymbol{Y}^{(j)}=y_1^{(j)}, y_2^{(j)}, \dots, y_M^{(j)}$:
\begin{equation*}
l = -\sum_{m=1}^M \log p(y_m^{(j)}|y_{m-1}^{(j)},\dots,y_1^{(j)},\boldsymbol{X}^{(j)};\boldsymbol{\theta})
\end{equation*}
$\boldsymbol{X}^{(j)}$ represents the inputs for the $j$th instance, which is a source sentence for our baseline, or a source sentence paired with an automatically parsed AMR graph for our model. $\boldsymbol{\theta}$ represents the model parameters.

\section{Experiments}

We empirically investigate the effectiveness of AMR for English-to-German translation.

\subsection{Setup}

\begin{table}
\centering
\begin{tabular}{lccc}
\hline
Dataset & \#Sent. & \#Tok. (EN) & \#Tok. (DE) \\
\hline
NC-v11 & 226K & 6.4M & 7.3M \\
Full & 4.17M & 109M & 118M \\
News2013 & 3000 & 84.7K & 95.6K \\
News2016 & 2999 & 88.1K & 98.8K \\
\hline
\end{tabular}
\caption{Statistics of the dataset. Numbers of tokens are after BPE processing.}
\label{tab:6_stat}
\end{table}

\begin{table}
\centering
\begin{tabular}{lcccc}
\hline
Dataset & EN-ori & EN & AMR & DE \\
\hline
NC-v11 & 79.8K & 8.4K & 36.6K & 8.3K \\
Full & 874K & 19.3K & 403K & 19.1K \\
\hline
\end{tabular}
\caption{Sizes of vocabularies. \emph{EN-ori} represents original English sentences without BPE.}
\label{tab:6_stat_vocab}
\end{table}

\subparagraph{Data}
We use the WMT16\footnote{http://www.statmt.org/wmt16/translation-task.html} English-to-German dataset, which contains around 4.5 million sentence pairs for training.
In addition, we use a subset of the full dataset (News Commentary v11 (NC-v11), containing around 243 thousand sentence pairs) for development and additional experiments.
For all experiments, we use newstest2013 and newstest2016 as the development and test sets, respectively.

To preprocess the data, the tokenizer from Moses\footnote{http://www.statmt.org/moses/} is used to tokenize both the English and German sides. 
The training sentence pairs where either side is longer than 50 words are filtered out after tokenization.
To deal with rare and compound words, byte-pair encoding (BPE)\footnote{https://github.com/rsennrich/subword-nmt} \cite{sennrich-haddow-birch:2016:P16-12} is applied to both sides.
In particular, 8000 and 16000 BPE merges are used on the News Commentary v11 subset and the full training set, respectively.
On the other hand, JAMR\footnote{https://github.com/jflanigan/jamr} \cite{flanigan-EtAl:2016:SemEval} is adopted to parse the English sentences into AMRs before BPE is applied.
The statistics of the training data and vocabularies after preprocessing are shown in Table \ref{tab:6_stat} and \ref{tab:6_stat_vocab}, respectively.
For the experiments with the full training set, we used the top 40K of the AMR vocabulary, which covers more than 99.6\% of the training set.

For our dependency-based and SRL-based baselines (which will be introduced in \textbf{Baseline systems}), we choose Stanford CoreNLP \cite{manning-EtAl:2014:P14-5} and IBM SIRE to generate dependency trees and semantic roles, respectively.
Since both dependency trees and semantic roles are based on the original English sentences without BPE, we used the top 100K frequent English words, which cover roughly 99.0\% of the training set.

\subparagraph{Hyperparameters}
We use the Adam optimizer \cite{kingma2014adam} with a learning rate of 0.0005.
The batch size is set to 128.
Between layers, we apply dropout with a probability of 0.2.
The best model is picked based on the cross-entropy loss on the development set.
For model hyperparameters, we set the graph state transition number to 10 according to development experiments.
Each node takes information from at most 6 neighbors. 
BLEU \cite{papineni2002bleu}, TER \cite{snover2006study} and Meteor \cite{denkowski:lavie:meteor-wmt:2014} are used as the metrics on cased and tokenized results.

For experiments with the NC-v11 subset, both word embedding and hidden vector sizes are set to 500, and the models are trained for at most 30 epochs.
For experiments with full training set, the word embedding and hidden state sizes are set to 800, and our models are trained for at most 10 epochs.
For all systems, the word embeddings are randomly initialized and updated during training.

\subparagraph{Baseline systems}
We compare our model with the following systems.
\emph{Seq2seq} represents our attention-based LSTM baseline ({\S \ref{sec:6_base}}), and \emph{Dual2seq} is our model, which takes both a sequential 
and a graph encoder and adopts a doubly-attentive decoder ({\S \ref{sec:amr}}).
To show the merit of AMR, we further contrast our model with the following baselines, all of which adopt the same doubly-attentive framework with a BiLSTM for encoding BPE-segmented source sentences: \emph{Dual2seq-LinAMR} uses another BiLSTM for encoding linearized AMRs. 
\emph{Dual2seq-Dep} and \emph{Dual2seq-SRL} adopt our graph recurrent network to encode original source sentences with dependency and semantic role annotations, respectively.
The three baselines are useful for contrasting different methods of encoding AMRs and for comparing AMRs with other popular structural information for NMT.

We also compare with Transformer \cite{NIPS2017_7181} and OpenNMT \cite{2017opennmt}, trained on the same dataset and with the same set of hyperparameters as our systems.
In particular, we compare with \emph{Transformer-tf}, one popular implementation\footnote{https://github.com/Kyubyong/transformer} of Transformer based on TensorFlow, and we choose \emph{OpenNMT-tf}, an official release\footnote{https://github.com/OpenNMT/OpenNMT-tf} of OpenNMT implemented with TensorFlow.
It is a popular open-source attention-based sequence-to-sequence system that has served as a baseline system in previous literature.
For a fair comparison, \emph{OpenNMT-tf} has 1 layer for both the encoder and the decoder, and \emph{Transformer-tf} has the default configuration (N=6), but with parameters being shared among different blocks.

\subsection{Development experiments}

\begin{figure}
\centering
\includegraphics[width=0.8\linewidth]{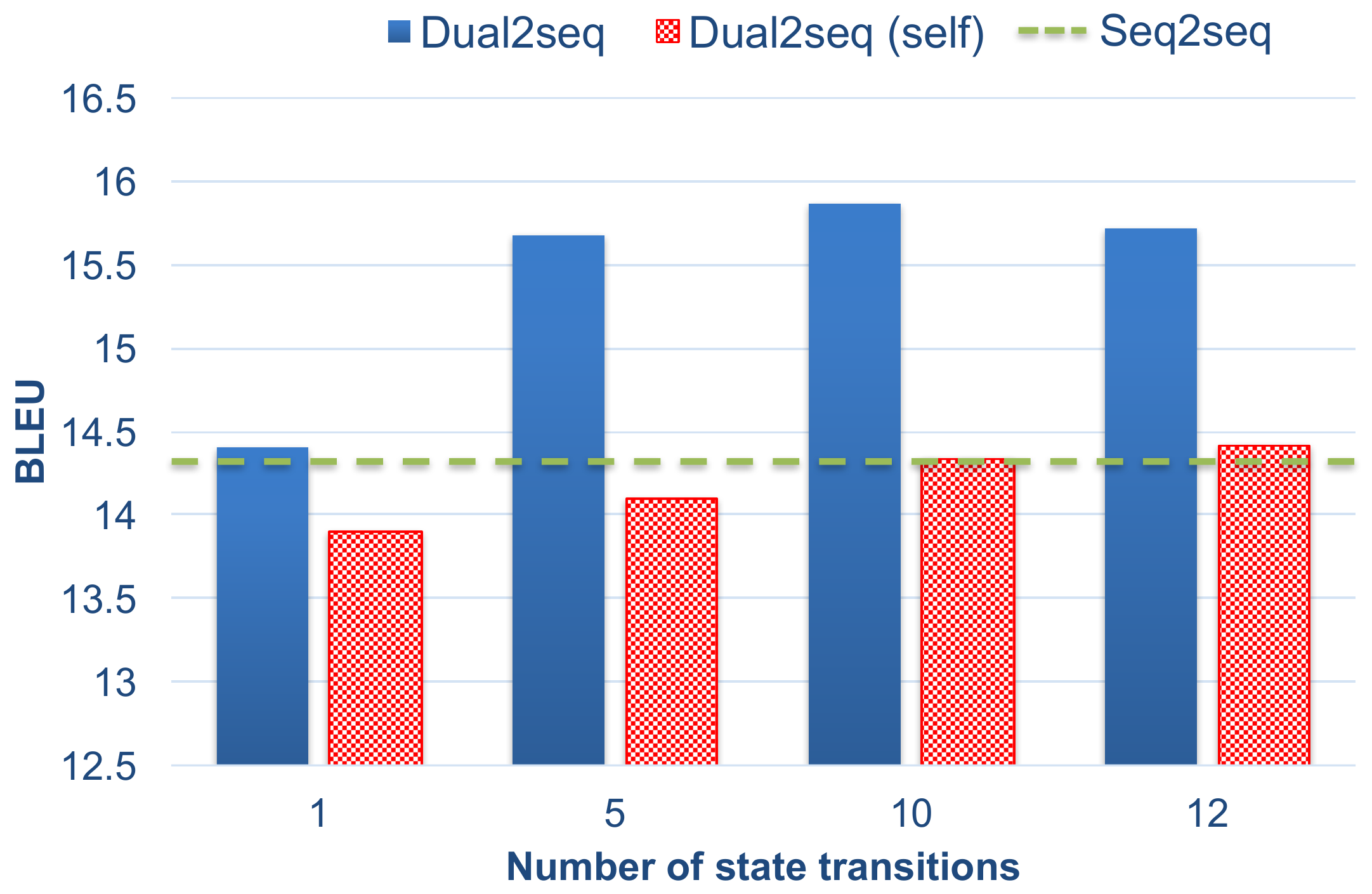}
\caption{\textsc{Dev} BLEU scores against transition steps for the graph encoders. 
The state transition is not applicable to \emph{Seq2seq}, so we draw a dashed line to represent its performance.}
\label{fig:6_dev_ana}
\end{figure}

Figure \ref{fig:6_dev_ana} shows the system performances as a function of the number of graph state transitions on the development set.
\emph{Dual2seq (self)} represents our dual-attentive model, but its graph encoder encodes the source sentence, which is treated as a chain graph, instead of an AMR graph.
Compared with \emph{Dual2seq}, \emph{Dual2seq (self)} has the same number of parameters, but without semantic information from AMR\@.
Due to hardware limitations, we do not perform an exhaustive search by evaluating every possible state transition number, but only transition numbers of 1, 5, 10 and 12.

Our \emph{Dual2seq} shows consistent performance improvement by increasing the transition number both from 1 to 5 (roughly +1.3 BLEU points) and from 5 to 10 (roughly 0.2 BLEU points).
The former shows greater improvement than the latter, showing that the performance starts to converge after 5 transition steps.
Further increasing transition steps from 10 to 12 gives a slight performance drop.
We set the number of state transition steps to 10 for all experiments according to these observations.

On the other hand, \emph{Dual2seq (self)} shows only small improvements
by increasing the state transition number, and it does not perform better than \emph{Seq2seq}.
Both results show that the performance gains of \emph{Dual2seq} are not due to an increased number of parameters.

\begin{table}
\centering
\begin{tabular}{l|c|c|c||c|c|c}
\multirow{2}{*}{System} & \multicolumn{3}{c||}{\textsc{NC-v11}} & \multicolumn{3}{c}{\textsc{Full}} \\
     & BLEU & TER$\downarrow$ & Meteor & BLEU & TER$\downarrow$ & Meteor \\
\hline
OpenNMT-tf & 15.1 & 0.6902 & 0.3040 & 24.3 & 0.5567 & 0.4225 \\
Transformer-tf & 17.1 & 0.6647 & 0.3578 & 25.1 & 0.5537 & 0.4344 \\
\hline
Seq2seq & 16.0 & 0.6695 & 0.3379 & 23.7 & 0.5590 & 0.4258 \\
Dual2seq-LinAMR & 17.3 & 0.6530 & 0.3612 & 24.0 & 0.5643 & 0.4246 \\
Duel2seq-SRL & 17.2 & 0.6591 & 0.3644 & 23.8 & 0.5626 &  0.4223 \\
Dual2seq-Dep & 17.8 & 0.6516 & 0.3673 & 25.0 & 0.5538 & 0.4328 \\
Dual2seq & \textbf{\phantom{*}19.2*} & \textbf{0.6305} & \textbf{0.3840} & \textbf{\phantom{*}25.5*} & \textbf{0.5480} & \textbf{0.4376} \\
\end{tabular}
\caption{\textsc{Test} performance. \emph{NC-v11} represents training only with the NC-v11 data, while \emph{Full} means using the full training data. * represents significant \cite{koehn2004statistical} result ($p < 0.01$) over \emph{Seq2seq}. $\downarrow$ indicates the lower the better.}
\label{tab:6_test}
\end{table}

\subsection{Main results}
\label{sec:main_res}

Table \ref{tab:6_test} shows the \textsc{test} BLEU, TER and Meteor scores of all systems trained on the small-scale \emph{News Commentary v11} subset or the large-scale full set.
\emph{Dual2seq} is consistently better than the other systems under all three metrics, showing the effectiveness of the semantic information provided by AMR\@.
Especially, \emph{Dual2seq} is better than both \emph{OpenNMT-tf} and \emph{Transformer-tf}\@.
The recurrent graph state transition of \emph{Dual2seq} is similar to Transformer in 
that it iteratively incorporates global information.
The improvement of \emph{Dual2seq} over \emph{Transformer-tf} undoubtedly comes from the use of AMRs,
 which provide complementary information to the textual inputs of the source language.

In terms of BLEU score, \emph{Dual2seq} is significantly better than \emph{Seq2seq} in both settings, which shows the effectiveness of incorporating AMR information.
In particular, the improvement is much larger under the small-scale setting (+3.2 BLEU) than that under the large-scale setting (+1.7 BLEU).
This is an evidence that structural and coarse-grained semantic information encoded in AMRs can be more helpful when training data are limited.

When trained on the NC-v11 subset, the gap between \emph{Seq2seq} and \emph{Dual2seq} under Meteor (around 5 points) is greater than that under BLEU (around 3 points).
Since Meteor gives partial credit to outputs that are synonyms to the reference or share identical stems, one possible explanation is that the structural information within AMRs helps to better translate the concepts from the source language, which may be synonyms or paronyms of reference words.

As shown in the second group of Table \ref{tab:6_test}, we further compare our model with other methods of leveraging syntactic or semantic information. 
\emph{Dual2seq-LinAMR} shows much worse performance than our model and only slightly outperforms the \emph{Seq2seq} baseline.
Both results show that simply taking advantage of the AMR concepts without their relations does not help very much.
One reason may be that AMR concepts, such as {\em John} and {\em Mary}, also appear in the textual input, and thus are also encoded by the other (sequential) encoder.\footnote{AMRs can contain multi-word concepts, such as {\em New York City}, but they are in the textual input.}
The gap between \emph{Dual2seq} and \emph{Dual2seq-LinAMR} comes from modeling the relations between concepts, which can be helpful for deciding target word order by enhancing the relations in source sentences.
We conclude that properly encoding AMRs is necessary to make them useful.

Encoding dependency trees instead of AMRs, \emph{Dual2seq-Dep} shows a larger performance gap with our model (17.8 vs 19.2) on small-scale training data than on large-scale training data (25.0 vs 25.5).
It is likely because AMRs are more useful on alleviating data sparsity than dependency trees, since words are lemmatized into unified concepts when parsing sentences into AMRs.
For modeling long-range dependencies, AMRs have one crucial advantage over dependency trees by modeling concept-concept relations more directly.
It is because AMRs drop function words, thus the distances between concepts are generally closer in AMRs than in dependency trees.
Finally, \emph{Dual2seq-SRL} is less effective than our model, because
the annotations labeled by SRL are a subset of AMRs.

We outperform \citet{diego-EtAl:2018:PAPERS} on the same datasets,
although our systems vary in a number of respects.
When trained on the \emph{NC-v11} data, they show BLEU scores of 14.9 only with their BiLSTM baseline, 16.1 using additional dependency information, 15.6 using additional semantic roles and 15.8 taking both as additional knowledge.
Using \emph{Full} as the training data, the scores become 23.3, 23.9, 24.5 and 24.9, respectively.
In addition to the different semantic representation being used (AMR vs SRL), \citet{diego-EtAl:2018:PAPERS} laid graph convolutional network (GCN) \cite{kipf2017semi} layers on top of a bidirectional LSTM (BiLSTM) layer, and then concatenated layer outputs as the attention memory. 
GCN layers encode the semantic role information, while BiLSTM layers encode the input sentence in the source language, and the concatenated hidden states of both layers contain information from both semantic role and source sentence.
For incorporating AMR, since there is no one-to-one word-to-node correspondence between a sentence and the corresponding AMR graph, we adopt separate attention models.
Our BLEU scores are higher than theirs, but we cannot conclude that the advantage primarily comes from AMR.

\subsection{Analysis}
\label{sec:analyze}

\begin{table}[t]
\centering
\begin{tabular}{l|c}
AMR Anno. & BLEU  \\
\hline
Automatic & 16.8 \\
Gold & \textbf{\phantom{*}17.5*} \\
\end{tabular}
\caption{BLEU scores of \emph{Dual2seq} on the \emph{little prince} data, when gold or automatic AMRs are available.}
\label{tab:6_amr_accu_ana}
\end{table}

\subparagraph{Influence of AMR parsing accuracy}
To analyze the influence of AMR parsing on our model performance, we further evaluate on a test set where the gold AMRs for the English side are available. 
In particular, we choose the \emph{Little Prince} corpus, which contains 1562 sentences with gold AMR annotations.\footnote{https://amr.isi.edu/download.html}
Since there are no parallel German sentences, we take a German-version \emph{Little Prince} novel, and then perform manual sentence alignment. 
Taking the whole \emph{Little Prince} corpus as the test set, we measure the influence of AMR parsing accuracy by evaluating on the test set when gold or automatically-parsed AMRs are available.
The automatic AMRs are generated by parsing the English sentences with JAMR.

Table \ref{tab:6_amr_accu_ana} shows the BLEU scores of our \emph{Dual2seq} model taking gold or automatic AMRs as inputs.
Not listed in Table \ref{tab:6_amr_accu_ana}, \emph{Seq2seq} achieves a BLEU score of 15.6, which is 1.2 BLEU points lower than using automatic AMR information.
The improvement from automatic AMR to gold AMR (+0.7 BLEU) is significant, which shows that the translation quality of our model can be further improved with an increase of AMR parsing accuracy.
However, the BLEU score with gold AMR does not indicate the potentially best performance that our model can achieve.
The primary reason is that even though the test set is coupled with gold AMRs, the training set is not.
Trained with automatic AMRs, our model can learn to selectively trust the AMR structure.
An additional reason is the domain difference: the \emph{Little Prince} data are in the literary domain while our training data are in the news domain.
There can be a further performance gain if the accuracy of the automatic AMRs on the training set is improved.

\begin{figure}
\centering
\includegraphics[width=0.8\linewidth]{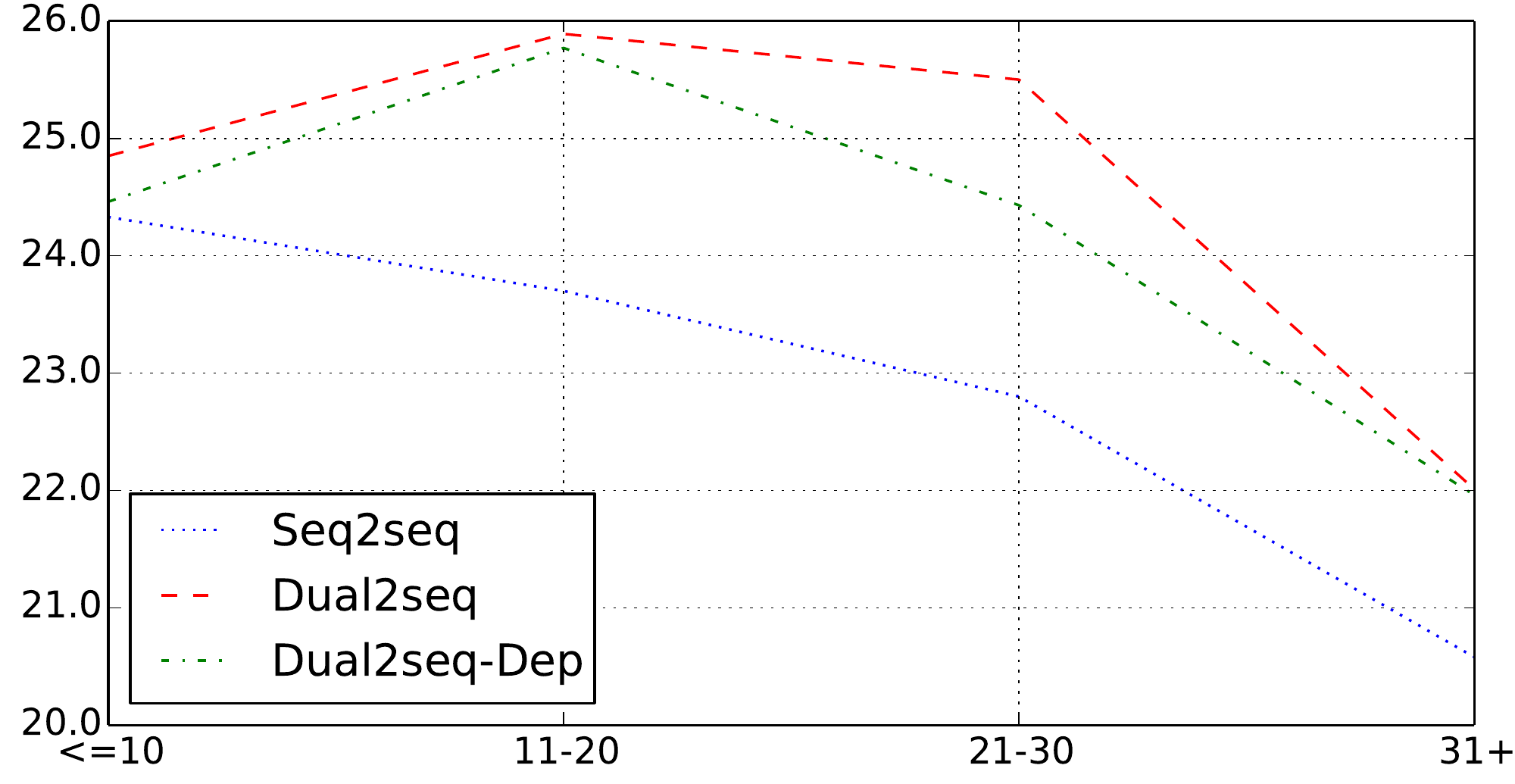}
\caption{Test BLEU score of various sentence lengths}
\label{fig:6_category}
\end{figure}

\subparagraph{Performance based on sentence length}
We hypothesize that AMRs should be more beneficial for longer sentences:
those are likely to contain long-distance dependencies (such as discourse information and predicate-argument structures) which may not be adequately captured by linear
chain RNNs but are directly encoded in AMRs.
To test this, we partition the test data into four buckets
by length and calculate BLEU for each of them. 
Figure \ref{fig:6_category} shows the performances of our model along with \emph{Dual2seq-Dep} and \emph{Seq2seq}.
Our model outperforms the \emph{Seq2seq} baseline rather uniformly across all buckets, except for the first one where they are roughly equal. 
This may be surprising.
On the one hand, \emph{Seq2seq} fails to capture some dependencies for medium-length instances;
on the other hand AMR parses are more noisy for longer sentences, which prevents us from obtaining extra improvements with AMRs.

Dependency trees have been proved useful in capturing long-range dependencies.
Figure \ref{fig:6_category} shows that AMRs are comparatively better than dependency trees, especially on medium-length (21-30) sentences.
The reason may be that the AMRs of medium-length sentences are much more accurate than longer sentences, thus are better at capturing the relations between concepts.
On the other hand, even though dependency trees are more accurate than AMRs, they still fail to represent relations for long sentences.
It is likely because relations for longer sentences are more difficult to detect.
Another possible reason is that dependency trees do not incorporate coreferences, which AMRs consider.

\begin{figure}[t]
{\footnotesize
\begin{tabularx}{\textwidth}{X}
\hline
\textbf{AMR}: (s2 / say-01
       :ARG0 (p3 / person
             :ARG1-of (h / have-rel-role-91
                   :ARG0 (p / person
                         :ARG1-of (m2 / meet-03
                               :ARG0 (t / they)
                               :ARG2 15)
                         :mod (m / mutual))
                   :ARG2 (f / friend))
             :name (n2 / name
                   :op1 "Carla"
                   :op2 "Hairston"))
       :ARG1 (a / and
             :op1 (p2 / person
                   :name (n / name
                         :op1 "Lamb")))
       :ARG2 (s / she)
       :time 20) \\
\textbf{Src}: Carla Hairston said she was 15 and Lamb was 20 when they met through mutual friends\\
\textbf{Ref}: Carla Hairston sagte , sie war 15 und Lamm war 20 , als sie sich durch gemeinsame Freunde trafen .\\
\textbf{Dual2seq}: Carla Hairston sagte , sie war 15 und Lamm war 20 , als sie sich durch gegenseitige Freunde trafen .\\
\textbf{Seq2seq}: Carla Hirston sagte , sie sei 15 und Lamb 20 , als sie durch gegenseitige Freunde trafen .\\
\hline
\textbf{AMR}: (s / say-01
       :ARG0 (m / media
             :ARG1-of (l / local-02))
       :ARG1 (c2 / come-01
             :ARG1 (v / vehicle
                   :mod (p / police))
             :manner (c3 / constant)
             :path (a / across
                   :op1 (r / refugee
                         :mod (n2 / new)))
             :time (s2 / since
                   :op1 (t3 / then))
             :topic (t / thing
                   :name (n / name
                         :op1 (c / Croatian)
                         :op2 (t2 / Tavarnik))))) \\
\textbf{Src}: Since then , according to local media , police vehicles are constantly coming across new refugees in Croatian Tavarnik .\\
\textbf{Ref}: Laut lokalen Medien treffen seitdem im kroatischen Tovarnik ständig Polizeifahrzeuge mit neuen Flüchtlingen ein .\\
\textbf{Dual2seq}: Seither kommen die Polizeifahrzeuge nach den örtlichen Medien ständig über neue Flüchtlinge in Kroatische Tavarnik .\\
\textbf{Seq2seq}: Seitdem sind die Polizeiautos nach den lokalen Medien ständig neue Flüchtlinge in Kroatien Tavarnik .\\
\hline
\textbf{AMR}: (b2 / breed-01
       :ARG0 (p2 / person
             :ARG0-of (h / have-org-role-91
                   :ARG2 (s3 / scientist)))
       :ARG1 (w2 / worm)
       :ARG2 (s2 / system
             :ARG1-of (c / control-01
                   :ARG0 (b / burst-01
                         :ARG1 (w / wave
                               :mod (s / sound)))
                   :ARG1-of (p / possible-01))
             :ARG1-of (n / nervous-01)
             :mod (m / modify-01
                   :ARG1 (g / genetics)))) \\
\textbf{Src}: Scientists have bred worms with genetically modified nervous systems that can be controlled by bursts of sound waves\\
\textbf{Ref}: Wissenschaftler haben Würmer mit genetisch veränderten Nervensystemen gezüchtet , die von Ausbrüchen von Schallwellen gesteuert werden können\\
\textbf{Dual2seq}: Die Wissenschaftler haben die Würmer mit genetisch veränderten Nervensystemen gezüchtet , die durch Verbrennungen von Schallwellen kontrolliert werden können\\
\textbf{Seq2seq}: Wissenschaftler haben sich mit genetisch modifiziertem Nervensystem gezüchtet , die durch Verbrennungen von Klangwellen gesteuert werden können\\
\hline
\end{tabularx}}
\caption{Sample system outputs}\label{fig:6_ex}
\end{figure}

\subparagraph{Human evaluation}
We further study the translation quality of predicate-argument structures by conducting a human evaluation on 100 instances from the testset.
In the evaluation, translations of both \emph{Dual2seq} and \emph{Seq2seq}, together with the source English sentence, the German reference, and an AMR are provided to a German-speaking annotator to decide which translation better captures the predicate-argument structures in the source sentence.
To avoid annotation bias, translation results of both models are swapped for some instances, and the German annotator does not know which model each translation belongs to.
The annotator either selects a ``winner'' or makes a ``tie'' decision, meaning that both results are equally good.

Out of the 100 instances, \emph{Dual2seq} wins on 46, \emph{Seq2seq} wins on 23, and there is a tie on the remaining 31.
\emph{Dual2seq} wins on almost half of the instances,
about twice as often as \emph{Seq2seq} wins, indicating that AMRs help in translating the predicate-argument structures on the source side.

\subparagraph{Case study}
The outputs of the baseline system (\emph{Seq2seq}) and our final system (\emph{Dual2seq})
are shown in Figure~\ref{fig:6_ex}.
In the first sentence, the AMR-based Dual2seq system correctly produces
the reflexive pronoun {\em sich} as an argument of 
the verb {\em trafen} ({\em meet}), despite the
distance between the words in the system output, and
despite the fact that the equivalent English words
{\em each other} do not appear in the system output.
This is facilitated by the argument structure in the AMR analysis.

In the second sentence, 
the AMR-based Dual2seq system produces an overly literal
translation for the English phrasal verb 
{\em come across}.  The Seq2seq
translation, however, incorrectly
states that the police vehicles {\em are} refugees.
The difficulty for the Seq2seq probably derives in part 
from the fact that {\em are} and {\em coming} are separated by
the word {\em constantly} in the input, while the main predicate
is clear in the AMR representation.

In the third sentence, the Dual2seq system correctly translates
the object of {\em breed} as {\em worms}, while the Seq2seq
translation incorrectly states that the scientists breed {\em themselves}.
Here the difficulty is likely the distance between the 
object and the verb in the German output, which causes the
Seq2seq system to lose track of the correct input position to translate.

\section{Conclusion}

We showed that AMRs can improve neural machine translation.
In particular, the structural semantic information from AMRs can be complementary to the source textual input by introducing a higher level of information abstraction.
A graph recurrent network (GRN) is leveraged to encode AMR graphs without breaking the original graph structure, and a sequential LSTM is used to encode the source input.
The decoder is a doubly-attentive LSTM, taking the encoding results of both the graph encoder and the sequential encoder as attention memories.
Experiments on a standard benchmark showed that AMRs are helpful regardless of the sentence length, and are more effective than other more popular choices, such as dependency trees and semantic roles.